\documentclass{article}

\usepackage{graphicx} %
\usepackage{subfigure}
\usepackage{wrapfig}
\usepackage{xspace}
\usepackage{caption}

\usepackage{algorithm}
\usepackage{algpseudocode}

\usepackage{color}
\usepackage{enumitem}
\usepackage[pagebackref=true,breaklinks=true,colorlinks,citecolor=gray]{hyperref}

\usepackage{fullpage}
\usepackage{xcolor}
\usepackage{tikz}
\usetikzlibrary{arrows.meta}
\usepackage{longtable,booktabs,array}
\usepackage[numbers]{natbib}
\usepackage[normalem]{ulem}

\usepackage{Definitions}

\renewcommand{\le}{\leqslant}

\renewcommand{\ge}{\geqslant}

\usepackage{amsmath,amsfonts,bm}

\def\Figref#1{Figure~\ref{#1}}

\def\Secref#1{Section~\ref{#1}}

\def\eqref#1{equation~\ref{#1}}
\def\Eqref#1{Equation~\ref{#1}}

\def\Algref#1{Algorithm~\ref{#1}}

\def\1{\bm{1}}

\def\rb{{\textnormal{b}}}

\def\va{{\bm{a}}}
\def\vb{{\bm{b}}}
\def\vc{{\bm{c}}}

\def\vf{{\bm{f}}}

\def\vv{{\bm{v}}}
\def\vw{{\bm{w}}}
\def\vx{{\bm{x}}}
\def\vy{{\bm{y}}}
\def\vz{{\bm{z}}}

\DeclareMathAlphabet{\mathsfit}{\encodingdefault}{\sfdefault}{m}{sl}
\SetMathAlphabet{\mathsfit}{bold}{\encodingdefault}{\sfdefault}{bx}{n}

\newcommand{\softmax}{\mathrm{softmax}}
\newcommand{\sigmoid}{\mathrm{sigm}}

\title{An Energy-Based Perspective on Machine Learning}

\begin{document}
\author{%
  Bo Dai \\
  Georgia Institute of Technology \\
  \texttt{bodai@cc.gatech.edu}
  \and
  Yilun Du \\
  Harvard University \\
  \texttt{ydu@seas.harvard.edu}
}
\maketitle

\begin{abstract}
Energy-based models (EBMs) define a distribution through a scalar energy, $p_\theta(\vx)\propto\exp\rbr{-E_\theta(\vx)}$, and are usually presented as one model family among many. This tutorial has a broader goal: beyond introducing EBMs, it uses them as a single lens that connects the different aspects of machine learning. Any EBM is specified by three choices, the energy, which determines what the model represents; the inference procedure, which samples from or optimizes the energy; and the learning method, which fits the energy to data. We first develop these components, including sampling, variational inference, and learning by maximum likelihood, noise contrastive estimation, pseudolikelihood, score matching, and adversarial training, and show that the learning methods share one contrastive structure. We then show that discriminative models, generative models such as GANs, normalizing flows, diffusion and autoregressive models, representation learning methods such as contrastive learning, word2vec, BERT, denoising auto-encoders, and diffusion representations, reinforcement learning with EBM policies and world models, inverse reinforcement learning, and planning, as well as inference-time computation with verifiers, are all EBMs that differ in these three choices, mostly in how they handle the partition function and sampling. This view explains existing methods, organizes them into a common framework, and suggests new algorithms by replacing one component while keeping the others, with composition of energies as a principled way to combine learned models at inference time.
\end{abstract}

\newpage
{\hypersetup{linkcolor=black}\tableofcontents}
\newpage

\section{Introduction}

Machine learning is often divided into supervised learning, generative modeling, representation learning, and decision making, each with its own models and training objectives. However, these areas share a common problem: learning relationships among variables and using those relationships to infer unknown quantities. Energy-based models (EBMs) provide a unified perspective on this problem by representing the compatibility of a configuration through a scalar energy function. Low-energy configurations correspond to plausible observations, compatible representations, or desirable outcomes. Learning shapes this energy landscape, while inference uses it to produce predictions, samples, or actions.

This perspective connects three ideas: \emph{representation}, \emph{inference}, and \emph{composition}. An energy function represents knowledge through relationships and constraints among variables. Inference uses this knowledge by searching for compatible configurations, either through optimization or sampling. Composition combines energy functions to express requirements that must be considered together. Separating these roles allows learned knowledge to be reused in different computations and combined with new information at inference time. For example, a joint energy over observations and latent variables can support both generating observations from a latent description and inferring a description from an observation. An energy over trajectories can be combined with a new goal to formulate a planning problem.

In this tutorial, which grew from the course \href{https://bo-dai.github.io/CSE6243-fall2024/}{CSE 6243} taught by Bo Dai at Georgia Tech, we develop this perspective across different generative models, representation learning methods, and planning and reinforcement learning methods. We primarily consider the probabilistic formulation, in which an energy defines an unnormalized distribution through $p_\theta(\vx)\propto\exp(-E_\theta(\vx))$. For generative modeling, this provides a common language for comparing models that parameterize densities, learn scores, or construct sampling procedures. Autoregressive models specify conditional factors whose negative logarithms sum to an energy, while invertible normalizing flows specify densities through a change of variables. Diffusion models learn gradients of $\log$ densities at different noise levels. These connections clarify both the relationships among generative models and the different assumptions they make about representation and inference. They also provide a unified way to develop new generative models.

For representation learning, energies express compatibility between observations and latent descriptions, or between different views of the same data. Contrastive methods learn features through scores for matched and mismatched observations, while reconstruction and denoising objectives train representations to support prediction of missing or corrupted information. Viewing these objectives through their associated energies connects representation learning to conditional prediction and generative modeling. It also suggests how learned generative models can provide objectives for inferring representations of new observations.

For planning and decision-making, energies score candidate actions or trajectories according to their predicted consequences. Costs and constraints contribute energy terms, while rewards enter with the opposite sign. Planning searches for configurations that satisfy these requirements, and entropy-regularized formulations connect this search to probabilistic inference. Learned dynamics, behavior models, and task objectives can therefore contribute different components of a common inference problem.

For inference-time computation, energies score different possible solutions for a question. Different methods of inference-time computation such as selecting among sampled responses, searching for partial solutions, or iteratively refining a candidate, correspond to different procedures to minimize this energy. Fine-tuning corresponds to directly fitting this combined energy, paying the cost of inference during training rather than for every input.

Composition enables us to use these connections to solve new tasks. Energy functions over compatible variables can be added to combine their requirements, corresponding to a product of their unnormalized densities when the resulting distribution is normalizable. For example, an energy describing plausible robot trajectories can be combined with energies expressing a destination and obstacle constraints. The resulting inference problem uses learned knowledge together with task-specific information, without requiring a separate predictor to be trained for each combination. This ability to represent reusable components and reason with their combinations motivates the energy-based perspective. Its practical success depends on the accuracy and relative scales of the components, as well as the ability to optimize or sample from the composed landscape.

We begin with the formulation of EBMs and methods for inference and learning~(\Secref{sec:prelim}). We then examine discriminative models~(\Secref{sec:disc_model}), generative models~(\Secref{sec:gen_model}), representation learning~(\Secref{sec:repr}), reinforcement learning and planning~(\Secref{sec:rl}), and inference-time compute (\Secref{sec:inference_time}) through this common framework, and close with a discussion~(\Secref{sec:discussion}). Throughout, we ask what each method represents, how inference uses that representation, and how its components can be combined or adapted to new tasks. Our aim is to explain connections among existing methods and show how those connections can guide the design of models that learn reusable knowledge and reason with it at inference time.

\section{A Brief Introduction to Energy-Based Models}\label{sec:prelim}

\subsection{Energy-Based Models}

We start with energy-based models of a single variable, which are also known as Gibbs distributions~\citep{georgii2011gibbs}.
Formally, the EBM defines the density as
\begin{equation}\label{eq:ebm}
    p_\theta(\vx) = \frac{\exp\rbr{-E_\theta(\vx)}}{Z_\theta}, \quad \text{with} \quad Z_\theta = \int \exp\rbr{-E_\theta(\vx)} d\vx,
\end{equation}
where $E_\theta(\vx)$, known as the energy function, is some function parametrized with $\theta$, and $Z_\theta$ is known as the partition function, which is often unknown and intractable to compute. The EBM~\Eqref{eq:ebm} defines a valid distribution, satisfying the positivity $p_\theta\rbr{\vx}\ge 0$ and normalization $\int p_\theta\rbr{\vx} d\vx = 1$. The value of the density $p_\theta(\vx)$ in~\Eqref{eq:ebm} up to a multiplicative constant is fully determined by the energy function $E_\theta(\vx)$, enabling us to sample from the density even without knowledge of the normalization constants.

EBMs can be easily generalized to joint or conditional models with multiple variables, respectively, \ie, 
\begin{align}
    p_\theta\rbr{\vx, \vy} = \frac{\exp\rbr{-E_\theta\rbr{\vx, \vy}}}{Z_\theta} \quad \text{with} \quad Z_\theta = \int \exp\rbr{-E_\theta(\vx, \vy)} d\vx d\vy, \label{eq:ebm_joint}\\
    p_\theta\rbr{\vy\mid\vx} = \frac{\exp\rbr{-E_\theta\rbr{\vx, \vy}}}{Z_\theta(\vx)} \quad \text{with} \quad Z_\theta(\vx) = \int \exp\rbr{-E_\theta(\vx, \vy)} d\vy. \label{eq:ebm_cond}
\end{align}
By comparing the joint and conditional EBMs in~\Eqref{eq:ebm_joint} and~\Eqref{eq:ebm_cond}, we emphasize the importance of partition function, leading to one key property of EBMs: with the same energy function but different partition functions, we can represent and sample from the corresponding conditional or joint distributions. Although this property seems trivial, it will play a vital role in enabling compositionality, adaptivity,  and improving training data sample complexity.

In comparison to many classes of generative models, EBMs~\Eqref{eq:ebm} directly model the probability density, without any forms of factorization of the distribution to enable efficient sampling. As a result, for many distributions, it can be simpler to represent the distributions as EBMs than other factorized parameterizations, at the cost of much more expensive inference.  Before we dive into the detailed discussion about compositional and adaptive properties of EBMs, we first provide examples of general EBMs. In fact, a variety of classic machine learning models can be represented as special forms of EBMs. 

\paragraph{Residual EBM.} For any EBM in~\Eqref{eq:ebm}, it can also be equivalently written as
\begin{equation}\label{eq:residual_ebm}
    p_\theta(\vx) = p_0\rbr{\vx}\frac{\exp\rbr{-\Etil_\theta\rbr{\vx}}}{Z_\theta}, \quad \text{with}\quad \Etil_\theta\rbr{\vx} =  E_\theta(\vx)+\log p_0\rbr{\vx},
\end{equation}
where $p_0\rbr{\vx}$ is some base probability measure. The equivalent model~\Eqref{eq:residual_ebm} is known as a \emph{residual EBM}~\citep{deng2020residual,arbel2020generalized}. We emphasize that although the vanilla EBM~\Eqref{eq:ebm} and residual EBM~\Eqref{eq:residual_ebm} share the same modeling flexibility, the residual EBM is more convenient when a good base distribution is available, as in the energy-inspired models of~\Secref{sec:eim} and the KL-regularized policies of~\Secref{sec:reg_rl}.

\paragraph{Exponential Family.} The exponential family is one of the most important classes of distributions in statistics and machine learning~\citep{wainwright2008graphical}, which possesses a number of useful properties~\citep{brown1986fundamentals} and is a special form of an EBM. 
A probability distribution belongs to the exponential family if it can be written in the following canonical form:
\[
p_\theta(\vx) = h(\vx) \exp\left( \theta^\top \mathbf{T}(\vx) - A(\theta) \right),
\]
where $\theta$ is the natural parameter, $\Tb\rbr{\vx}$ are the sufficient statistics, $h\rbr{\vx}$ denotes some base measure of $\vx$, and $A\rbr{\theta} = \log \int \exp\left( \theta^\top \mathbf{T}(\vx)\right) h\rbr{\vx}d\vx$ is the $\log$-partition function. The exponential family includes many commonly used distributions, such as the Gaussian, Poisson and multinomial distributions, to name a few. 

Obviously, the exponential family is a special case of EBMs whose energy is linear in $\theta$, \ie, $E_\theta\rbr{\vx} = -\theta^\top \Tb\rbr{\vx} - \log h\rbr{\vx}$.

\paragraph{Maximum Entropy Models.} The Maximum Entropy Model (MaxEnt)~\citep{altun2006unifying,dudik2007maximum} is a statistical modeling technique that selects the probability distribution that has the highest entropy among all distributions that satisfy a set of observed constraints derived from the training data. Formally, MaxEnt is defined as
\begin{align}\label{eq:maxent}
    \max_{q\in \Delta\rbr{\Xcal}}&\,\, H\rbr{q}\defeq -\int q(\vx)\log q(\vx)d\vx\\
    \st& \,\, \EE_{q(\vx)}\sbr{f_i(\vx)} = a_i, \quad \forall i= 1,\ldots, k,
\end{align}
which induces the Lagrangian as 
\begin{equation}
    L\rbr{q, \lambda} = -H(q) - \sum_{i=1}^k \lambda_i \rbr{\EE_{q(\vx)}\sbr{f_i(\vx)} -  a_i} - \lambda_0\rbr{\int q\rbr{\vx}d\vx -1}.
\end{equation}
Taking the partial derivative of $L\rbr{q, \lambda}$ w.r.t. $q\rbr{\vx}$ and setting it to zero gives
\begin{equation}\label{eq:maxent_sol}
    \frac{\partial L\rbr{q, \lambda}}{\partial q\rbr{\vx}} = \log q(\vx) + 1 - \sum_{i=1}^k \lambda_i f_i\rbr{\vx} - \lambda_0  = 0 \Rightarrow q^*\rbr{\vx} = \frac{\exp\rbr{\sum_{i=1}^k \lambda_i f_i\rbr{\vx}}}{Z},
\end{equation}
which obviously is an EBM with energy function $E\rbr{\vx} = -\bm{\lambda}^\top \fb\rbr{\vx}$ with parameters $\bm{\lambda}$ and features $\fb\rbr{\vx}$.

\paragraph{Markov Random Fields and Conditional Random Fields.} Markov random fields~(MRFs) are  distributions with explicit dependency among variables. Formally, an MRF defines a probability distribution on $\vx = \cbr{x_1, x_2, \ldots, x_d}$ by an undirected graph $G$, in which nodes correspond to variables $x_i$ and edges define the dependency between two variables, leading to 
\begin{equation}\label{eq:mrf}
    p(\vx) = \frac{1}{Z}\prod_{c\in C}\psi_c\rbr{x_c} = \frac{1}{Z}\exp\rbr{\sum_{c\in C} \log \psi_c\rbr{x_c}} ,
\end{equation}
with $Z$ as partition function, $C$ denoting the set of cliques of $G$, and each factor $\psi_c\rbr{\cdot}$ is a non-negative function over the variables in the corresponding subgraph. 

Similarly, conditional random fields~(CRFs) generalize MRFs and define the conditional distribution as
\begin{equation}\label{eq:crf}
    p(\vy\mid\vx) = \frac{1}{Z(\vx)}\prod_{c\in C}\psi_c\rbr{\vx, \vy} = \frac{1}{Z(\vx)}\exp\rbr{\sum_{c\in C}\log \psi_c\rbr{\vx, \vy}}
\end{equation}
Clearly, both MRF~\Eqref{eq:mrf} and CRF~\Eqref{eq:crf} can be reduced to corresponding EBMs equivalently with explicit dependency structure captured by the energy function $E\rbr{\vx} = -\sum_{c\in C} \log \psi_c\rbr{x_c}$ and $E\rbr{\vx, \vy} = -\sum_{c\in C}\log \psi_c\rbr{\vx, \vy}$.

\paragraph{Hopfield Network and (Restricted) Boltzmann Machine.} Hopfield Network was proposed as binary threshold units with recurrent connections between them. It has been revealed that if the connections are symmetric, there is a corresponding global energy function for binary $\vx$, 
\[
    E\rbr{\vx} = -\frac{1}{2} \vx^\top W\vx - \vx^\top b, \quad \text{with}\quad \vx\in \cbr{-1, +1}^d,
\]
which induces a distribution $p(\vx)\propto \exp\rbr{-E\rbr{\vx}}$ that aligns with the stationary distribution of the stochastic Hopfield network updates, \ie, $p(x_i=1\mid\vx_{-i}) = \frac{1}{1+ \exp\rbr{-2\rbr{\sum_{j\neq i}w_{ij}x_j + b_i}}}$. 
The induced stochastic rule and stationary distribution are also known as Boltzmann machine and Boltzmann distribution, respectively. 

Furthermore, if we split the variables as two disjoint groups $\cbr{\vx, \vy}$ and construct bipartite connection, we obtain the energy function, 
\begin{equation}
     E\rbr{\vx, \vy} = -\vx^\top b - \vy^\top a - \vx^\top A \vy,
\end{equation}
where $\vx$ and $\vy$ can be binary, categorical, or Gaussian. Since the structure of the energy function is restricted such that there is no intra-layer connections in $\vx$ or $\vy$, the model is named as restricted Boltzmann machine. Meanwhile, due to such special structure, the components of $\vy$ are conditionally independent given $\vx$, and vice versa.

Although EBMs define flexible distributions, it is notorious that the partition function in~\Eqref{eq:ebm_joint} and~\Eqref{eq:ebm_cond} is intractable, therefore, both sampling from EBMs and calculating the probability are not straightforward. Consequently, learning with EBMs is complicated. We discuss sampling in~\Secref{sec:inference} and learning in~\Secref{sec:learning}.

\subsection{Beyond EBMs: Generalized Maximum Entropy}\label{sec:beyond_maxent}

The maximum entropy model~\Eqref{eq:maxent} of~\Secref{sec:prelim} makes two choices: it maximizes the Shannon entropy, and it imposes the moment constraints exactly; its solution~\Eqref{eq:maxent_sol} is then an EBM in the exponential family. EBMs extend beyond both choices. Maximizing the Shannon entropy is equivalent to minimizing the KL divergence from a uniform base measure, which can be replaced by other divergences, and the exact constraints can be relaxed to approximate ones; different choices induce different model families~\citep{grunwald2004game,altun2006unifying,dudik2007maximum,frongillo2014convex}. Consider the generalized problem
\begin{equation}\label{eq:gen_maxent}
    \min_{q\in\Delta\rbr{\Xcal}}\; D\rbr{q\,\|\,p_0} \quad \st \quad \nbr{\EE_{q(\vx)}\sbr{\fb\rbr{\vx}} - \va}_{\Bcal} \le \epsilon,
\end{equation}
where $D$ is a convex divergence from a base measure $p_0$, $\nbr{\cdot}_{\Bcal}$ is a norm, and $\epsilon\ge0$ allows approximate moment matching. The maximum entropy model~\Eqref{eq:maxent} is the case of the KL divergence, a uniform $p_0$, and $\epsilon=0$.
The two choices in~\Eqref{eq:gen_maxent} change the model in different ways.
\begin{itemize}
    \item \textbf{The divergence determines the model family.} For an $f$-divergence with convex generator $\varphi$, the optimum is
    \begin{equation}\label{eq:f_family}
        D_\varphi\rbr{q\,\|\,p_0} = \int p_0(\vx)\,\varphi\rbr{\frac{q(\vx)}{p_0(\vx)}}d\vx
        \quad\Longrightarrow\quad
        q^*(\vx) = p_0(\vx)\,(\varphi')^{-1}\rbr{\bm{\lambda}^\top\fb(\vx) - \lambda_0},
    \end{equation}
    with $\lambda_0$ chosen for normalization, which is an exponential family only for $\varphi(u) = u\log u$. More generally, regular exponential families are in one-to-one correspondence with regular Bregman divergences~\citep{banerjee2005clustering}, and other divergences and generalized entropies induce generalized, or deformed, exponential families~\citep{grunwald2004game,naudts2008generalised,frongillo2014convex}. For the $\alpha$-divergence, $\varphi_\alpha(u) = \rbr{u^\alpha - \alpha u + \alpha - 1}/\rbr{\alpha(\alpha-1)}$, it is the power-law family
    \begin{equation}\label{eq:alpha_family}
        q^*(\vx) = p_0(\vx)\sbr{1 + (\alpha-1)\rbr{\bm{\lambda}^\top\fb(\vx) - \lambda_0}}_+^{1/(\alpha-1)},
    \end{equation}
    which recovers the exponential family as $\alpha\to1$, and otherwise has heavier tails or bounded support. It corresponds to the $q$-exponential family of Tsallis statistics~\citep{naudts2008generalised}. Replacing the exponential family by such a heavy-tailed family changes the behavior of the learned model, \eg, $t$-logistic regression~\citep{ding2010t} replaces the softmax of logistic regression by a $t$-exponential family, which makes the classifier robust to label noise.
    \item \textbf{The norm determines the regularization and the feature space.} With $\epsilon>0$, the dual of~\Eqref{eq:gen_maxent} becomes maximum likelihood regularized by the dual norm~\citep{altun2006unifying}, \eg, with an $\ell_1$ penalty that yields sparse models~\citep{dudik2007maximum}. Taking $\fb(\vx) = k(\vx,\cdot)$ as the feature map of a reproducing kernel Hilbert space~(RKHS) $\Hcal$ and $\Bcal$ as its norm gives the infinite-dimensional kernel exponential family $p(\vx)\propto p_0(\vx)\exp\rbr{\inner{\lambda}{k(\vx,\cdot)}_{\Hcal}}$ with an RKHS-norm penalty~\citep{altun2006unifying,canu2006kernel,sriperumbudur2017density,dai2019kernel}, an EBM whose energy is a nonparametric function in $\Hcal$.
\end{itemize}

\subsection{Adaptability and Composability of EBMs}\label{sec:compose}
A major benefit of learning EBMs is their flexibility in representing different probability distributions using the same energy function. As noted after~\Eqref{eq:ebm_cond}, a joint energy also defines the conditional distribution obtained by fixing any subset of its variables, and only the partition function changes. Energies can further be added, subtracted, and combined, which forms new distributions from learned ones without retraining~\citep{du2025learning,du2024compositional}. Below, we describe these operations on sets and on distributions, discuss when a composition is correct and how it differs from a product of experts, and show how the same operations adapt a pretrained model and how composed energies are sampled.
 
\begin{figure}[t]
    \centering
    \includegraphics[width=\textwidth]{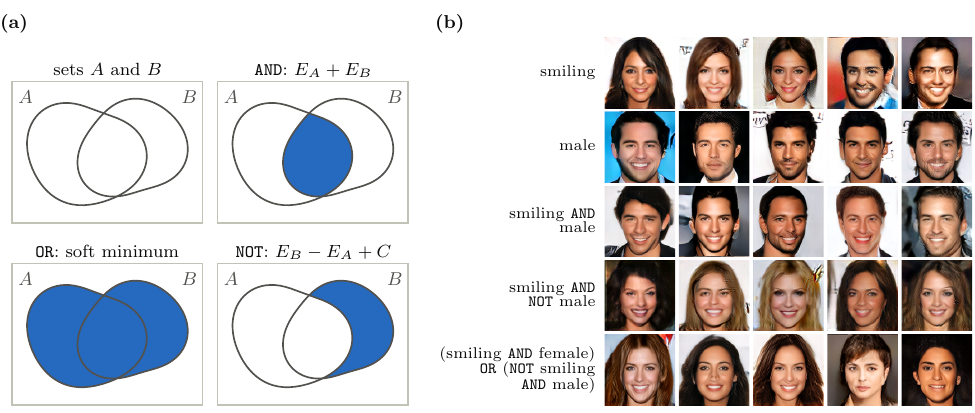}
    \caption{\textbf{Logical composition of energy functions.} \textbf{(a)}~\texttt{AND}, \texttt{OR}, and \texttt{NOT} applied to two sets $A$ and $B$; the composed energy is low in the shaded region. \textbf{(b)}~Faces sampled from compositions of conditional energies, each trained independently on one attribute of CelebA. Figures from~\citet{du2020compositional}.}
    \label{fig:compose_logic}
    \vspace{-10pt}
\end{figure}

\paragraph{Composing EBM Set Functions.} We first illustrate how composition over energy functions can be used to implement logical set operations~\citep{du2020compositional}. Consider sets $A$ and $B$ and learned energy functions $E_A(\vx)$ and $E_B(\vx)$ for these sets, for instance
\begin{equation}
E_A(\vx)=
\begin{cases}
0, & \text{for } \vx \in A\\
C, & \text{for } \vx \not\in A,\\
\end{cases}
\label{eqn:set_val}
\end{equation}
where $C$ is a large numerical value. Sampling under these learned energy functions $E_A(\vx)$ or $E_B(\vx)$ would correspond to drawing samples from either set $A$ or $B$. Logical operations on the sets then become arithmetic on the energies, as illustrated in Figure~\ref{fig:compose_logic}(a).
 
\begin{enumerate}
    \item \textbf{Logical \texttt{AND}.} The intersection $A\cap B$ is represented by the sum of the energies,
    \begin{equation}\label{eqn:compose_and}
        E_{A\wedge B}(\vx) = E_A(\vx) + E_B(\vx),
    \end{equation}
    which is $0$ on $A\cap B$ and at least $C$ elsewhere, so that low-energy samples satisfy both requirements.
    \item \textbf{Logical \texttt{OR}.} The union $A\cup B$ is represented by the soft minimum of the energies,
    \begin{equation}\label{eqn:compose_or}
        E_{A\vee B}(\vx) = -\log\rbr{\exp\rbr{-E_A(\vx)} + \exp\rbr{-E_B(\vx)}},
    \end{equation}
    which is at most $0$ on $A\cup B$ and equals $C-\log 2$ elsewhere. It differs from $\min\cbr{E_A(\vx), E_B(\vx)}$ by at most $\log 2$, so low-energy samples satisfy either requirement.
    \item \textbf{Logical \texttt{NOT}.} Negation subtracts an energy. The complement of a bounded set is unbounded, so negation is applied relative to another set: $B$ without $A$ is represented by
    \begin{equation}\label{eqn:compose_not}
        E_{B\wedge\neg A}(\vx) = E_B(\vx) - E_A(\vx) + C,
    \end{equation}
    which is $0$ on $B\setminus A$ and at least $C$ elsewhere. The subtracted energy acts as a negation operation: it raises the energy wherever $A$ holds and leaves it unchanged elsewhere.
\end{enumerate}
Each operation returns an energy, so the operations can be nested, and they apply to learned energies as well as to the idealized ones of~\Eqref{eqn:set_val}. Figure~\ref{fig:compose_logic}(b) shows faces sampled from nested compositions of conditional energies that were trained independently, each on a single attribute~\citep{du2020compositional}.

\begin{figure}[t]
    \centering
    \includegraphics[width=\textwidth]{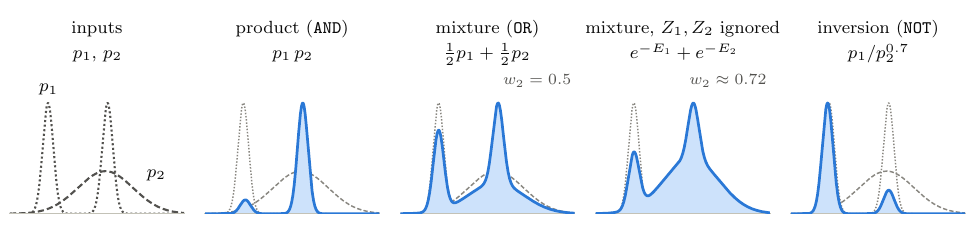}
    \caption{\textbf{Composing two one-dimensional densities.} Left: $p_1$ (dotted) and $p_2$ (dashed), repeated in gray in the other panels, where each composition (blue) is scaled to its own maximum. The product keeps the mass on which both agree, and the mixture keeps the mass of both. Adding the unnormalized densities, $e^{-E_1}+e^{-E_2}$, instead weights each component by its partition function~\Eqref{eqn:compose_mixture}, so the broad $p_2$ gets weight $w_2\approx 0.72$ rather than $\tfrac12$ (here each $E_i$ has minimum zero). The inversion $p_1/p_2^{0.7}$ keeps the mass of $p_1$ where $p_2$ is low.}
    \label{fig:compose_ops}
    \vspace{-10pt}
\end{figure}

\paragraph{Composing EBM Probability Distributions.} We can likewise combine distributions rather than sets. Product, mixture, and inversion are the distributional counterparts of \texttt{AND}, \texttt{OR}, and \texttt{NOT}, but the energies are now $\log$-densities, so their values, and not only their minimizers, matter, as illustrated in Figure~\ref{fig:compose_ops}. Let $p_i(\vx) = \exp\rbr{-E_i(\vx)}/Z_i$ for $i=1,2$.
 
\begin{enumerate}
    \item \textbf{Product of Distributions.} The product of the two distributions can be written as
\begin{equation}
    p_{\text{product}}(\vx) \propto p_1(\vx) p_2(\vx) \propto e^{-(E_1(\vx) + E_2(\vx))}.
\end{equation}
Therefore, we can construct an energy function $E'(\vx)$ which represents the product of the densities using the expression
\begin{equation}
    E'(\vx) = E_1(\vx) + E_2(\vx),
    \label{eqn:product_energy_comp}
\end{equation}
providing a way to compose energy functions to implement the product of densities. The product of densities allows us to construct a new density that combines the information in both distributions -- sampling $\vx$ which is highly likely under both distributions~\citep{du2024compositional, yang2023compositional}. The partition functions $Z_1$ and $Z_2$ play no role, since adding a constant to either energy only changes the normalizer of the product, and the product is normalizable whenever one of the densities is bounded, since then $p_1p_2\le\rbr{\sup_\vx p_1(\vx)}\,p_2$. Weighting the energies, $w_1E_1 + w_2E_2$ with $w_1,w_2>0$, gives $p_1^{w_1}p_2^{w_2}$, in which each weight acts as an inverse temperature that sharpens or flattens its component.
    \item \textbf{Mixture of Distributions.} We can represent the mixture $p_{\text{mix}}(\vx) = w_1p_1(\vx) + w_2p_2(\vx)$, with $w_1+w_2 = 1$, by the energy
    \begin{equation}\label{eqn:compose_mixture}
        E_{\text{mix}}(\vx) = -\log\rbr{w_1\frac{\exp\rbr{-E_1(\vx)}}{Z_1} + w_2\frac{\exp\rbr{-E_2(\vx)}}{Z_2}},
    \end{equation}
    which is the soft minimum~\Eqref{eqn:compose_or} with each energy $E_i$ shifted by $\log Z_i - \log w_i$. Unlike the product, the mixture depends on the partition functions. An energy is determined only up to an additive constant, which does not change its own distribution but shifts the weight of its component. For instance, in the energy composition operator in ~\Eqref{eqn:compose_or}. the soft minimum of $E_1$ and $E_2$ alone is the mixture with weights proportional to $Z_1$ and $Z_2$. For general distributions, a mixture with given weights requires $\log Z_1 - \log Z_2$, which can be learned together with the energy, as the constant $c$ of binary NCE~\Eqref{eqn:nce_density}, or estimated with the variational bound~\Eqref{eq:vi_bound}. If the mixture is only sampled, the partition functions can be avoided by ancestral sampling: draw $i$ with probability $w_i$, then draw $\vx\sim p_i$. They are needed once the mixture is composed further, \eg, multiplied by a third distribution, since MCMC on the composed energy evaluates $E_{\text{mix}}$ itself.
    \item \textbf{Inversion of Distributions.} We can represent a distribution that favors points that are likely under $p_1$ but unlikely under $p_2$ by raising $p_2$ to a negative power,
    \begin{equation}\label{eqn:compose_inversion}
        p_{\text{inv}}(\vx) \propto \frac{p_1(\vx)}{p_2(\vx)^{\alpha}}, \qquad E_{\text{inv}}(\vx) = E_1(\vx) - \alpha E_2(\vx),
    \end{equation}
    with $\alpha>0$, which is the distributional counterpart of~\Eqref{eqn:compose_not}. Like the product, the inversion does not depend on the partition functions, but may not be normalizable. For Gaussians $p_i = \Ncal\rbr{0,\sigma_i^2 I}$, $p_{\text{inv}}$ is normalizable only if $\alpha\sigma_1^2 < \sigma_2^2$, \ie, the negated expert must be broad relative to the positive one. Classifier-free guidance ~\Eqref{eqn:cfg_poe}  is an inversion of this kind raised to the power $1+w$, since $p_t(\vx\mid y)^{1+w}/p_t(\vx)^{w} = \rbr{p_t(\vx\mid y)/p_t(\vx)^{\alpha}}^{1+w}$ with $\alpha = w/(1+w) < 1$, and its negated distribution is the unconditional model.
\end{enumerate}

\paragraph{When Is a Composition Correct?} We can understand what a product does through Bayes' rule. Let $c_1,\ldots,c_n$ be conditions on $\vx$, such as attributes of an image or requirements on a trajectory, and suppose we have an unconditional EBM $p(\vx)\propto\exp\rbr{-E(\vx)}$ and a conditional EBM $p(\vx\mid c_i)\propto\exp\rbr{-E(\vx\mid c_i)}$ for each condition alone. If the conditions are conditionally independent given $\vx$, then
\begin{equation}\label{eqn:compose_bayes}
\begin{aligned}
    p(\vx\mid c_1,\ldots,c_n) &\propto p(\vx)\prod_{i=1}^{n}p(c_i\mid\vx) \propto p(\vx)\prod_{i=1}^{n}\frac{p(\vx\mid c_i)}{p(\vx)}, \\
    E(\vx\mid c_1,\ldots,c_n) &= E(\vx) + \sum_{i=1}^{n}\rbr{E(\vx\mid c_i) - E(\vx)},
\end{aligned}
\end{equation}
up to constants that do not depend on $\vx$~\citep{liu2022compositional}. Each difference $E(\vx\mid c_i) - E(\vx)$ equals $-\log p(c_i\mid\vx)$ up to a constant, \ie, it is the energy of an implicit classifier, so the conjunction adds one classifier energy per condition to a single prior, which is counted once rather than $n$ times. Weighting the differences gives the guided distributions of~\Secref{sec:diffusion}, with $n=1$ for classifier-free guidance, and adding the difference for $c_1$ while subtracting the one for $c_2$ gives the negation $p(\vx)\,p(c_1\mid\vx)/p(c_2\mid\vx)$, an inversion~\Eqref{eqn:compose_inversion} whose negated expert is the implicit classifier of $c_2$. The derivation makes two requirements explicit. First, the conditions must be conditionally independent given $\vx$, which holds, for example, when each condition is a function of $\vx$. Second, each $p(\vx\mid c_i)$ must be accurate where the composition places its mass, a region that may contain few training examples for any single condition. The first is a modeling assumption, and the second determines how far a composition generalizes beyond its training data.

\paragraph{What Does Composition Do?} Composition reuses learned components in new combinations. Each component describes one aspect of the data or the task, such as a concept, a constraint, a dynamics model, or a goal, and may be a separately trained model or a conditional model evaluated under one condition. Components may also be defined over different subsets of the variables, in which case the composed energy is an MRF~\Eqref{eq:mrf} whose factors are learned separately and whose graph is chosen for each task, as when the constraints among objects in a scene are composed into a factor graph over their poses~\citep{yang2023compositional}. Since the components are learned separately, training needs to cover each component but not each combination. Suppose $\vx$ is described by $K$ conditions, each taking one of $M$ values. A single model of $p(\vx\mid c_1,\ldots,c_K)$ must be trained on enough of the $M^K$ combinations, whereas the composition~\Eqref{eqn:compose_bayes} only requires the $KM$ conditionals $p(\vx\mid c_k)$, and it generates combinations that never appear together in the training data~\citep{du2020compositional,liu2022compositional}. The same operations recur throughout this tutorial: guided sampling multiplies a diffusion model by a classifier~(\Secref{sec:diffusion}), planners compose dynamics, goals, and obstacles~(\Secref{sec:planning}), and reasoning composes the energies of individual constraints into larger problems~(\Secref{sec:itc_compose}). The product~\Eqref{eqn:product_energy_comp} has the form of a product of experts~\citep{hinton2006training}, which uses it to model a single distribution with jointly trained experts; composition uses it to combine models trained separately. See~\citet{du2024compositional} for a broader discussion.
 
\paragraph{Adaptation.} The same operations adapt a pretrained model to new data or requirements without retraining it. Conditioning is the simplest case: by~\Eqref{eq:ebm_joint} and~\Eqref{eq:ebm_cond}, clamping any subset of the variables of a joint energy, such as the visible pixels of an image or the start and goal of a trajectory, turns the same energy into a conditional model, and MCMC never needs the new partition function. A residual energy adapts a model to new data: by~\Eqref{eq:residual_ebm}, $p(\vx)\propto p_0(\vx)\exp\rbr{-\Etil_\theta(\vx)}$ keeps a pretrained $p_0$ fixed and learns only $\Etil_\theta$, which can be small when the new distribution is close to the old one. This is the structure of the KL-regularized fine-tuning of language models in~\Secref{sec:reg_rl}, where $\Etil = -r/\lambda$, and of the probabilistic adaptation of video models, in which a large pretrained text-to-video model, used only through its score and without fine-tuning, is composed as a product with a small model trained on domain-specific videos~\citep{yang2024probabilistic}. Finally, a new requirement is a new energy term: the planning problems of~\Secref{sec:planning} adapt a model of trajectories to a goal and constraints by adding their energies.
 
\paragraph{Sampling from Compositions.} A composed energy is sampled or minimized with the methods of~\Secref{sec:inference}, which need the composed energy or its gradient but not the partition function. Three practical issues remain. First, reading a sum of energies as a product of distributions requires each energy to be a $\log$-density up to a constant. Energies trained by regression~(\Secref{sec:learning}) or with a margin~(\Secref{sec:disc_model}) are not maximum likelihood fits, and for them the sum is better read as an objective, as in the set operations above: its minimizers satisfy all requirements when the requirements are compatible, and prediction by energy minimization~\Eqref{eq:ebm_prediction_gd} applies directly. Second, separately trained energies have no common scale, so the weights in $\sum_i w_iE_i$ act as inverse temperatures that must be chosen at inference time. Third, the modes of a product lie where all components agree, which can be a small region far from the modes of each component, so the composed landscape is harder to sample than its parts. Annealing through smoothed energies helps, and diffusion models provide such a sequence for each component; however, adding their scores at a noise level does not give the score of the smoothed product, so Langevin or MCMC corrections on the summed energies are needed to sample the composition correctly~(\Secref{sec:diffusion})~\citep{du2023reduce}.

\subsection{Inference with EBMs}\label{sec:inference}

In this section, we first introduce the approximate algorithms for calculating the corresponding probability distributions induced by EBMs, which can be largely categorized in two classes: \textbf{Metropolis–Hastings algorithm} via generating samples and \textbf{variational inference} via distribution approximation. Then, we introduce prediction by optimization, which returns a single output rather than samples. We mainly discuss the inference methods for unconditional EBMs~\Eqref{eq:ebm_joint}, however, the algorithms can be easily generalized for conditional EBMs~\Eqref{eq:ebm_cond}. 

\paragraph{Metropolis-Hastings Sampling.}
Since $Z_\theta$ is intractable, we cannot sample from or calculate the probability of $p_\theta(\vx)\propto\exp\rbr{-E_\theta(\vx)}$ directly. Markov chain Monte Carlo~(MCMC) instead constructs a Markov chain whose stationary distribution is $p_\theta$, and treats the states of the chain after a burn-in period as approximate samples. The Metropolis-Hastings~(MH) algorithm~\citep{metropolis1953equation,hastings1970monte} is the general recipe: given the current state $\vx$, it draws a candidate $\vx'\sim q(\vx'\mid\vx)$ from a proposal distribution and accepts it with probability
\begin{equation}\label{eq:mh_accept}
    \alpha\rbr{\vx, \vx'} = \min\cbr{1, \frac{\exp\rbr{-E_\theta(\vx')}\, q(\vx\mid\vx')}{\exp\rbr{-E_\theta(\vx)}\, q(\vx'\mid\vx)}},
\end{equation}
otherwise staying at $\vx$. The partition function cancels in the ratio, so only energy differences $E_\theta(\vx') - E_\theta(\vx)$ are required, and the acceptance step guarantees detailed balance with respect to $p_\theta$ for any valid proposal. Different MH samplers differ only in \textbf{the choice of $q$}, which determines how the energy function is used.

The Gibbs sampler~\citep{geman1984stochastic} updates one variable (or block of variables) at a time, using the full conditional $p_\theta(x_i\mid\vx_{-i})$ as the proposal. With this choice of $q$, the acceptance probability in~\Eqref{eq:mh_accept} is always one, which means the candidate is always accepted. The conditionals only involve the energy terms that contain $x_i$, which makes Gibbs sampling natural for MRFs and Boltzmann machines. Particularly, in RBMs, the conditional independence of $\vy$ given $\vx$ and vice versa allows each layer to be sampled in one block.

For continuous $\vx$ and differentiable energies, the gradient $\nabla_\vx E_\theta(\vx)$ provides a much better proposal than random perturbations. Langevin dynamics follows the gradient with injected noise,
\begin{equation}\label{eq:langevin}
    \vx^{(t+1)} = \vx^{(t)} - \eta\nabla_\vx E_\theta\rbr{\vx^{(t)}} + \sqrt{2\eta}\,\bm{\xi}^{(t)}, \quad \bm{\xi}^{(t)}\sim\Ncal\rbr{0, I},
\end{equation}
which converges to $p_\theta$ as $\eta\to 0$. With a finite step size, the update is biased; using~\Eqref{eq:langevin} as the proposal of an MH step removes the bias, yielding the Metropolis-adjusted Langevin algorithm~(MALA)~\citep{roberts1996exponential}. In practice, many EBM learning methods run the unadjusted version for a fixed number of steps.

Langevin dynamics moves by a small random step at each iteration, so it explores the space slowly, like a random walk. Hamiltonian Monte Carlo~(HMC)~\citep{duane1987hybrid,neal2011mcmc} avoids this by borrowing the dynamics of a physical particle. It introduces an auxiliary momentum $\vv\in\RR^d$ and defines the Hamiltonian
\begin{equation}\label{eq:hamiltonian}
    H(\vx, \vv) = E_\theta(\vx) + \frac{1}{2}\|\vv\|_2^2,
\end{equation}
\ie, a potential energy $E_\theta(\vx)$ plus a kinetic energy. The joint distribution $p(\vx, \vv)\propto\exp\rbr{-H(\vx, \vv)} = p_\theta(\vx)\,\Ncal\rbr{\vv; 0, I}$ has $p_\theta$ as its marginal on $\vx$, so it suffices to sample from the joint and discard $\vv$. Each HMC iteration first resamples the momentum $\vv\sim\Ncal\rbr{0, I}$, and then moves $(\vx, \vv)$ along Hamilton's equations,
\begin{equation}\label{eq:hamilton_eq}
    \frac{d\vx}{dt} = \vv, \qquad \frac{d\vv}{dt} = -\nabla_\vx E_\theta(\vx).
\end{equation}
The particle speeds up when rolling down the energy landscape and slows down when climbing, while the total energy $H$ stays constant. Hence it can travel far from its starting point to a state of similar probability. Since~\Eqref{eq:hamilton_eq} cannot be solved exactly, it is simulated for $L$ steps of the leapfrog integrator with step size $\epsilon$,
\begin{equation}\label{eq:leapfrog}
    \vv \leftarrow \vv - \frac{\epsilon}{2}\nabla_\vx E_\theta(\vx), \qquad
    \vx \leftarrow \vx + \epsilon\vv, \qquad
    \vv \leftarrow \vv - \frac{\epsilon}{2}\nabla_\vx E_\theta(\vx).
\end{equation}
The leapfrog integrator is reversible and preserves volume, so the end point $(\vx', \vv')$ can be used as an MH proposal with a simple acceptance probability $\min\cbr{1, \exp\rbr{H(\vx, \vv) - H(\vx', \vv')}}$, which corrects the discretization error. Because the leapfrog integrator nearly conserves $H$, this probability stays close to one even for distant proposals. HMC thus trades more gradient evaluations per sample ($L$ per iteration) for much less correlated samples. Its performance depends on $\epsilon$ and $L$: a large $\epsilon$ lowers the acceptance rate, while a poorly chosen $L$ wastes computation or makes the trajectory turn back on itself. The No-U-Turn sampler~\citep{hoffman2014no} balances this trade-off by setting $L$ adaptively. 
With a single leapfrog step ($L=1$), HMC reduces to MALA with step size $\eta = \epsilon^2/2$.

The exact samplers above, general MH, Gibbs sampling, MALA, and HMC, are all special cases of the MH algorithm; unadjusted Langevin dynamics is not, and does not sample exactly from $p_\theta$ at a finite step size. These samplers place different requirements on the energy function: general MH and Gibbs sampling only need to evaluate energies (or conditionals), and therefore apply to discrete and structured spaces, whereas Langevin dynamics and HMC need the gradient $\nabla_\vx E_\theta$, and therefore require continuous inputs and differentiable energies.

\paragraph{Amortized Sampling via Variational Inference. }
MCMC samplers pay their cost for every sample: each new sample requires running a chain for many steps, and consecutive samples are correlated. Variational inference takes a different route: it fits a tractable distribution $q_\phi(\vx)$ to approximate EBM $p_\theta$, so that afterwards drawing a sample costs only one pass through $q_\phi$; the cost of inference is thus \emph{amortized} over all samples. The standard objective is minimizing the reverse Kullback-Leibler divergence,
\begin{equation}\label{eq:vi_kl}
    \min_{q_\phi}\,\,\mathrm{KL}\rbr{q_\phi\|p_\theta} = \EE_{\vx\sim q_\phi}\sbr{E_\theta(\vx)} - H\rbr{q_\phi} + \log Z_\theta,
\end{equation}
where $H\rbr{q_\phi} = -\EE_{q_\phi}\sbr{\log q_\phi(\vx)}$ is the entropy of $q_\phi$. Since $\log Z_\theta$ does not depend on $\phi$, we can fit $q_\phi$ by minimizing $\EE_{q_\phi}\sbr{E_\theta(\vx)} - H\rbr{q_\phi}$, which never evaluates the intractable partition function. The first term pulls the samples toward low-energy regions, and the entropy term keeps them spread out instead of collapsing onto a single minimum. Because the KL divergence is non-negative,~\Eqref{eq:vi_kl} also yields the variational lower bound
\begin{equation}\label{eq:vi_bound}
    \log Z_\theta \ge \EE_{\vx\sim q_\phi}\sbr{-E_\theta(\vx)} + H\rbr{q_\phi},
\end{equation}
which is tight if and only if $q_\phi = p_\theta$. Therefore, the same optimization provides both a sampler and an estimate of the $\log$-partition function, which can also be used for learning EBMs.

The only remaining part is that the parametrization choice of $q_\phi$ needs to make both sampling and the entropy tractable. The classical choice for MRFs is the mean-field approximation $q_\phi(\vx) = \prod_i q_{\phi_i}(x_i)$, which ignores the dependencies among variables~\citep{wainwright2008graphical}. Modern choices are neural samplers: normalizing flows~\citep{rezende2015variational} and autoregressive models give exact $\log$ densities, and hence exact entropies, variational auto-encoders~\citep{kingma2014auto} give a lower bound on the $\log$ density, and implicit generators $\vx = g_\phi(\vz)$ with $\vz\sim\Ncal\rbr{0, I}$ are more flexible but require estimating the entropy~\citep{kim2016deep,dai2019kernel,dai2019exponential}. 

To optimize the variational objective~\Eqref{eq:vi_kl}, we can exploit the stochastic gradient descent method, which requires estimating the gradient w.r.t. $\phi$. The difficulty is that $\phi$ appears in the distribution the expectation is taken over. There are two standard estimators. For continuous $\vx$ and an implicit or reparameterizable $q_\phi$, \ie, $\vx = g_\phi(\vz)$ with $\vz\sim\Ncal\rbr{0, I}$, the reparameterization trick~\citep{kingma2014auto} moves $\phi$ inside the expectation,
\begin{equation}\label{eq:reparam_grad}
    \nabla_\phi\EE_{\vz}\sbr{E_\theta\rbr{g_\phi(\vz)}} = \EE_{\vz}\sbr{\rbr{\frac{\partial g_\phi(\vz)}{\partial\phi}}^\top\nabla_\vx E_\theta\rbr{g_\phi(\vz)}},
\end{equation}
which again uses $\nabla_\vx E_\theta$. For discrete $\vx$, or whenever $\nabla_\vx E_\theta$ is unavailable, we use the REINFORCE, or score-function, estimator~\citep{williams1992simple}. It relies on the identity $\nabla_\phi q_\phi(\vx) = q_\phi(\vx)\nabla_\phi\log q_\phi(\vx)$, which gives
\begin{equation}\label{eq:reinforce_grad}
    \nabla_\phi\EE_{\vx\sim q_\phi}\sbr{E_\theta(\vx) + \log q_\phi(\vx)} = \EE_{\vx\sim q_\phi}\sbr{\rbr{E_\theta(\vx) + \log q_\phi(\vx) - b}\nabla_\phi\log q_\phi(\vx)},
\end{equation}
where the left-hand side is exactly the variational objective, \ie, $\EE_{q_\phi}\sbr{E_\theta(\vx)} - H\rbr{q_\phi}$. The derivative of $\log q_\phi$ inside the expectation drops out because $\EE_{q_\phi}\sbr{\nabla_\phi\log q_\phi(\vx)} = 0$; for the same reason, any constant baseline $b$ can be subtracted without introducing bias. REINFORCE only needs to evaluate the energy and the $\log$-density of $q_\phi$ at the samples, so it applies to any $\vx$, including discrete and structured outputs. Its price is high variance, as it only uses the value of the energy rather than its gradient. Therefore, in practice, a baseline $b$, \eg, the average of $E_\theta(\vx) + \log q_\phi(\vx)$ over the other samples in a batch, or other control variates are necessary to make it work. In short, the two estimators trade generality for variance: the reparameterization estimator~\Eqref{eq:reparam_grad} uses the gradient of the energy and usually has much lower variance, but requires continuous $\vx$, a differentiable $E_\theta$, and a reparameterizable $q_\phi$; REINFORCE~\Eqref{eq:reinforce_grad} only needs energy values and $\log q_\phi$, and so applies to discrete or non-differentiable settings, at the cost of higher variance. When both are applicable, the reparameterization estimator is generally preferred.

Compared with MCMC, variational inference is fast at sampling time but biased: the error is limited by the family of $q_\phi$, and the reverse KL~\Eqref{eq:vi_kl} is mode-seeking, so $q_\phi$ tends to miss some modes of $p_\theta$. The two can be combined~\citep{dai2019exponential}: a learned $q_\phi$ can provide initializations for short MCMC chains, or serve as the proposal in an MH step, whose acceptance probability~\Eqref{eq:mh_accept} then corrects the bias of $q_\phi$.

\paragraph{Prediction.}
In addition to drawing samples, we can use an EBM to make a prediction given an input $\vx$. For a conditional model
\begin{equation}
    p_\theta(\vy\mid\vx)
    = \frac{\exp(-E_\theta(\vx,\vy))}{Z_\theta(\vx)},
\end{equation}
a maximum a posteriori (MAP) prediction is
\begin{equation}\label{eq:ebm_map_prediction}
    \hat{\vy}_{\mathrm{MAP}} \in
    \operatorname*{arg\,max}_{\vy} p_\theta(\vy\mid\vx)
    = \operatorname*{arg\,min}_{\vy} E_\theta(\vx,\vy).
\end{equation}
The partition function depends on $\vx$ but not on the candidate output $\vy$, so it is not needed to compare predictions. More generally, energy minimization defines a prediction rule even when the energy does not induce a normalizable probability distribution, provided a minimum exists. The probabilistic interpretation additionally requires $0<Z_\theta(\vx)<\infty$.

The computation required to find a prediction depends on the energy and the output space. For a small set of discrete labels, we can evaluate every candidate and select the one with the lowest energy. For continuous outputs and differentiable energies, we can iteratively update a candidate through gradient descent,
\begin{equation}\label{eq:ebm_prediction_gd}
    \vy^{(t+1)} = \vy^{(t)}
    - \eta_t\nabla_{\vy}E_\theta(\vx,\vy^{(t)}),
\end{equation}
where $\eta_t$ is a step size. For structured discrete outputs, we instead need to use combinatorial search or another approximate minimization method. Overall, for general nonconvex energies, these minimization procedures may not find a global minimum, making initialization significantly affect the final prediction.

Standard feedforward prediction is a special case in which the minimum is available directly. Consider the quadratic energy
\begin{equation}\label{eq:regression_energy}
    E_\theta(\vx,\vy)
    = \frac{1}{2\sigma^2}\|\vy-f_\theta(\vx)\|_2^2,
    \qquad \sigma>0.
\end{equation}
Its unique minimizer is $\hat{\vy}=f_\theta(\vx)$, which can be computed with one network evaluation. In contrast, a general energy $E_\theta(\vx,\vy)$ does not need to have this quadratic form and can express multiple complex dependencies among output variables.

\subsection{Learning for EBMs}\label{sec:learning}

Learning an EBM means fitting the energy $E_\theta$ so that $p_\theta(\vx)\propto\exp\rbr{-E_\theta(\vx)}$ matches the data distribution. The obstacle is again the partition function: $Z_\theta$ depends on $\theta$, so the likelihood and its gradient cannot be computed directly. Learning methods for EBMs differ mainly in how they avoid or approximate $Z_\theta$, as we will discuss in the following.  

\paragraph{Maximum Likelihood Estimation.} Given a dataset $\mathcal{D} = \{X\}$, the most direct approach to learning an energy function $E_\theta(\vx)$ is through maximum likelihood estimation~(MLE). We can directly fit a probability density $p_\theta(\vx) \propto \exp (-E_\theta(\vx))$ to maximize the $\log$-likelihood of samples drawn from a dataset $\mathcal{D}$, \ie, minimizing the negative $\log$-likelihood.

The full form of the likelihood of $E_\theta(\vx)$ as defined in~\Eqref{eq:ebm} is $\exp(-E_\theta(\vx)) / Z$, where $Z$ is the partition function. The negative $\log$-likelihood loss objective is given by
\begin{equation}\label{eq:mle_ebm}
    \mathcal{L}_{NLL}(\mathcal{D}) = \EE_{\vx \sim \mathcal{D}}[-\log (p_\theta(\vx))],
\end{equation}
which can be equivalently written as
\begin{equation}
    \mathcal{L}_{NLL}(\mathcal{D}) = \EE_{\vx \sim \mathcal{D}}[E_\theta(\vx)] + \log \left (\int \exp(-E_\theta(\vx)) d\vx \right).
    \label{eqn:nll_ebm}
\end{equation}
Taking the gradient of the above objective, we obtain the expression
\begin{equation}
    \nabla_\theta \mathcal{L}_{NLL}(\mathcal{D}) = \EE_{\vx \sim \mathcal{D}}[\nabla_\theta E_\theta(\vx)] - \frac{\int \exp (-E_\theta(\vx)) \nabla_\theta E_\theta(\vx) d\vx}{\int \exp (-E_\theta(\vx)) d\vx},
    \label{eqn:gradient_ebm}
\end{equation}
where we can reduce the right-hand side to the expectation
\begin{equation}
    \nabla_\theta \mathcal{L}_{NLL}(\mathcal{D}) = \EE_{\vx \sim \mathcal{D}}[\nabla_\theta E_\theta(\vx)] - \EE_{\vx \sim p_\theta(\vx)}[\nabla_\theta E_\theta(\vx)].
    \label{eqn:gradient_ebm_reduce}
\end{equation}
Using the above expression, the gradient of $\mathcal{L}_{NLL}$ can be computed equivalently from the surrogate loss
\begin{equation}
    \widetilde{\mathcal{L}}_{NLL}(\mathcal{D}) = \EE_{\vx \sim \mathcal{D}}[E_\theta(\vx)] - \EE_{\vx \sim \texttt{stopgrad}\rbr{p_\theta(\vx)}}[ E_\theta(\vx)],
    \label{eqn:loss_ebm_reduce}
\end{equation}
which gives us a simple objective for training $E_\theta(\vx)$ through maximum likelihood estimation, where we contrastively decrease the energy of ground truth data samples while increasing the energy of samples drawn from the model distribution $\vx \sim p_\theta(\vx)$. However, in practice, drawing samples from the distribution $p_\theta(\vx)$ can be challenging -- several approaches to drawing these samples were discussed in~\Secref{sec:inference}.

Particularly, in the original contrastive divergence, these negative samples are initialized from ground truth data~\citep{hinton2006training}, while one can also initialize samples from random noise or an approximate distribution and use Langevin dynamics to draw negative samples to train the EBM~\citep{du2019implicit, dai2019exponential}. Instead of sampling from $p_\theta$, one can approximate~\Eqref{eqn:gradient_ebm} by amortizing its denominator, \ie, by learning $\log Z_\theta$ jointly with the energy. For an unconditional model, $\log Z_\theta$ is a single scalar, which can be fitted as one extra parameter $\tau$. For a conditional EBM $p_\theta(\vy\mid\vx)$, the partition function $Z_\theta(\vx)$ changes with every input, and a second network $\tau(\vx)$ approximates $\log Z_\theta(\vx)$ variationally~\citep{sander2025joint,sun2025amorlip}. Concretely, \citet{sander2025joint} start from the convex conjugate of the logarithm, $\log Z = \min_\tau\cbr{\tau + Z e^{-\tau} - 1}$; with $Z_\theta(\vx) = \EE_{q(\vy\mid\vx)}\sbr{\exp\rbr{-E_\theta(\vx,\vy)}}$ for a base distribution $q$, this replaces $\log Z_\theta(\vx)$ in the negative $\log$-likelihood by $\tau(\vx) + \EE_{q(\vy\mid\vx)}\sbr{\exp\rbr{-E_\theta(\vx,\vy) - \tau(\vx)}} - 1$, minimized jointly over $\theta$ and $\tau$. Writing $\tilde p_\tau(\vy\mid\vx) = q(\vy\mid\vx)\exp\rbr{-E_\theta(\vx,\vy) - \tau(\vx)}$ for the amortized conditional, which is unnormalized unless $\tau(\vx) = \log Z_\theta(\vx)$, this objective equals, up to a constant, the generalized KL divergence $\mathrm{KL}\rbr{p_\theta\|\tilde p_\tau}$. Since the expectation enters linearly, this objective can be estimated without bias from samples of $q$, but the exponential makes it numerically fragile: when the energy varies widely, $\exp\rbr{-E_\theta - \tau}$ can overflow and its estimate has high variance. AmorLIP~\citep{sun2025amorlip} instead minimizes the KL divergence in the other direction, $\mathrm{KL}\rbr{\tilde p_\tau\|p_\theta}$, and, to address the instability, further proposes its second-order approximation, a regression of $\tau(\vx)$ onto $\log Z_\theta(\vx)$ in log space, which trades a small bias for a numerically stable objective with lower variance.

\paragraph{Noise Contrastive Estimation.}
Maximum likelihood needs samples from the model $p_\theta$ in every gradient step, which is what makes it expensive. Noise contrastive estimation~(NCE)~\citep{gutmann2010noise} avoids this by recasting the learning problem as classification between data and noise, from which the energy function can be recovered. Concretely, by formulating binary classification and multi-class classification problems, we obtain two variants of NCE.

\textit{Binary NCE.} The original formulation~\citep{gutmann2010noise} poses a binary classification between data and noise. It treats the $\log$-partition function as an extra scalar parameter $c$, so that the model's normalized $\log$-density is
\begin{equation}\label{eqn:nce_density}
    \log p_{\theta,c}(\vx) = -E_\theta(\vx) - c.
\end{equation}
Suppose each data point comes with $\nu$ noise samples, so that a sample is a priori $\nu$ times more likely to be noise. By Bayes' rule, the probability that a sample $\vx$ is data rather than noise is
\begin{equation}\label{eqn:nce_posterior}
    P(y=1\mid\vx) = \frac{p_{\theta,c}(\vx)}{p_{\theta,c}(\vx) + \nu\,p_n(\vx)} = \sigmoid\rbr{h_{\theta,c}(\vx)}, \qquad
    h_{\theta,c}(\vx) = \underbrace{-E_\theta(\vx) - \log p_n(\vx)}_{\text{unnormalized $\log$-ratio to noise}} + \underbrace{\rbr{-c - \log\nu}}_{\text{bias } b},
\end{equation}
where $\sigmoid$ is the logistic function. The posterior is thus a sigmoid of the model's $\log$-density ratio to the noise, plus a bias term $b = -c - \log\nu$, which implies we can recover the EBM parameters from the classifier $P(y=1\mid\vx)$.
We obtain the Binary NCE, which fits $(\theta, c)$ by logistic regression, \ie, minimizing the cross-entropy,
\begin{equation}\label{eqn:lnce_bin}
    \mathcal{L}_{\text{NCE}}^{\text{bin}}(\theta, c) = -\EE_{\vx\sim p_d}\sbr{\log\sigmoid\rbr{h_{\theta,c}(\vx)}} - \nu\,\EE_{\vx\sim p_n}\sbr{\log\rbr{1-\sigmoid\rbr{h_{\theta,c}(\vx)}}},
\end{equation}
with the gradient with respect to $\theta$ as
\begin{equation}\label{eqn:nce_bin_grad}
    \nabla_\theta\mathcal{L}_{\text{NCE}}^{\text{bin}} = \EE_{\vx\sim p_d}\sbr{\rbr{1-\sigmoid\rbr{h_{\theta,c}(\vx)}}\nabla_\theta E_\theta(\vx)} - \nu\,\EE_{\vx\sim p_n}\sbr{\sigmoid\rbr{h_{\theta,c}(\vx)}\nabla_\theta E_\theta(\vx)}.
\end{equation}
We emphasize that the negative samples in~\Eqref{eqn:nce_bin_grad} now come from $p_n$ instead of $p_\theta$, so no MCMC is needed. If the model family contains $p_d$, the minimizer satisfies $p_{\theta,c} = p_d$, so the estimator is consistent~\citep{gutmann2010noise}, and the learned constant recovers the $\log$-partition function, $c^\star = \log Z_\theta = \log\int\exp\rbr{-E_\theta(\vx)}d\vx$. Binary NCE therefore learns both the energy and its normalizer, which allows the likelihood of new samples to be evaluated afterwards.

\textit{Ranking NCE.} The second form poses a multi-class problem instead~\citep{ma2018noise}: given one data point $\vx_0\sim p_d$ hidden among $K$ noise samples $\vx_1,\ldots,\vx_K\sim p_n$, the model must identify which of the $K+1$ samples is the real one. To see what this probability is, let $i\in\{0,\ldots,K\}$ be the unknown position of the data point, with a uniform prior. If the data were drawn from the model $p_\theta$, the $K+1$ samples would have joint density $p_\theta(\vx_i)\prod_{j\neq i}p_n(\vx_j)$ given $i$. By Bayes' rule, the posterior over the position is
\begin{equation}\label{eqn:nce_rank_posterior}
    P\rbr{i\mid\vx_{0:K}}
    = \frac{p_\theta(\vx_i)\prod_{j\neq i}p_n(\vx_j)}{\sum_{l=0}^{K}p_\theta(\vx_l)\prod_{j\neq l}p_n(\vx_j)}
    = \frac{p_\theta(\vx_i)/p_n(\vx_i)}{\sum_{l=0}^{K}p_\theta(\vx_l)/p_n(\vx_l)}
    = \frac{\exp\rbr{-E_\theta(\vx_i)}/p_n(\vx_i)}{\sum_{l=0}^{K}\exp\rbr{-E_\theta(\vx_l)}/p_n(\vx_l)},
\end{equation}
where the second equality divides the numerator and the denominator by $\prod_{j=0}^{K}p_n(\vx_j)$, and the third substitutes $p_\theta = \exp\rbr{-E_\theta}/Z_\theta$, so that $Z_\theta$ cancels. This is a softmax over the samples with logits $-E_\theta(\vx_k) - \log p_n(\vx_k)$. Ranking NCE maximizes the probability of identifying the data point, $i=0$, \ie, it minimizes the negative $\log$-probability,
\begin{equation}\label{eqn:lnce_rank}
    \mathcal{L}_{\text{NCE}}^{\text{rank}}(\theta) = -\EE\sbr{\log\frac{\exp\rbr{-E_\theta(\vx_0)}/p_n(\vx_0)}{\sum_{k=0}^{K}\exp\rbr{-E_\theta(\vx_k)}/p_n(\vx_k)}}.
\end{equation}
Since any constant in the energy appears in both the numerator and the denominator, the partition function cancels: ranking NCE learns $E_\theta$ alone, with no parameter $c$. Its gradient,
\begin{equation}\label{eqn:nce_rank_grad}
    \nabla_\theta\mathcal{L}_{\text{NCE}}^{\text{rank}} = \EE\sbr{\nabla_\theta E_\theta(\vx_0) - \sum_{k=0}^{K} w_k\,\nabla_\theta E_\theta(\vx_k)}, \qquad
    w_k = \frac{\exp\rbr{-E_\theta(\vx_k)}/p_n(\vx_k)}{\sum_{j=0}^{K}\exp\rbr{-E_\theta(\vx_j)}/p_n(\vx_j)},
\end{equation}
is again the maximum likelihood gradient, with the expectation under $p_\theta$ replaced by a self-normalized importance sampling estimate that uses $p_n$ as the proposal. Ranking NCE is consistent as well, and it is especially convenient for conditional models $p_\theta(\vy\mid\vx)$, where the partition function $Z_\theta(\vx)$ differs for every input and binary NCE would need a separate constant for each $\vx$~\citep{ma2018noise}. The price is that it does not recover $Z_\theta$, which has to be estimated separately, \eg, by importance sampling, if normalized likelihoods are needed.

\textit{Choosing the noise.} Both forms depend heavily on the noise distribution. $p_n$ must be positive wherever $p_d$ is, and it should be close to $p_d$: if the noise is easy to tell apart from the data, the classifier succeeds without learning much about the data, and the gradients carry little information. Larger $\nu$ or $K$ makes both estimators closer to maximum likelihood, at the cost of more noise samples per data point.

\paragraph{Pseudolikelihood Estimation.}
Given an unnormalized EBM $E_\theta(\vx)$  over variables $\vx = \cbr{x_1, x_2, \ldots, x_d}$, in pseudolikelihood estimation~\citep{besag1975statistical}, we define the pseudolikelihood as
\begin{equation}
    PL(\theta; \mathbf{x}) = \prod_{j=1}^{d} p_\theta(x_j \mid \mathbf{x}_{-j}).
\end{equation}
The objective we train our estimator with is then the negative $\log$-pseudolikelihood
\begin{equation}
\mathcal{L}_{PL}(\theta) = - \sum_{i=1}^{N} \sum_{j=1}^{d} \log p_\theta(x_j^{(i)} \mid \mathbf{x}_{-j}^{(i)}).
\label{eqn:lpseudo}
\end{equation}
The objective in \Eqref{eqn:lpseudo} is substantially easier to optimize than \Eqref{eq:mle_ebm} since we can compute $p_\theta(x_j^{(i)} \mid \mathbf{x}_{-j}^{(i)})$ in closed form as
$$p(x_j = k \mid \mathbf{x}_{-j}) = \frac{p(x_j=k, \mathbf{x}_{-j})}{\sum_{v \in \mathcal{V}} p(x_j=v, \mathbf{x}_{-j})} = \frac{\exp(-E(x_j=k, \mathbf{x}_{-j}))}{\sum_{v \in \mathcal{V}} \exp(-E(x_j=v, \mathbf{x}_{-j}))},$$
which can be computed by enumerating through the vocabulary size $\mathcal{V}$ of $x_j$.

In general, pseudolikelihood is a consistent estimator of the overall distribution $p(\vx)$, as long as the distribution is positive at every data point $\vx$~\citep{besag1975statistical,hyvarinen2006consistency}. However, in comparison to other approaches for training EBMs, pseudolikelihood focuses on local interactions between variables. This can make the method data inefficient when fitting distributions where there are global interactions between multiple variables.

\textit{Connection to masked discrete diffusion.} Pseudolikelihood is closely related to masked, or absorbing-state, discrete diffusion models~\citep{austin2021structured,sahoo2024simple,shi2024simplified}. Both are trained to predict hidden variables from visible ones. Pseudolikelihood hides exactly one variable $x_j$ and fits $p_\theta(x_j\mid\vx_{-j})$. Masked diffusion hides a random subset $M$ of the variables, whose size ranges from one variable to all of them, and fits $p_\theta(x_j\mid\vx_{-M})$ for every $j\in M$. With the linear noise schedule, its training objective is a weighted average of such masked-prediction losses over masking ratios,
\begin{equation}\label{eqn:masked_diffusion}
    \mathcal{L}_{\text{MD}}(\theta) = \EE_{t\sim\mathcal{U}[0,1]}\sbr{\frac{1}{t}\,\EE_{M\sim q_t}\sbr{-\sum_{j\in M}\log p_\theta\rbr{x_j\mid\vx_{-M}}}},
\end{equation}
where $q_t$ hides each variable independently with probability $t$. Pseudolikelihood~\Eqref{eqn:lpseudo} is the special case in which exactly one variable is hidden at a time. The link is even closer in continuous time: the score of a discrete diffusion model can be written through the single-variable conditionals $p_t(x_j\mid\vx_{-j})$, exactly the quantities pseudolikelihood fits, and it can be learned by a categorical ratio matching objective built on them~\citep{sun2023score}.

The difference matters for what each objective learns. By the chain rule, for any ordering $\pi$ of the variables, $\log p(\vx) = \sum_{k=1}^{d}\log p\rbr{x_{\pi(k)}\mid\vx_{\pi(<k)}}$. The masked diffusion objective~\Eqref{eqn:masked_diffusion} is equivalent to averaging this decomposition over random orderings~\citep{uria2014deep,hoogeboom2022autoregressive,ou2024your}, so it is an upper bound on the negative $\log$-likelihood. Pseudolikelihood keeps only the last term of each ordering, $k=d$, where every other variable is visible. It is therefore not a bound on the likelihood, and it never trains the model to predict a variable whose neighbors are also unknown, which is why it struggles with the global interactions mentioned above. In this sense, masked diffusion extends pseudolikelihood from one masking level to all of them.

The two also differ in another respect. In an EBM, every conditional is derived from a single energy function, so the conditionals are always consistent with one joint distribution. Masked diffusion models usually parameterize the conditionals directly with a neural network, and the learned conditionals need not agree with any joint distribution. The two also sample differently: Gibbs sampling resamples one variable at a time from the pseudolikelihood conditionals and keeps cycling, whereas masked diffusion starts with every variable hidden and reveals them progressively until none are left.

\paragraph{Score Matching.}
Score matching~\citep{hyvarinen2005estimation} avoids the partition function by fitting the gradient of the $\log$-density instead of the density itself. For continuous $\vx$, this gradient is the \emph{score},
\begin{equation}\label{eqn:ebm_score}
    \nabla_\vx\log p_\theta(\vx) = -\nabla_\vx E_\theta(\vx) - \nabla_\vx\log Z_\theta = -\nabla_\vx E_\theta(\vx),
\end{equation}
and $Z_\theta$ drops out because it does not depend on $\vx$. Score matching fits the model score to the data score by minimizing their expected squared distance, the Fisher divergence,
\begin{equation}\label{eqn:esm}
    \mathcal{L}_{\text{ESM}}(\theta) = \frac{1}{2}\,\EE_{\vx\sim p_d}\sbr{\big\|\nabla_\vx\log p_\theta(\vx) - \nabla_\vx\log p_d(\vx)\big\|_2^2}.
\end{equation}
The data score $\nabla_\vx\log p_d$ is unknown, but \citet{hyvarinen2005estimation} showed by integration by parts that, under mild smoothness and boundary conditions, \Eqref{eqn:esm} equals, up to a constant that does not depend on $\theta$,
\begin{equation}\label{eqn:ism}
    \mathcal{L}_{\text{SM}}(\theta) = \EE_{\vx\sim p_d}\sbr{\frac{1}{2}\big\|\nabla_\vx E_\theta(\vx)\big\|_2^2 - \tr\rbr{\nabla_\vx^2 E_\theta(\vx)}},
\end{equation}
which only involves the energy and the data. Like NCE, score matching needs no samples from the model; unlike NCE, it needs no noise distribution either. It is consistent when the model family contains $p_d$~\citep{hyvarinen2005estimation}.

\textit{Scalable variants.} The trace of the Hessian in~\Eqref{eqn:ism} needs $d$ backward passes, which is expensive in high dimensions. Sliced score matching~\citep{song2019sliced} projects the scores onto random directions $\vv$ and replaces the trace with $\vv^\top\nabla_\vx^2 E_\theta(\vx)\,\vv$, which needs a single Hessian-vector product. Denoising score matching~\citep{vincent2011connection} removes the second derivative altogether. It adds Gaussian noise to the data, $\tilde\vx = \vx + \sigma\bm{\epsilon}$ with $\bm{\epsilon}\sim\Ncal\rbr{0, I}$, and matches the model score at $\tilde\vx$ to the score of the noising kernel, which is known in closed form,
\begin{equation}\label{eqn:dsm}
    \mathcal{L}_{\text{DSM}}(\theta) = \frac{1}{2}\,\EE_{\vx\sim p_d,\,\bm{\epsilon}}\sbr{\Big\|-\nabla_{\tilde\vx}E_\theta(\tilde\vx) + \frac{\tilde\vx - \vx}{\sigma^2}\Big\|_2^2}.
\end{equation}
Its minimizer is the score of the noisy data distribution rather than of $p_d$ itself, which approaches the data score as $\sigma\to0$. Training on a range of noise levels, from large to small, gives the noise-conditional score models behind score-based diffusion models~\citep{song2019generative,song2021score}, which can be read as EBMs whose energy is learned at every noise level and sampled with annealed Langevin dynamics~\Eqref{eq:langevin}.

\textit{Limitations.} Because the score is a local quantity, score matching sees each region of the space only through how the density changes there. Where the data are scarce, as in the low-density regions between separated modes, the learned score is inaccurate, and Langevin sampling from it mixes poorly between modes~\citep{song2019generative}; the noise in denoising score matching helps by filling in these regions. Score matching also requires continuous $\vx$ and a differentiable energy. For discrete data, ratio matching~\citep{hyvarinen2007some} plays the same role by matching ratios $p_\theta(\vx')/p_\theta(\vx)$ between states that differ in a single variable, in which $Z_\theta$ again cancels. These ratios are determined by the single-variable conditionals $p_\theta(x_j\mid\vx_{-j})$, which ties score matching for discrete data back to pseudolikelihood and to the categorical ratio matching used in discrete diffusion~\citep{sun2023score}.

\paragraph{Adversarial Training.}
Adversarial training starts from maximum likelihood and removes the partition function with the variational bound from~\Secref{sec:inference}. The negative $\log$-likelihood~\Eqref{eqn:nll_ebm} contains $\log Z_\theta$, and by~\Eqref{eq:vi_bound}, $\log Z_\theta$ is itself the solution of an optimization problem over distributions,
\begin{equation}\label{eqn:logz_variational}
    \log Z_\theta = \max_{q}\;\EE_{\vx\sim q}\sbr{-E_\theta(\vx)} + H\rbr{q},
\end{equation}
where the maximum is attained at $q = p_\theta$. Substituting~\Eqref{eqn:logz_variational} into~\Eqref{eqn:nll_ebm} turns maximum likelihood into a min-max problem,
\begin{equation}\label{eqn:adv_minmax}
    \min_\theta\;\mathcal{L}_{NLL}(\theta) = \min_\theta\,\max_{q}\;\EE_{\vx\sim p_d}\sbr{E_\theta(\vx)} - \EE_{\vx\sim q}\sbr{E_\theta(\vx)} + H\rbr{q}.
\end{equation}
The two players have clear roles. For a fixed energy, the inner maximization over $q$ is exactly the variational inference problem~\Eqref{eq:vi_kl}: it fits $q$ to $p_\theta$ by minimizing $\mathrm{KL}\rbr{q\|p_\theta}$, so $q$ becomes an amortized sampler for the current EBM. For a fixed $q$, the gradient of the outer objective with respect to $\theta$ is
\begin{equation}\label{eqn:adv_grad}
    \EE_{\vx\sim p_d}\sbr{\nabla_\theta E_\theta(\vx)} - \EE_{\vx\sim q}\sbr{\nabla_\theta E_\theta(\vx)},
\end{equation}
which is the maximum likelihood gradient~\Eqref{eqn:gradient_ebm_reduce} with the samples from $p_\theta$ replaced by samples from $q$. The energy thus lowers itself on data and raises itself on the sampler's outputs, while the sampler moves its samples toward low energy and keeps them spread out through the entropy term. No MCMC is run during training; the sampler is learned instead.

This is the structure of a generative adversarial network~\citep{goodfellow2014generative}: the energy plays the discriminator and $q$ plays the generator. The difference is the entropy term $H\rbr{q}$. Without it, the inner maximization would put all of $q$'s mass at the minimum of the energy, which is the mode collapse familiar from GANs, and the game would no longer correspond to maximum likelihood. With it, the solution of~\Eqref{eqn:adv_minmax} is the maximum likelihood estimate, and the learned generator is a sampler for the learned EBM~\citep{kim2016deep,dai2017calibrating}.

The entropy is also what makes the method hard in practice. It is easy to compute only for restricted families of $q$, such as normalizing flows~\citep{rezende2015variational}. For flexible implicit generators $\vx = g_\phi(\vz)$, it has to be estimated or bounded, and different estimates lead to different algorithms~\citep{kim2016deep,dai2017calibrating}. One alternative defines $q$ by a few steps of learnable sampling dynamics, such as Langevin or Hamiltonian updates started from a simple distribution, whose entropy can be tracked through the dynamics~\citep{dai2019exponential}; the dual view of~\Eqref{eqn:adv_minmax} also extends to kernel exponential families~\citep{dai2019kernel}. A second difficulty is that for a fixed $q$ the objective in~\Eqref{eqn:adv_minmax} is only a lower bound on the negative $\log$-likelihood, since $q$ only lower-bounds $\log Z_\theta$. If $q$ falls behind the current energy, minimizing over $\theta$ lowers the bound rather than the likelihood. In practice, $\theta$ and $q$ are updated in alternation, and the quality of the learned EBM depends on how closely $q$ tracks $p_\theta$.

\paragraph{Regression Training.}

Energy functions can also be learned through regression objectives that supervise the outputs or updates of energy minimization. The supervision need not specify scalar energy values: it can instead describe where optimization should arrive or how a candidate should move~\citep{dai2016discriminative}. IREM and Equilibrium Matching illustrate these two approaches.

\emph{Regressing optimized predictions.}
Iterative Reasoning through Energy Minimization (IREM)~\citep{du2022learning} learns a conditional energy $E_\theta(\vx,\vy)$ by supervising the prediction obtained after minimizing it. Given a paired input and solution $(\vx,\vy)$, initialize $\hat{\vy}^{(0)}$ and perform $K$ updates,
\begin{equation}
    \hat{\vy}^{(k+1)}
    = \hat{\vy}^{(k)}-\eta\nabla_{\vy}E_\theta(\vx,\hat{\vy}^{(k)}).
\end{equation}
The regression objective is
\begin{equation}\label{eq:irem_regression}
    \mathcal{L}_{\mathrm{IREM}}(\theta)
    =\EE_{(\vx,\vy)\sim\mathcal{D},\,\hat{\vy}^{(0)}}
    \left[\|\hat{\vy}^{(K)}-\vy\|_2^2\right].
\end{equation}
Training differentiates through optimization to shape the energy landscape. In practice, IREM truncates backpropagation to the final update and uses a replay buffer of optimized candidates alongside random initializations. The regression target is the solution reached by optimization, rather than the energy assigned to that solution. At inference time, the learned energy can be optimized for a different number of steps, although finite-step training alone does not guarantee that longer optimization improves predictions.

\emph{Regressing an update field.}
Equilibrium Matching (EqM)~\citep{wang2025equilibrium} instead supervises a time-independent field at corrupted data points. Given data $\vx$, Gaussian noise $\bm{\epsilon}$, and $\gamma\sim\mathcal{U}[0,1]$, let
\begin{equation}
    \vx_\gamma=(1-\gamma)\vx+\gamma\bm{\epsilon}.
\end{equation}
EqM regresses $f_\theta$ toward $(\bm{\epsilon}-\vx)c(\gamma)$, with $c(0)=0$. Its explicit-energy variant, EqM-E, sets $f_\theta=\nabla E_\theta$ and minimizes
\begin{equation}\label{eq:eqm_regression}
    \mathcal{L}_{\mathrm{EqM\text{-}E}}(\theta)
    =\EE_{\vx,\bm{\epsilon},\gamma}
    \left[\left\|\nabla_{\vx_\gamma}E_\theta(\vx_\gamma)
    -(\bm{\epsilon}-\vx)c(\gamma)\right\|_2^2\right].
\end{equation}
The model does not receive $\gamma$. The target magnitude vanishes at the data endpoint, encouraging equilibrium behavior. Generation starts from noise and follows the learned descent field. Training requires no unrolled generation trajectory. EqM-E enforces a scalar potential by construction; an unconstrained vector predictor need not be exactly conservative.

Thus, IREM supervises the \emph{result} of energy minimization, whereas EqM supervises its \emph{local update field}. Both avoid partition-function estimation, but neither objective is itself maximum likelihood for $\exp(-E_\theta)/Z_\theta$. In particular, optimization from random initializations in these models is not the same as equilibrium sampling from that Boltzmann density of the learned energy function.

\subsection{Unified View through Contrastive Learning}\label{sec:unified}
\begin{figure}[t]
    \centering
    \includegraphics[width=\textwidth]{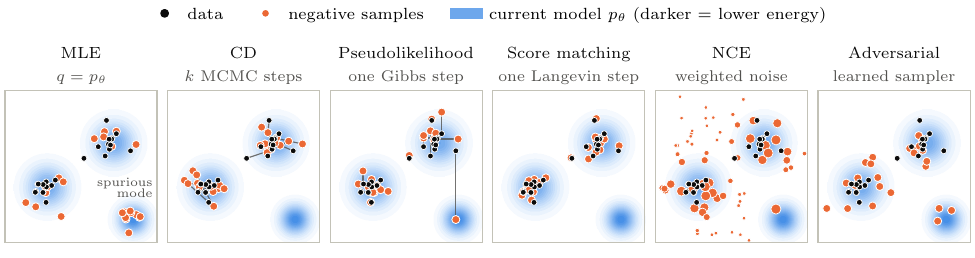}
    \caption{\textbf{Illustration of Negative Samples.} The negative samples of the learning methods in~\tabref{tab:negative_samplers}, for the same data and current model. Maximum likelihood draws from the model itself, including a spurious low-energy region far from the data. Contrastive divergence and score matching start at the data and move a short distance (lines connect each data point to its negative sample), so they never visit that region, and pseudolikelihood resamples one coordinate at a time, so it moves only along the axes. Noise-contrastive estimation weights noise samples by the classifier (marker size), and adversarial training draws from a learned sampler, which can lag behind the model.}
    \label{fig:negatives}
\end{figure}

The learning methods above look different, but their gradients share one structure~\citep{dai2019exponential}. The regression methods at the end of~\Secref{sec:learning} are the exception: they fit the update of a sampler rather than the likelihood of the energy, and fall outside this view. Each lowers the energy of the data and raises the energy of a set of \emph{negative samples},
\begin{equation}\label{eqn:unified_grad}
    \nabla_\theta\mathcal{L}(\theta) = \EE_{\vx\sim p_d}\sbr{\nabla_\theta E_\theta(\vx)} - \EE_{\vx\sim q}\sbr{\nabla_\theta E_\theta(\vx)},
\end{equation}
or, in the form of~\Eqref{eqn:loss_ebm_reduce}, $\mathcal{L}(\theta) = \EE_{p_d}\sbr{E_\theta} - \EE_{\texttt{stopgrad}\rbr{q}}\sbr{E_\theta}$. Maximum likelihood is the case $q = p_\theta$~\Eqref{eqn:gradient_ebm_reduce}. The other methods replace $p_\theta$ with a \emph{negative sampler} $q$ that is cheaper to draw from, and they differ mainly in this choice. \tabref{tab:negative_samplers} lists the samplers, which we explain below, and Figure~\ref{fig:negatives} shows them on a two-dimensional example. The view also tells where each method loses accuracy: wherever $q$ differs from $p_\theta$, the gradient differs from the maximum likelihood gradient.

\begin{table}[t]
\centering
\caption{Learning methods for EBMs as choices of the negative sampler $q$ in~\Eqref{eqn:unified_grad}, following~\citet{dai2019exponential}. $p_d$ is the data distribution, $T_\theta^k$ denotes $k$ steps of an MCMC kernel that leaves $p_\theta$ invariant, $\vx_{-j}$ is $\vx$ without its $j$-th variable, and $\sigmoid(h_{\theta,c})$ is the NCE classifier in~\Eqref{eqn:nce_posterior}.}
\label{tab:negative_samplers}
\small
\begin{tabular}{lll}
\hline
Method & Negative sampler $q$ & Gap from $p_\theta$ \\
\hline
Maximum likelihood & $p_\theta$ & none (exact) \\
Contrastive divergence & $T_\theta^k\, p_d$: $k$ MCMC steps from data & short chains \\
Persistent CD & $T_\theta^k$ from the previous iteration's samples & chains lag behind $\theta$ \\
Pseudolikelihood & $\frac{1}{d}\sum_{j} p_\theta(x_j\mid\vx_{-j})\,p_d(\vx_{-j})$: one Gibbs step from data & one variable at a time \\
Composite likelihood & block Gibbs step from data & one block at a time \\
Score matching & one small Langevin step from data & local, step size $\eta\to0$ \\
Binary NCE & $\rbr{p_d + \nu p_n}\,\sigmoid(h_{\theta,c})$: reweighted data and noise & exact only at $p_{\theta,c} = p_d$ \\
Ranking NCE & self-normalized importance sampling from $p_n$ & finite $K$, choice of $p_n$ \\
Adversarial training & learned sampler $q$ & how well $q$ tracks $p_\theta$ \\
\hline
\end{tabular}
\end{table}

\paragraph{Short MCMC Chains.} Contrastive divergence~(CD)~\citep{hinton2006training} starts the chains at the data and runs only $k$ steps, often a single one, so $q = T_\theta^k p_d$. Persistent CD~\citep{tieleman2008training} continues the chains from the previous iteration instead of restarting them, and short-run Langevin from noise or from a replay buffer~\citep{du2019implicit} is a further variant. These samplers are cheap because they never run to convergence. The price is that $q$ stays near where the chains started, so energy minima far from the data are rarely visited and rarely corrected. When short chains are restarted from noise, the learned energy need not be an accurate density, but the energy together with its short-run sampler is still a valid generator~\citep{nijkamp2019learning}.

\paragraph{Pseudolikelihood as One Gibbs Step.} The gradient of one term of~\Eqref{eqn:lpseudo} is
\begin{equation}\label{eqn:pl_grad}
    -\nabla_\theta\log p_\theta\rbr{x_j\mid\vx_{-j}} = \nabla_\theta E_\theta(\vx) - \EE_{x_j'\sim p_\theta(\cdot\mid\vx_{-j})}\sbr{\nabla_\theta E_\theta\rbr{x_j', \vx_{-j}}}.
\end{equation}
Averaging over $j$ gives~\Eqref{eqn:unified_grad} with the following sampler: take a data point, pick a variable $j$ uniformly at random, and resample it from its conditional. This is a single step of random-scan Gibbs sampling started at the data~\citep{hyvarinen2007some}. Composite likelihood~\citep{lindsay1988composite} resamples a block of variables instead of one, and non-local contrastive objectives~\citep{vickrey2010non} resample within a region chosen around the data point. Each enlarges the neighborhood the sampler can reach, which is why they capture more global structure than pseudolikelihood.

\paragraph{Score Matching as One Langevin Step.} Let $q$ take one Langevin step~\Eqref{eq:langevin} with step size $\eta/2$ from each data point, $\vx' = \vx - \frac{\eta}{2}\nabla_\vx E_\theta(\vx) + \sqrt{\eta}\,\bm{\xi}$ with $\bm{\xi}\sim\Ncal\rbr{0, I}$, and treat $\vx'$ as fixed, as in~\Eqref{eqn:loss_ebm_reduce}. A second-order Taylor expansion of $\nabla_\theta E_\theta(\vx')$ around $\vx$ gives~\citep{hyvarinen2007some,dai2019exponential}
\begin{equation}\label{eqn:sm_as_cd}
    \EE_{\vx\sim p_d}\sbr{\nabla_\theta E_\theta(\vx)} - \EE_{\vx'}\sbr{\nabla_\theta E_\theta(\vx')} = \frac{\eta}{2}\,\nabla_\theta\mathcal{L}_{\text{SM}}(\theta) + o(\eta),
\end{equation}
so score matching~\Eqref{eqn:ism} is the small-step limit of one-step Langevin CD. Its locality, noted in its limitations above, is the locality of this sampler. Minimum probability flow~\citep{sohl2011new} and minimum Stein discrepancy estimators~\citep{barp2019minimum} also correspond to fixed one-step samplers from the data, a symmetric transition kernel and one step of Stein variational gradient descent, respectively~\citep{dai2019exponential}.

\paragraph{NCE as Importance Sampling.} Splitting the data weight $1-\sigmoid$ in~\Eqref{eqn:nce_bin_grad} into $1$ and $-\sigmoid$ and collecting the $\sigmoid$ terms gives
\begin{equation}\label{eqn:nce_as_sampler}
    \nabla_\theta\mathcal{L}_{\text{NCE}}^{\text{bin}} = \EE_{\vx\sim p_d}\sbr{\nabla_\theta E_\theta(\vx)} - \int\rbr{p_d(\vx) + \nu\,p_n(\vx)}\,\sigmoid\rbr{h_{\theta,c}(\vx)}\,\nabla_\theta E_\theta(\vx)\,d\vx.
\end{equation}
The negatives are thus data and noise samples weighted by the classifier. Since $\sigmoid(h_{\theta,c}) = p_{\theta,c}/(p_{\theta,c} + \nu p_n)$, the weighted measure $\rbr{p_d + \nu p_n}\sigmoid(h_{\theta,c})$ equals $p_{\theta,c}$ exactly when $p_{\theta,c} = p_d$, so the negatives become model samples as the model approaches the data. Ranking NCE~\Eqref{eqn:nce_rank_grad} is explicitly self-normalized importance sampling with proposal $p_n$. In both cases, the quality of the sampler depends on how well $p_n$ covers $p_\theta$, which is the noise-choice problem discussed above.

\paragraph{Learned Samplers.} Adversarial training~\Eqref{eqn:adv_grad} is the only case where $q$ is trained to approach $p_\theta$, by the inner maximization of~\Eqref{eqn:adv_minmax}. From this view, the methods above fix $q$ by hand, which makes them cheap but leaves a gap that does not shrink with more training, while a learned sampler can close the gap at the cost of a min-max problem~\citep{dai2019exponential}.

The same template also explains contrastive representation learning. InfoNCE~\citep{oord2018representation} is ranking NCE~\Eqref{eqn:lnce_rank} for a conditional energy $E_\theta(\vx,\vy)$, where the negatives for $\vx$ are the $\vy$'s of other samples in the batch, \ie, $p_n$ is the marginal of the data. We return to this in~\Secref{sec:repr}.

\section{Discriminative Models}\label{sec:disc_model}

We first instantiate the EBM framework for discriminative tasks, where the goal is to predict an output $\vy$ from an input $\vx$. The energy is now defined over input--output pairs, $E_\theta(\vx,\vy)$, and it induces a conditional distribution
\begin{equation}\label{eqn:cond_ebm}
    p_\theta(\vy\mid\vx) = \frac{\exp\rbr{-E_\theta(\vx,\vy)}}{Z_\theta(\vx)}, \qquad Z_\theta(\vx) = \sum_{\vy}\exp\rbr{-E_\theta(\vx,\vy)}.
\end{equation}
which is the conditional EBM~\Eqref{eq:ebm_cond} with the integral replaced by a sum over outputs. Different choices of the output space and of the energy recover familiar models, including logistic regression, conditional random fields~(CRFs), and structured prediction energy networks.

\subsection{Logistic Regression and CRFs}\label{sec:crf}

When $\vy$ is one of a few classes, \Eqref{eqn:cond_ebm} is logistic regression; when $\vy$ is a structured output, such as a label sequence, and the energy decomposes over neighboring labels, it is a CRF~\citep{lafferty2001conditional}. The two differ in how the partition function $Z_\theta(\vx)$ is computed. For logistic regression, it is a sum over $K$ classes and is obtained by direct enumeration. For a linear-chain CRF, it has $K^T$ terms and cannot be enumerated, but the chain structure lets dynamic programming compute it exactly. In both cases, maximum likelihood can be carried out exactly, without the sampling-based approximations of~\Secref{sec:learning}.

\paragraph{Logistic Regression.} Let $y\in\cbr{1,\ldots,K}$ and let the energy be linear in a feature map $\fb(\vx)$,
\begin{equation}\label{eqn:lr_energy}
    E_\theta(\vx, y) = -\vw_y^\top\fb(\vx) - b_y, \qquad
    p_\theta(y\mid\vx) = \frac{\exp\rbr{\vw_y^\top\fb(\vx) + b_y}}{\sum_{k=1}^{K}\exp\rbr{\vw_k^\top\fb(\vx) + b_k}},
\end{equation}
with $\theta = \cbr{\vw_k, b_k}_{k=1}^K$. The conditional distribution is the softmax of the negative energies, and $K=2$ gives the sigmoid. The partition function is a sum of $K$ terms, so the negative $\log$-likelihood can be computed exactly,
\begin{equation}\label{eqn:lr_nll}
    \mathcal{L}(\theta) = \EE_{(\vx,y)\sim\mathcal{D}}\sbr{E_\theta(\vx,y) + \log\sum_{k=1}^{K}\exp\rbr{-E_\theta(\vx,k)}}.
\end{equation}
Its gradient has the form of~\Eqref{eqn:gradient_ebm_reduce}, with the expectation under the model computed exactly by summing over the $K$ classes,
\begin{equation}\label{eqn:lr_grad}
\begin{aligned}
    \nabla_\theta\mathcal{L} &= \EE_{(\vx,y)\sim\mathcal{D}}\sbr{\nabla_\theta E_\theta(\vx,y) - \sum_{k=1}^{K}p_\theta(k\mid\vx)\,\nabla_\theta E_\theta(\vx,k)}, \\
    \nabla_{\vw_k}\mathcal{L} &= \EE_{(\vx,y)\sim\mathcal{D}}\sbr{\rbr{p_\theta(k\mid\vx) - \1[y=k]}\fb(\vx)}.
\end{aligned}
\end{equation}
The gradient for class $k$ is the gap between the predicted probability and the label, times the features. For fixed features, $\mathcal{L}$ is convex, and \Algref{alg:lr_mle} gives the stochastic gradient version of MLE; for small problems, Newton's method, known as iteratively reweighted least squares, converges faster. When $\fb$ is a neural network trained jointly, the same algorithm is the standard softmax cross-entropy training of classifiers.

\begin{algorithm}[t]
\caption{Maximum likelihood for logistic regression by stochastic gradient descent}
\label{alg:lr_mle}
\begin{algorithmic}[1]
\Require data $\mathcal{D}$, features $\fb$, step size $\eta$, batch size $B$
\State Initialize $\vw_k \gets \mathbf{0}$, $b_k \gets 0$ for $k = 1,\ldots,K$
\Repeat
    \State Sample a minibatch $\cbr{(\vx_i, y_i)}_{i=1}^{B}$ from $\mathcal{D}$
    \State Compute $p_{ik} \gets p_\theta(k\mid\vx_i)$ by~\Eqref{eqn:lr_energy} for all $i, k$ \Comment{exact negative phase}
    \State $\vw_k \gets \vw_k - \frac{\eta}{B}\sum_{i=1}^{B}\rbr{p_{ik} - \1[y_i=k]}\fb(\vx_i)$ for all $k$
    \State $b_k \gets b_k - \frac{\eta}{B}\sum_{i=1}^{B}\rbr{p_{ik} - \1[y_i=k]}$ for all $k$
\Until{converged}
\end{algorithmic}
\end{algorithm}

\paragraph{Linear-Chain CRF.} For a label sequence $\vy = (y_1,\ldots,y_T)$ with $y_t\in\cbr{1,\ldots,K}$, a linear-chain CRF~\citep{lafferty2001conditional}, \ie, the CRF~\Eqref{eq:crf} with cliques over neighboring labels, lets the energy decompose into a unary term for each position and a pairwise term for each pair of neighboring labels,
\begin{equation}\label{eqn:crf_energy}
    E_\theta(\vx,\vy) = -\sum_{t=1}^{T} s_\theta(\vx, t, y_t) - \sum_{t=2}^{T} A_{y_{t-1}, y_t},
\end{equation}
where $s_\theta(\vx,t,k)$ scores label $k$ at position $t$, through a linear model or a neural network, and $A\in\RR^{K\times K}$ scores label transitions. The sum in $Z_\theta(\vx)$ now has $K^T$ terms, but the chain structure lets the forward algorithm compute it by dynamic programming in $O(TK^2)$ time,
\begin{equation}\label{eqn:crf_forward}
    \alpha_1(k) = e^{s_\theta(\vx,1,k)}, \qquad
    \alpha_t(k) = e^{s_\theta(\vx,t,k)}\sum_{j=1}^{K}\alpha_{t-1}(j)\,e^{A_{jk}}, \qquad
    Z_\theta(\vx) = \sum_{k=1}^{K}\alpha_T(k),
\end{equation}
carried out in $\log$ space for numerical stability. MLE then applies directly. The gradient again has the form of~\Eqref{eqn:gradient_ebm_reduce}, and because the energy is a sum of local terms, the expectation under the model only needs the marginals $p_\theta(y_t\mid\vx)$ and $p_\theta(y_{t-1},y_t\mid\vx)$, which the forward-backward algorithm computes exactly in $O(TK^2)$. In practice, one computes $\log Z_\theta(\vx)$ with~\Eqref{eqn:crf_forward} and lets automatic differentiation produce this gradient, then optimizes with L-BFGS or stochastic gradient descent. Prediction is energy minimization, $\vy^\star = \argmin_\vy E_\theta(\vx,\vy)$, solved exactly by the Viterbi algorithm, which replaces the sum in~\Eqref{eqn:crf_forward} with a max.

\paragraph{General CRFs.} Exact MLE relies on the chain structure, and it extends to trees and graphs of low treewidth through the junction tree algorithm. For CRFs on loopy graphs, such as grid CRFs for image segmentation, $Z_\theta(\vx)$ and the marginals are intractable, and the learning methods of~\Secref{sec:learning} apply with the label $\vy$ in place of $\vx$: approximate marginals from loopy belief propagation or mean-field~\citep{wainwright2008graphical}, pseudolikelihood over each label given its neighbors~\citep{besag1975statistical}, or contrastive divergence with Gibbs sampling.

\paragraph{Max-Margin Training.} Alternatively, one can drop the probabilistic model and train with only the minimum-energy output. The structured perceptron~\citep{collins2002discriminative}, max-margin Markov networks~\citep{taskar2003max}, and structured support vector machines~(structured SVMs)~\citep{tsochantaridis2005large} replace the expectation under the model in~\Eqref{eqn:gradient_ebm_reduce} with the energy minimizer, or its loss-augmented version for the margin-based methods, which only requires MAP inference. These methods are still EBMs: they learn an energy over input--output pairs and predict by minimizing it, but they shape the energy with a margin loss instead of the likelihood, so the energy never needs to be normalized~\citep{lecun2006tutorial}. This makes them applicable whenever the energy can be minimized, even when $Z_\theta(\vx)$ cannot be computed. The same idea underlies the SPENs of~\Secref{sec:spen}, which keep the structured SVM loss and replace its linear energy with a deep network.

\subsection{Structured Prediction Energy Networks}\label{sec:spen}

A CRF fixes the interactions among labels in advance through a graph over the labels. In a linear-chain CRF~\Eqref{eqn:crf_energy}, only neighboring labels interact, which is what makes $Z_\theta(\vx)$ tractable; richer graphs capture more interactions but need the approximations discussed above. Structured prediction energy networks~(SPENs)~\citep{belanger2016structured} drop this restriction. The energy is a deep network over the whole output, so it can capture interactions among all labels, and inference becomes optimization over $\vy$ instead of dynamic programming.

\paragraph{Energy.} For multi-label classification with $L$ labels, SPENs relax the binary output to $\vy\in[0,1]^L$ and use an energy with two terms,
\begin{equation}\label{eqn:spen_energy}
    E_\theta(\vx,\vy) = -\sum_{i=1}^{L} y_i\,\vb_i^\top F(\vx) - \vc_2^\top g\rbr{C_1\vy},
\end{equation}
where $F(\vx)$ is a feature network, $\vb_i$, $\vc_2$, and $C_1$ are learned parameters, and $g$ is an elementwise nonlinearity. The first term is local: it scores each label from the input, as in logistic regression. The second term is global: $\vc_2^\top g\rbr{C_1\vy}$ is a one-hidden-layer network over the whole label vector, which can learn which labels tend to appear together, without specifying a graph.

\paragraph{Inference.} Prediction is energy minimization, as in~\Secref{sec:inference},
\begin{equation}\label{eqn:spen_inference}
    \hat{\vy} = \argmin_{\vy\in[0,1]^L}\; E_\theta(\vx,\vy),
\end{equation}
solved by projected gradient descent on $\vy$, with $\vy$ rounded to binary labels at the end. Since $E_\theta$ is not convex in $\vy$, this finds a local minimum, and the result depends on the initialization and the number of steps.

\paragraph{Learning.} $Z_\theta(\vx)$ is an integral over $\vy$ with a non-decomposable energy, so maximum likelihood is out of reach, and SPENs are trained without a probabilistic model. The original SPEN is trained as a structured SVM~\citep{tsochantaridis2005large}, one of the max-margin methods mentioned at the end of the previous subsection, with the structured hinge loss
\begin{equation}\label{eqn:spen_hinge}
    \mathcal{L}(\theta) = \sum_{(\vx,\vy)\in\mathcal{D}}\max_{\vy'}\sbr{\Delta(\vy,\vy') + E_\theta(\vx,\vy) - E_\theta(\vx,\vy')}_+,
\end{equation}
where $\Delta(\vy,\vy')$ is a task loss, such as the Hamming distance. Its gradient is $\nabla_\theta E_\theta(\vx,\vy) - \nabla_\theta E_\theta(\vx,\hat\vy')$, where $\hat\vy'$ is the maximizer, found by gradient descent on $\vy'$. This is the contrastive form~\Eqref{eqn:unified_grad} with a single negative sample: the output with low energy but high task loss.

A SPEN is thus a structured SVM with a deep energy. A structured SVM uses an energy that is linear in the parameters, $E_\vw(\vx,\vy) = -\vw^\top\Psi(\vx,\vy)$, with a joint feature map $\Psi$ designed so that the maximization over $\vy'$ in~\Eqref{eqn:spen_hinge}, known as loss-augmented inference, can be solved exactly, \eg, by dynamic programming or graph cuts. The training problem is then convex. SPENs keep the same loss and replace the linear energy with the deep energy~\Eqref{eqn:spen_energy}. This removes the need to design $\Psi$ and lets the energy learn interactions among all labels, but loss-augmented inference becomes gradient descent over a relaxed $\vy'$, which is only approximate, and the training problem is no longer convex.

End-to-end SPENs~\citep{belanger2017end} replace the hinge loss altogether: they unroll a fixed number of gradient steps of~\Eqref{eqn:spen_inference} and backpropagate a task loss on the final prediction through them, which is the regression training of~\Secref{sec:learning} applied to a discriminative energy.

Two later lines of work change how the energy is trained or used. Deep value networks~\citep{gygli2017deep} train the energy to regress the task score of an output directly. Inference networks~\citep{tu2018learning} train a separate network $A_\omega(\vx)$ to approximate the minimizer in~\Eqref{eqn:spen_inference}, so prediction is a single forward pass. The latter mirrors adversarial training in~\Secref{sec:learning}: the energy is trained against a learned network that searches for its low-energy outputs, except that here the network finds a minimizer rather than a sampler.

\section{Generative Models}\label{sec:gen_model}

Many existing generative models can be viewed as special cases of EBMs. Each defines, explicitly or implicitly, a distribution $p_\theta(\vx)\propto\exp\rbr{-E_\theta(\vx)}$, and they differ mainly in how they handle two parts that~\Secref{sec:prelim} treated separately: the potential function $E_\theta$ and the sampler that draws from it. A general EBM parameterizes the potential freely, so it can express any shape, but it must pay at sampling time, with MCMC or a learned sampler, and at training time, with the methods of~\Secref{sec:learning}. Other generative models restrict or learn one of the two parts so that the other becomes easy.

The following subsections take the main families in turn. In GANs, the generator plays the role of the learned sampler in adversarial training and the discriminator plays the role of the potential; adding an entropy term on the generator turns the pair into an EBM trained by maximum likelihood. Normalizing flows build the sampler as an invertible map, which makes the potential, including its normalization, exact by change of variables. Diffusion and flow-matching models learn the gradient of the potential at many noise levels and sample by following it, as in annealed Langevin dynamics. Energy-inspired models take a sampler that uses a learned potential as the model itself, so that sampling is exact and the likelihood has a tractable bound; modern Hopfield networks, discussed alongside them, retrieve stored patterns by energy minimization; and deep equilibrium models define a layer by the fixed point of an iterative update, such as energy minimization. Finally, autoregressive models write the potential as a sum of normalized conditionals, which makes sampling one variable at a time exact.

\subsection{Generative Adversarial Nets}\label{sec:gan}

The adversarial training of~\Secref{sec:learning} already has the structure of a GAN. We make the correspondence explicit, and then use it to read the original GAN and its variants from the EBM view.

\paragraph{Entropy-Regularized GAN.} Take the sampler $q$ in~\Eqref{eqn:adv_minmax} to be an implicit generator, $\vx = g_\phi(\vz)$ with $\vz\sim\Ncal\rbr{0, I}$, and write $q_\phi$ for its distribution. Maximum likelihood for the EBM becomes
\begin{equation}\label{eqn:ergan}
    \min_\theta\,\max_\phi\;\EE_{\vx\sim p_d}\sbr{E_\theta(\vx)} - \EE_{\vz}\sbr{E_\theta\rbr{g_\phi(\vz)}} + H\rbr{q_\phi}.
\end{equation}
The energy acts as the discriminator: it is lowered on data and raised on generated samples. The generator moves its samples toward low energy, while the entropy term keeps them spread out. The two players have the same roles as in~\Secref{sec:learning}, and so does the result. If both families are rich enough, the solution satisfies $q_\phi = p_\theta = p_d$: the energy recovers $-\log p_d$ up to a constant, and the generator is a sampler for it. Training yields both parts of the EBM, the potential and the sampler~\citep{kim2016deep,dai2017calibrating,dai2019exponential}. The practical difficulty is the one noted in~\Secref{sec:learning}: the entropy of an implicit generator has no closed form and must be estimated or bounded.

\paragraph{The Original GAN.} The original GAN~\citep{goodfellow2014generative} trains a discriminator $D(\vx) = \sigmoid\rbr{\ell(\vx)}$ to tell data from generated samples,
\begin{equation}\label{eqn:gan}
    \min_\phi\,\max_\ell\;\EE_{\vx\sim p_d}\sbr{\log\sigmoid\rbr{\ell(\vx)}} + \EE_{\vz}\sbr{\log\rbr{1 - \sigmoid\rbr{\ell(g_\phi(\vz))}}}.
\end{equation}
For a fixed generator, the optimal discriminator is $D^\star = p_d/(p_d + q_\phi)$, so its logit is the $\log$-density ratio, $\ell^\star = \log p_d - \log q_\phi$, and the generator then minimizes the Jensen--Shannon divergence between $p_d$ and $q_\phi$~\citep{goodfellow2014generative}.

The discriminator objective in~\Eqref{eqn:gan} is binary NCE~\Eqref{eqn:lnce_bin} with $\nu = 1$ and the generator as the noise distribution. Comparing $\ell$ with the NCE logit~\Eqref{eqn:nce_posterior} shows what is missing: NCE parameterizes $\ell(\vx) = -E_\theta(\vx) - \log p_n(\vx) - c$, which requires the noise density, whereas an implicit generator has no tractable $q_\phi(\vx)$. The GAN therefore learns the ratio $\ell$ directly instead of an energy. This has two consequences. First, as $q_\phi$ approaches $p_d$, the $\log$-ratio goes to zero and the discriminator becomes constant, $D^\star = 1/2$, so at the solution the potential is discarded and only the sampler remains. The energy could in principle be read off as $-E_\theta = \ell + \log q_\phi$ up to a constant, but this requires $\log q_\phi$. Without the generator's density, the EBM cannot be obtained from a trained GAN. Second, the generator update has no entropy term, so the generator's best response is not tied to an EBM. For a fixed energy, the entropy-regularized generator maximizes $-\EE_{q}\sbr{E_\theta} + H\rbr{q} = -\mathrm{KL}\rbr{q\|p_\theta} + \log Z_\theta$, whose unique maximizer is the EBM itself, $q = p_\theta\propto\exp\rbr{-E_\theta}$. Without the entropy, the generator only maximizes $-\EE_q\sbr{E_\theta}$, or, in~\Eqref{eqn:gan}, minimizes $\EE_q\sbr{\log\rbr{1-\sigmoid(\ell)}}$; in both cases, the best response is a point mass at the minimum of the energy, or at the maximum of $\ell$. The EBM is then just one of the distributions the generator may settle on, with no force pulling it there, and the game can reach solutions outside the space of EBMs, such as a generator that covers only some modes of the data; this is one source of mode collapse.

The entropy term in~\Eqref{eqn:ergan} addresses both points: it makes the generator's objective $\mathrm{KL}\rbr{q_\phi\|p_\theta}$, so every solution of the game is an EBM, and it keeps the discriminator a valid energy at the solution. Both require the generator's density: the entropy $H\rbr{q_\phi} = -\EE_{q_\phi}\sbr{\log q_\phi}$ requires $\log q_\phi$, which is exactly what an implicit generator lacks. This is why entropy-regularized GANs either use generators with tractable densities or estimate the entropy, as discussed in~\Secref{sec:learning}.

\paragraph{Other GAN Variants.} Several GAN variants change the discriminator loss, and the EBM view classifies them by how the loss shapes the potential. Each loss relaxes the constraints that an EBM places on the model: the potential need not be normalized, and the generator need not sample from it. The $f$-GAN~\citep{nowozin2016f} replaces the Jensen--Shannon divergence with any $f$-divergence through its variational representation. Its optimal discriminator is again a function of the density ratio $p_d/q_\phi$, and its generator again has no entropy term, so both points above apply unchanged: the energy cannot be recovered without $\log q_\phi$, and the solutions of the game need not be EBMs. The Wasserstein GAN~\citep{arjovsky2017wasserstein} uses a critic $f$ and the objective $\EE_{p_d}\sbr{f} - \EE_{q_\phi}\sbr{f}$ over $1$-Lipschitz $f$. With $f = -E_\theta$, the critic objective looks like the contrastive maximum likelihood loss~\Eqref{eqn:loss_ebm_reduce}, but the resemblance is only in form. Maximum likelihood requires the negative samples to come from the EBM itself, $p_\theta\propto\exp\rbr{-E_\theta}$. In the Wasserstein GAN, the generator is trained to reduce the Wasserstein distance to $p_d$, and like the original GAN it has no entropy term, so its best response to a fixed critic is again a point mass rather than $p_\theta$. The generator is therefore not a sampler for the EBM defined by the critic, and the critic's gradient is not the maximum likelihood gradient. The Lipschitz constraint, rather than normalization, keeps the critic bounded, and at the solution the critic is the potential that defines the Wasserstein distance, not a $\log$-density.

\begin{table}[t]
\centering
\caption{GANs from the EBM view: the discriminator loss, whether the generator has an entropy term (Entropy), and what the discriminator represents at the solution.}
\label{tab:gans}
\small
\begin{tabular}{llll}
\hline
Model & Discriminator loss & Entropy & Discriminator at the solution \\
\hline
Original GAN & binary classification & no & constant, $D^\star = 1/2$ \\
Wasserstein GAN & $\EE_{p_d}\sbr{f} - \EE_{q_\phi}\sbr{f}$, Lipschitz $f$ & no & critic for a distance \\
Entropy-regularized GAN & maximum likelihood & yes & energy, $E_\theta = -\log p_d + \text{const}$ \\
\hline
\end{tabular}
\end{table}

\tabref{tab:gans} summarizes the comparison. In every case the generator is a learned sampler. What separates the entropy-regularized GAN is that the discriminator is trained by maximum likelihood and the generator by $\mathrm{KL}\rbr{q_\phi\|p_\theta}$, so both parts of the EBM survive training. The learned energy can then be used beyond sampling, for example to evaluate unnormalized likelihoods, to refine generated samples with a few MCMC steps, or to be composed with other energies as in~\Secref{sec:compose}.

\subsection{Normalizing Flows}\label{sec:flow}

For a general EBM, the partition function $Z_\theta$ is intractable, and this is the source of the difficulties in sampling and learning discussed in~\Secref{sec:prelim}. Normalizing flows avoid it with a special parametrization: the model is an invertible transformation of a simple distribution, so the change-of-variables formula gives its density, normalization included.

\paragraph{Change of Variables.} Let $\vz\sim p_0$ be a simple base distribution, such as $\Ncal\rbr{0, I}$, and let $\vx = g_\theta(\vz)$, where $g_\theta:\RR^d\to\RR^d$ is invertible and differentiable. Since $g_\theta$ only moves probability mass around without creating or removing any, the density of $\vx$ is
\begin{equation}\label{eqn:flow_cov}
    p_\theta(\vx) = p_0\rbr{g_\theta^{-1}(\vx)}\,\abr{\det\frac{\partial g_\theta^{-1}(\vx)}{\partial\vx}}.
\end{equation}
This density is tractable when $g_\theta^{-1}$ and the determinant of its Jacobian are cheap to compute, and normalizing flows are built from layers that meet both requirements. Coupling layers~\citep{dinh2015nice,dinh2017density} transform half of the variables with a function of the other half, so the Jacobian is triangular and its determinant is the product of the diagonal. Autoregressive flows~\citep{papamakarios2017masked,kingma2016improved} transform each variable conditioned on the previous ones, which also gives a triangular Jacobian. Glow~\citep{kingma2018glow} adds invertible $1\times1$ convolutions. Continuous normalizing flows~\citep{chen2018neural,grathwohl2019ffjord} define $g_\theta$ by an ordinary differential equation, in which case the $\log$-determinant becomes the integral of a trace along the trajectory. A deep flow composes many such layers, and the $\log$-determinants add up.

\paragraph{Normalizing Flows as EBMs.} Taking the negative $\log$ of~\Eqref{eqn:flow_cov} writes the flow in the form of~\Eqref{eq:ebm},
\begin{equation}\label{eqn:flow_energy}
    p_\theta(\vx) = \frac{\exp\rbr{-E_\theta(\vx)}}{Z_\theta}, \qquad
    E_\theta(\vx) = -\log p_0\rbr{g_\theta^{-1}(\vx)} - \log\abr{\det\frac{\partial g_\theta^{-1}(\vx)}{\partial\vx}}, \qquad
    Z_\theta = 1.
\end{equation}
The energy has two terms: how unlikely the corresponding latent point is under the base distribution, and how much the flow compresses or expands volume at $\vx$. This special parametrization makes the partition function of the EBM tractable: it equals one for every $\theta$, because the change of variables preserves total mass. More generally, if the base distribution is itself an EBM, $p_0\propto\exp\rbr{-E_0}$ with partition function $Z_0$, then the same argument shows that the flow has partition function $Z_0$ as well. A flow never changes the partition function; it moves the normalization into the base distribution, where it is known.

\paragraph{Learning and Sampling.} Because $Z_\theta$ does not depend on $\theta$, maximum likelihood is exact. In the gradient~\Eqref{eqn:gradient_ebm_reduce}, the expectation under the model equals $-\nabla_\theta\log Z_\theta = 0$, so the negative phase vanishes and the gradient reduces to $\EE_{\vx\sim p_d}\sbr{\nabla_\theta E_\theta(\vx)}$, with no samples from the model needed during training. Sampling is also exact and takes one pass: draw $\vz\sim p_0$ and return $g_\theta(\vz)$. A flow therefore makes both parts of the EBM easy at once, the potential with its normalization and the sampler.

\paragraph{Normalizing Flows and General EBMs.} Since a flow has an exact density, and hence an exact entropy, it is a natural choice for the learned sampler in adversarial training~\Eqref{eqn:adv_minmax} and in variational inference~\citep{rezende2015variational}, which is the first difficulty noted for implicit generators in~\Secref{sec:gan}. A flow can also serve as the noise distribution of NCE~\Eqref{eqn:lnce_bin}, where its exact density is what NCE needs; flow contrastive estimation trains the flow and the EBM jointly, so that the noise stays close to the data~\citep{gao2020flow}.

\paragraph{Limitations.} The tractability comes from architectural constraints. The map must be invertible, so $\vx$ and $\vz$ have the same dimension and there is no bottleneck, and each layer must have a cheap Jacobian determinant, which limits how much each layer can do. Invertibility also preserves topology: mapping a unimodal Gaussian onto data with separated modes requires very steep transformations between the modes. A general EBM has none of these constraints, but pays with an intractable partition function. This is the trade-off described at the start of this section: flows restrict the potential so that its normalization and its sampler come for free.

\subsection{Diffusion and Flow Models}\label{sec:diffusion}

Diffusion and flow models generate data by gradually transforming noise into samples. We first introduce both families in their usual form, and then use Tweedie's formula to show that they learn a sequence of EBMs, one for each noise level, and differ from EBMs mainly in how they parameterize it.

\paragraph{Diffusion Models.} A diffusion model~\citep{sohl2015deep,ho2020denoising,song2021score} fixes a forward process that corrupts a data point $\vx_0\sim p_d$ with Gaussian noise of increasing scale. All we need is its marginal at each time,
\begin{equation}\label{eqn:diff_forward}
    \vx_t = \alpha_t\vx_0 + \sigma_t\bm{\epsilon}, \qquad \bm{\epsilon}\sim\Ncal\rbr{0, I}, \qquad t\in[0,1],
\end{equation}
that is, $\vx_t\mid\vx_0\sim\Ncal\rbr{\alpha_t\vx_0,\sigma_t^2 I}$. The schedule satisfies $\alpha_0=1$ and $\sigma_0=0$, and the signal-to-noise ratio $\alpha_t^2/\sigma_t^2$ decreases to nearly zero at $t=1$. Variance-preserving diffusion takes $\alpha_t^2+\sigma_t^2=1$, variance-exploding diffusion takes $\alpha_t=1$ with $\sigma_t$ growing large, and the linear interpolation $\alpha_t=1-t$, $\sigma_t=t$ is used by rectified flow. Let $p_t$ denote the distribution of $\vx_t$. To generate data, the process is run backward in time, from noise at $t=1$ to data at $t=0$, by the reverse-time stochastic differential equation or the deterministic probability-flow ordinary differential equation~\citep{song2021score}. Both only require the score $\nabla_\vx\log p_t(\vx)$ at every $t$. Diffusion models learn it with a network $\bm{\epsilon}_\theta(\vx_t, t)$ trained to predict the noise that was added,
\begin{equation}\label{eqn:diff_loss}
    \mathcal{L}(\theta) = \EE_{t,\,\vx_0\sim p_d,\,\bm{\epsilon}}\sbr{w(t)\,\big\|\bm{\epsilon}_\theta\rbr{\alpha_t\vx_0 + \sigma_t\bm{\epsilon}, t} - \bm{\epsilon}\big\|_2^2},
\end{equation}
with a weighting $w(t)$ over noise levels. At each $t$, this is denoising score matching~\Eqref{eqn:dsm}~\citep{vincent2011connection}, which corresponds to $\alpha_t = 1$, generalized to $\alpha_t\neq1$, as made precise below.

\paragraph{Flow Models.} Flow matching and rectified flow~\citep{lipman2023flow,liu2023flow,albergo2023building} use the same interpolation~\Eqref{eqn:diff_forward} between data and noise, but learn the velocity of the path instead of the noise. The network $\vv_\theta(\vx_t, t)$ is regressed onto the time derivative of the interpolation,
\begin{equation}\label{eqn:fm_loss}
    \mathcal{L}(\theta) = \EE_{t,\,\vx_0\sim p_d,\,\bm{\epsilon}}\sbr{\big\|\vv_\theta\rbr{\vx_t, t} - \rbr{\dot\alpha_t\vx_0 + \dot\sigma_t\bm{\epsilon}}\big\|_2^2},
\end{equation}
and samples are generated by integrating the ordinary differential equation $d\vx_t/dt = \vv_\theta(\vx_t, t)$ from noise to data. Some works use the reverse time convention, with noise at $t=0$; we follow the diffusion convention throughout. The minimizer of~\Eqref{eqn:fm_loss} is the conditional expectation $\vv^\star(\vx,t) = \dot\alpha_t\EE\sbr{\vx_0\mid\vx_t=\vx} + \dot\sigma_t\EE\sbr{\bm{\epsilon}\mid\vx_t=\vx}$.

\paragraph{Tweedie's Formula.} The two families, and their different targets, are connected by Tweedie's formula~\citep{efron2011tweedie}. Since $p_t(\vx) = \int p_d(\vx_0)\,\Ncal\rbr{\vx;\alpha_t\vx_0, \sigma_t^2 I}\,d\vx_0$, differentiating under the integral gives $\nabla_\vx p_t(\vx) = \int p_d(\vx_0)\,\Ncal\rbr{\vx;\alpha_t\vx_0,\sigma_t^2 I}\,\frac{\alpha_t\vx_0 - \vx}{\sigma_t^2}\,d\vx_0$, and dividing by $p_t(\vx)$ yields
\begin{equation}\label{eqn:tweedie}
    \nabla_\vx\log p_t(\vx) = \frac{\alpha_t\,\EE\sbr{\vx_0\mid\vx_t = \vx} - \vx}{\sigma_t^2}, \qquad\text{equivalently}\qquad
    \EE\sbr{\vx_0\mid\vx_t=\vx} = \frac{\vx + \sigma_t^2\nabla_\vx\log p_t(\vx)}{\alpha_t}.
\end{equation}
The score of the noisy distribution is the direction from $\vx$ toward its denoised estimate. Taking the conditional expectation of $\vx_t=\alpha_t\vx_0+\sigma_t\bm{\epsilon}$ given $\vx_t=\vx$ gives $\vx=\alpha_t\EE\sbr{\vx_0\mid\vx}+\sigma_t\EE\sbr{\bm{\epsilon}\mid\vx}$, so the expected noise is $\EE\sbr{\bm{\epsilon}\mid\vx}=\rbr{\vx-\alpha_t\EE\sbr{\vx_0\mid\vx}}/\sigma_t=-\sigma_t\nabla_\vx\log p_t(\vx)$ by~\Eqref{eqn:tweedie}. Since this conditional expectation is the minimizer of~\Eqref{eqn:diff_loss}, the optimal noise predictor is the score scaled by $-\sigma_t$. The noise prediction, the data prediction, and the velocity are therefore all affine functions of the score, with coefficients that depend only on $t$.

\paragraph{Diffusion Models as EBMs.} Now write each noisy distribution as an EBM with a time-dependent energy,
\begin{equation}\label{eqn:diff_ebm}
    p_t(\vx) = \frac{\exp\rbr{-E_\theta(\vx, t)}}{Z_\theta(t)}, \qquad \nabla_\vx\log p_t(\vx) = -\nabla_\vx E_\theta(\vx, t),
\end{equation}
where, as in score matching, $Z_\theta(t)$ drops out of the score. We emphasize that this connection, like score matching itself, relies on the gradient $\nabla_\vx$ and the Gaussian corruption~\Eqref{eqn:diff_forward}, so it holds only for continuous $\vx$. For discrete data, there is no gradient; the score is replaced by ratios of probabilities between neighboring states, as in the ratio matching discussed with score matching in~\Secref{sec:learning}, and discrete diffusion models learn such ratios or the corresponding conditionals instead~\citep{austin2021structured,sun2023score,lou2024discrete}. By Tweedie's formula, every target used by diffusion and flow models is determined by the gradient of this energy,
\begin{equation}\label{eqn:diff_param}
\begin{aligned}
    \bm{\epsilon}^\star(\vx,t) &= \sigma_t\nabla_\vx E_\theta(\vx,t), \qquad
    \vx_0^\star(\vx,t) = \frac{\vx - \sigma_t^2\nabla_\vx E_\theta(\vx,t)}{\alpha_t}, \\
    \vv^\star(\vx,t) &= \frac{\dot\alpha_t}{\alpha_t}\vx - \sigma_t\rbr{\frac{\dot\alpha_t\sigma_t}{\alpha_t} - \dot\sigma_t}\nabla_\vx E_\theta(\vx,t).
\end{aligned}
\end{equation}
Substituting the first expression into~\Eqref{eqn:diff_loss} recovers denoising score matching~\citep{vincent2011connection}, \ie,~\Eqref{eqn:dsm}, for the energy $E_\theta(\cdot, t)$, weighted over noise levels. A diffusion model is therefore a family of EBMs, $\cbr{p_t}_{t\in[0,1]}$, learned by denoising score matching~\citep{vincent2011connection,song2019generative}. The energy at $t=0$ describes the data, and the energies at larger $t$ are smoothed versions of it whose landscapes are easier to traverse. Sampling moves through this family from smooth to sharp, which is the annealed Langevin dynamics of~\citet{song2019generative} in its original form, and the reverse-time dynamics~\citep{song2021score} in general. This connection has also been used to learn EBMs, by parameterizing the scalar energy $E_\theta$ and fitting $-\nabla_\vx E_\theta$ by diffusion or flow matching~\citep{thornton2025composition,balcerak2025energy}; since the loss involves $\nabla_\vx E_\theta$, each training step requires the mixed second derivative $\nabla_\theta\nabla_\vx E_\theta$, \ie, backpropagation through $\nabla_\vx E_\theta$, which is expensive in computation and memory and may introduce numerical instability.

\begin{figure}[t]
    \centering
    \includegraphics[width=0.85\textwidth]{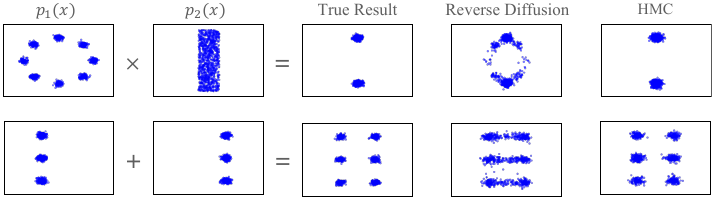}
    \caption{\textbf{Sampling compositions of two diffusion models.} Left to right: the two component distributions, samples from the true composition, and samples from reverse diffusion and from annealed MCMC (HMC) on the composed models. Top: product; bottom: mixture. Reverse diffusion fails to sample either composition, whereas MCMC on the composed energies recovers both. For the product, summing the scores of the two models at each noise level does not give the score of the noised product, since adding noise does not commute with taking products. Reproduced from~\citet{du2023reduce}.}
    \label{fig:rrr_samplers}
    \vspace{-10pt}
\end{figure}

\paragraph{The Difference Is the Parameterization.} In terms of the two parts of~\Secref{sec:gen_model}, diffusion and flow models keep the EBM's potential but parameterize it through its gradient: the network outputs the vector $\bm{\epsilon}_\theta$, $\vx_{0,\theta}$, or $\vv_\theta$ directly, and~\Eqref{eqn:diff_param} is read in reverse to define an implied score. An EBM instead parameterizes the scalar $E_\theta(\vx, t)$ and obtains the score by differentiation. The two are trained with the same loss, and differ in three ways.
\begin{itemize}
    \item \textbf{Conservativeness.} A score obtained as $-\nabla_\vx E_\theta$ is a gradient field. A vector-valued network is not constrained to be one, so the implied score need not be the gradient of any energy, for the same reason as for the deep equilibrium layers of~\Secref{sec:deq}.
    \item \textbf{Access to the energy.} With a scalar energy, $\log p_t$ can be evaluated up to a constant. This enables Metropolis--Hastings corrections~\Eqref{eq:mh_accept} and Hamiltonian Monte Carlo during sampling, and comparisons between samples, none of which is possible with the vector parameterization.
    \item \textbf{Cost.} The scalar parameterization needs an extra backward pass to obtain the score, and in practice it has reached comparable but not better sample quality~\citep{salimans2021should,du2023reduce}.
\end{itemize}
The difference matters most for composition~(\Secref{sec:compose}). The product of two EBMs has energy $E_1 + E_2$, but adding the scores of two diffusion models at time $t$ does not give the score of the noised product, because adding noise does not commute with taking products. Sampling from the composition by following the summed scores is therefore biased. With an energy parameterization, the summed energies define a valid unnormalized density at each $t$, and MCMC corrections can sample from it correctly~\citep{du2023reduce} as Figure~\ref{fig:rrr_samplers} shows.

\paragraph{Guided Sampling as a Product of Distributions.} The EBM view also explains why guided sampling works. Classifier guidance~\citep{dhariwal2021diffusion} draws class-conditional samples from an unconditional diffusion model by adding the gradient of a classifier $p_t(y\mid\vx)$ trained on noisy inputs, scaled by a guidance weight $w$,
\begin{equation}\label{eqn:cg}
    \nabla_\vx\log p_t(\vx) + w\,\nabla_\vx\log p_t(y\mid\vx) = -\nabla_\vx\sbr{E_\theta(\vx,t) + w\,E_\omega(\vx,y,t)}, \qquad E_\omega(\vx,y,t) = -\log p_t(y\mid\vx).
\end{equation}
The classifier is the conditional EBM of~\Secref{sec:disc_model}, and the right-hand side is the gradient of a sum of two energies. By~\Secref{sec:compose}, a sum of energies is a product of distributions, so~\Eqref{eqn:cg} is 
\begin{equation}\label{eqn:cg_poe}
    \tilde p_t(\vx\mid y) \propto p_t(\vx)\,p_t(y\mid\vx)^{w}.
\end{equation}
For $w=1$, this is the posterior $p_t(\vx\mid y)$ by Bayes' rule; for $w>1$, the classifier expert is sharpened, which trades diversity for fidelity. Classifier-free guidance~\citep{ho2021classifier} trains a conditional and an unconditional model together and combines their outputs as $(1+w)\,\bm{\epsilon}_\theta(\vx,y,t) - w\,\bm{\epsilon}_\theta(\vx,t)$. By~\Eqref{eqn:diff_param}, this corresponds to the energy $(1+w)\,E_\theta(\vx\mid y,t) - w\,E_\theta(\vx,t)$, that is, to the distribution
\begin{equation}\label{eqn:cfg_poe}
    \tilde p_t(\vx\mid y) \propto \frac{p_t(\vx\mid y)^{1+w}}{p_t(\vx)^{w}} \propto p_t(\vx)\,p_t(y\mid\vx)^{1+w},
\end{equation}
where the second step uses Bayes' rule. Classifier-free guidance is therefore classifier guidance with an implicit classifier, defined by the ratio of the two models, and the negative weight on the unconditional model is a negated expert, as in the logical \texttt{NOT} of~\Secref{sec:compose}.

In both cases, the EBM view gives guidance a precise meaning: at each noise level $t$, the guided score is the score of a product of distributions, and Langevin dynamics~\Eqref{eq:langevin} on the summed energy samples, asymptotically or with an MH correction, from~\Eqref{eqn:cg_poe} or~\Eqref{eqn:cfg_poe}. The same product combines a large pretrained diffusion model with a small domain-specific one through their scores alone, which adapts black-box text-to-video models without fine-tuning~\citep{yang2024probabilistic}. It also shows the limit of the usual implementation. Guided diffusion samplers plug the summed score into the reverse process, which assumes that the guided distributions at different $t$ are noised versions of one another. They are not, for the same reason as for composition above: adding noise does not commute with taking products. Guided reverse diffusion therefore only approximately samples the product distribution, and running Langevin or MCMC steps on the summed energy at each $t$, corrected by Metropolis--Hastings when the energy is available, removes this bias~\citep{du2023reduce}.

\subsection{Energy-Inspired Models}\label{sec:eim}\label{sec:hopfield}

Diffusion models in~\Secref{sec:diffusion} can be read as EBMs with a sampler-friendly parameterization: the potential is learned through its gradients at every noise level, which is exactly what the reverse-time sampler uses. Energy-inspired models~(EIMs)~\citep{lawson2019energy}, proposed before this view of diffusion models became common, share the same spirit from a different direction: they focus on the sampling procedure of the EBM during learning. A general EBM is sampled with an approximate procedure, such as MCMC or importance sampling, so the samples we actually use come from the sampler, not from the model $p_\theta$, and training the energy as if the samples were exact creates a mismatch between the model and the inference procedure. EIMs remove the mismatch by taking the distribution induced by the sampler, which uses a learned energy, as the model itself, and maximizing its likelihood directly. In terms of the two parts of~\Secref{sec:gen_model}, an EIM keeps the energy but defines the distribution through the sampler, so that sampling is exact by construction.

\paragraph{Self-Normalized Importance Sampling.} Let $\pi(\vx)$ be a tractable proposal, for example a Gaussian or a VAE, so that the target EBM is $p_\theta(\vx)\propto\pi(\vx)\exp\rbr{-E_\theta(\vx)}$. The simplest EIM draws $K$ candidates $\vx_1,\ldots,\vx_K\sim\pi$ and returns $\vx_i$ with probability proportional to $\exp\rbr{-E_\theta(\vx_i)}$. The likelihood of this sampler is intractable, because it marginalizes over the candidates, but treating the candidates as latent variables gives the evidence lower bound
\begin{equation}\label{eqn:eim_snis}
    \log q_\theta^{\text{SNIS}}(\vx) \ge \EE_{\vx_{2:K}\sim\pi}\sbr{\log\frac{\pi(\vx)\exp\rbr{-E_\theta(\vx)}}{\frac{1}{K}\rbr{\exp\rbr{-E_\theta(\vx)} + \sum_{j=2}^{K}\exp\rbr{-E_\theta(\vx_j)}}}}.
\end{equation}
As $K\to\infty$, the denominator converges to $Z_\theta$ and $q_\theta^{\text{SNIS}}$ to the EBM $p_\theta$; for finite $K$, the model interpolates between the proposal $\pi$ and the EBM. Up to terms that do not depend on the energy, \Eqref{eqn:eim_snis} is the ranking NCE objective~\Eqref{eqn:lnce_rank} with the proposal as the noise distribution, $p_n = \pi$. The EIM view therefore gives ranking NCE a new reading: it is a lower bound on the likelihood of a well-defined model, the SNIS sampler, whether or not the data distribution lies in the model family, and it provides a principled objective for learning the noise distribution together with the energy, which NCE alone does not. A truncated rejection sampler, which accepts proposals one at a time with probability $\sigmoid\rbr{-E_\theta(\vx)}$, gives another EIM and recovers the generative process of learned accept/reject sampling~\citep{bauer2019resampled}.

\paragraph{Hamiltonian Importance Sampling.} Importance sampling scales poorly with dimension, because all candidates come from the proposal. Hamiltonian importance sampling~\citep{neal2005hamiltonian} instead moves a single sample: it draws $\vx^{(0)}\sim\pi$ and a momentum $\vv^{(0)}\sim\Ncal\rbr{0,I}$, and runs leapfrog steps~\Eqref{eq:leapfrog} on the energy $E_\theta$ with a learned schedule that rescales the momentum at each step. When the rescaling factors multiply to one, the map from $(\vx^{(0)},\vv^{(0)})$ to $(\vx^{(T)},\vv^{(T)})$ preserves volume, so the joint density of the endpoint equals that of the starting point. Writing $\Phi_\theta$ for this map, which depends on the energy through its gradient, the likelihood of a data point $\vx = \vx^{(T)}$ is bounded by treating the final momentum as a latent variable,
\begin{equation}\label{eqn:eim_his}
    \log q_\theta^{\text{HIS}}(\vx) \ge \EE_{\vv^{(T)}\sim r(\cdot\mid\vx)}\sbr{\log\pi(\vx^{(0)}) + \log\Ncal\rbr{\vv^{(0)}; 0, I} - \log r(\vv^{(T)}\mid\vx)}, \qquad (\vx^{(0)},\vv^{(0)}) = \Phi_\theta^{-1}(\vx,\vv^{(T)}),
\end{equation}
where $r(\vv^{(T)}\mid\vx)$ is a variational distribution over the final momentum and $(\vx^{(0)},\vv^{(0)})$ is obtained by running the dynamics backward from $(\vx,\vv^{(T)})$. The bound is maximized jointly over the energy, the proposal, the step size and schedule, and $r$, by backpropagating through the trajectory. The proposal provides a coarse sample, and the gradient of the energy refines it, which helps when the proposal is poorly matched to the data. This construction is closely related to normalizing flows~(\Secref{sec:flow}), as a volume-preserving flow driven by the energy, and to the dynamics-based samplers of adversarial training~\citep{dai2019exponential}.

Hamiltonian importance sampling also resembles a diffusion model~(\Secref{sec:diffusion}): both start from a simple distribution and move the sample over several steps guided by the gradient of an energy. The difference is in learning. Hamiltonian importance sampling is trained by maximizing the variational lower bound~\Eqref{eqn:eim_his} on the likelihood of the sampler's output, which requires simulating the whole trajectory and backpropagating through it. A diffusion model never simulates its sampler during training: it fixes a forward noising process and fits the reverse process to it one noise level at a time, by the denoising regression~\Eqref{eqn:diff_loss}.

\paragraph{Modern Hopfield Networks.} A related line of work uses an energy to design the update of a network rather than its distribution. The modern Hopfield network~\citep{ramsauer2021hopfield}, which builds on classical Hopfield networks~\citep{hopfield1982neural} and dense associative memories~\citep{krotov2016dense,demircigil2017model}, stores patterns $\vx_1,\ldots,\vx_N$ as the columns of $X$ and assigns a state $\bm{\xi}$ the energy
\begin{equation}\label{eqn:hopfield_energy}
    E(\bm{\xi}) = -\frac{1}{\beta}\log\sum_{i=1}^{N}\exp\rbr{\beta\,\vx_i^\top\bm{\xi}} + \frac{1}{2}\bm{\xi}^\top\bm{\xi} + \text{const}.
\end{equation}
The concave-convex procedure minimizes this energy with the update $\bm{\xi}^{\text{new}} = X\softmax\rbr{\beta X^\top\bm{\xi}}$, which converges, stores exponentially many patterns, and typically retrieves a well-separated pattern in one step. The softmax is a Gibbs distribution over the stored patterns, and the update replaces the state by the expected pattern under it. With queries $Q$, keys $K$, values $V$ and $\beta = 1/\sqrt{d_k}$, one update is exactly the attention $\softmax\rbr{QK^\top/\sqrt{d_k}}V$ of transformers~\citep{ramsauer2021hopfield}, so an attention layer performs one step of energy minimization. Iterating such an update to its fixed point leads to the deep equilibrium models of~\Secref{sec:deq}.

\subsection{Deep Equilibrium Models}\label{sec:deq}

Inference in an EBM is often energy minimization, carried out by iterating a fixed update until it stops changing. Deep equilibrium models~(DEQs)~\citep{bai2019deep} take this iteration as the definition of a network layer: the output is the point where the update no longer changes it. Used as a generator, a DEQ can be viewed as an energy-inspired model~(\Secref{sec:eim}) whose sampler is a deterministic fixed-point solver: draw noise $\vz\sim p_0$, iterate the update to its fixed point, and return it. When the update is a gradient step on an energy, this sampler is energy minimization started from the noise.

\paragraph{Fixed-Point Layers.} Given an input $\vx$ and a layer $f_\theta$, a DEQ defines its output as the fixed point
\begin{equation}\label{eqn:deq}
    \vz^\star = f_\theta\rbr{\vz^\star, \vx}.
\end{equation}
This corresponds to applying the same layer infinitely many times, a weight-tied network of infinite depth, but the fixed point is found directly by a root-finding method, such as Broyden's method or Anderson acceleration, applied to $g_\theta(\vz) = f_\theta(\vz,\vx) - \vz$. Training does not backpropagate through the solver's iterations. By the implicit function theorem, the gradient of a loss $\ell(\vz^\star)$ is
\begin{equation}\label{eqn:deq_grad}
    \frac{\partial\ell}{\partial\theta} = \frac{\partial\ell}{\partial\vz^\star}\rbr{I - \frac{\partial f_\theta(\vz^\star,\vx)}{\partial\vz^\star}}^{-1}\frac{\partial f_\theta(\vz^\star,\vx)}{\partial\theta},
\end{equation}
which depends only on the fixed point, so memory does not grow with the number of solver steps. The vector--inverse-Jacobian product is itself computed by solving a linear fixed-point equation.

\paragraph{Learning DEQs.} How a DEQ is trained depends on whether a target is available for its fixed point. In the conditional setting, each input $\vx$ comes with a target output $\vy$, and learning is easy: a task loss is placed on the fixed point and differentiated through it,
\begin{equation}\label{eqn:deq_loss}
    \mathcal{L}(\theta) = \EE_{(\vx,\vy)\sim\mathcal{D}}\sbr{\ell\rbr{h_\theta\rbr{\vz^\star_\theta(\vx)}, \vy}}, \qquad \vz^\star_\theta(\vx) = f_\theta\rbr{\vz^\star_\theta(\vx), \vx},
\end{equation}
where $h_\theta$ is an output head and $\ell$ is, for example, the cross-entropy for classification or the squared error for regression. The exact gradient is the implicit gradient~\Eqref{eqn:deq_grad}~\citep{bai2019deep}; cheaper approximations drop or truncate the inverse-Jacobian term, such as Jacobian-free backpropagation~\citep{fung2022jfb} and phantom gradients~\citep{geng2021training}, and penalizing the Jacobian of the update keeps the fixed point stable during training~\citep{bai2021stabilizing}. Energy minimization trained by regression onto the solution, as in IREM~\citep{du2022learning}, falls in this setting as well.

The unconditional setting, where the goal is to learn a distribution and no output is paired with the noise, is harder, and existing methods first turn it into a problem that has a target or a likelihood: distillation from a pretrained diffusion model, which supplies noise--sample pairs and reduces generation to regression~\citep{geng2023one}; an invertible fixed-point map, which gives an exact likelihood as in normalizing flows~\citep{lu2021implicit}; a fixed-point layer inside the denoiser of a diffusion model, trained from scratch with the denoising loss of~\Secref{sec:diffusion}~\citep{bai2024fixed}; and Equilibrium Matching, which learns a time-independent gradient field whose descent from noise reaches the data~\citep{wang2025equilibrium}.

Equilibrium Matching is the closest to a DEQ trained without a teacher, since its sampler is exactly the fixed-point iteration of~\Eqref{eqn:deq} with the energy-descent layer $f_\theta(\vz) = \vz - \eta\nabla_\vz E_\theta(\vz)$. As introduced in~\Secref{sec:learning}, it corrupts data as $\vx_\gamma = (1-\gamma)\vx + \gamma\bm{\epsilon}$ with $\gamma\sim\mathcal{U}[0,1]$ and $\bm{\epsilon}\sim\Ncal\rbr{0,I}$, and regresses the gradient of the energy at $\vx_\gamma$ onto $(\bm{\epsilon}-\vx)\,c(\gamma)$,~\Eqref{eq:eqm_regression}. Two choices turn this regression into a fixed-point model. First, the network does not receive $\gamma$, so the minimizer is the single field $\nabla E^\star(\vz) = \EE\sbr{(\bm{\epsilon}-\vx)\,c(\gamma)\mid\vx_\gamma=\vz}$, averaged over all noise levels that could have produced $\vz$, rather than a family indexed by time as in diffusion. Second, the magnitude satisfies $c(0)=0$, for example $c(\gamma)=\gamma$, so the target vanishes at the data, and data points are equilibria of the field. This holds only approximately: since the field averages over noise levels, a data point is an exact equilibrium only if the corrupted samples that land on it come almost entirely from $\gamma\approx0$. In high dimension, the noise level can be inferred from $\vx_\gamma$ itself, because the norm of the added noise concentrates, so the approximation is accurate, but it can be loose in low dimension. Samples are generated by iterating
\begin{equation}\label{eqn:eqm_sample}
    \vz_{k+1} = \vz_k - \eta\,\nabla_\vz E_\theta(\vz_k), \qquad \vz_0\sim\Ncal\rbr{0,I},
\end{equation}
until the gradient is small, that is, until a fixed point is reached; the step size and the number of steps can be adapted at sampling time, and accelerated methods such as Nesterov's can be used. Training needs only data and noise, with no pretrained model and no unrolled trajectory.

\paragraph{Connection to EBMs.} Choose the layer to be one step of gradient descent on an energy, $f_\theta(\vz,\vx) = \vz - \eta\nabla_\vz E_\theta(\vx,\vz)$. Its fixed points are exactly the stationary points of the energy, $\nabla_\vz E_\theta(\vx,\vz^\star) = 0$, so the forward pass of the DEQ is energy minimization, as in~\Secref{sec:inference}, and~\Eqref{eqn:deq_grad} differentiates through the minimizer without unrolling it. This is the exact counterpart of the unrolled training of end-to-end SPENs in~\Secref{sec:spen} and of IREM in~\Secref{sec:learning}, which backpropagate through a fixed number of minimization steps. The converse does not hold: a general layer $f_\theta$ is not the gradient of any scalar function, because the Jacobian of a gradient field is symmetric while that of a general layer is not. A general DEQ therefore computes an equilibrium that is not the minimum of any energy, and its fixed point may not exist or be unique. Monotone operator equilibrium networks~\citep{winston2020monotone} restrict the layer so that a unique fixed point is guaranteed, and the modern Hopfield networks of~\Secref{sec:hopfield} are an instance in which the update does descend an explicit energy.

\subsection{Autoregressive Models as Approximate Ancestral Samplers}\label{sec:ar}

Autoregressive models can be viewed as another reparameterization of an EBM from the sampler's point of view, in the spirit of the energy-inspired models of~\Secref{sec:eim}. Instead of parameterizing a global energy and then designing a sampler for it, the parameterization follows ancestral sampling: each step models the conditional distribution of one variable given the ones already generated, and the joint distribution is the one induced by a single sweep in a fixed order.

\paragraph{Fixed-Order Factorization of an EBM.} For $\vx = (x_1,\ldots,x_d)$ with each $x_i$ in a finite vocabulary $\mathcal{V}$, any EBM~\Eqref{eq:ebm} factorizes exactly in a fixed order by the chain rule,
\begin{equation}\label{eqn:ebm_chain}
    p_\theta(\vx) = \prod_{i=1}^{d} p_\theta(x_i\mid\vx_{<i}), \qquad
    p_\theta(x_i\mid\vx_{<i}) = \frac{\sum_{\vx_{>i}}\exp\rbr{-E_\theta(\vx)}}{\sum_{x_i,\,\vx_{>i}}\exp\rbr{-E_\theta(\vx)}}.
\end{equation}
Drawing $x_1$, then $x_2$ given $x_1$, and so on, is ancestral sampling: each variable is drawn given the ones already visited, with the unvisited ones marginalized out, and this single pass returns an exact sample. Unlike a Gibbs sweep~(\Secref{sec:inference}), which conditions on the current values of all other variables and revisits them, it draws each variable once. For a general EBM, however, the sum over $\vx_{>i}$ is as intractable as $Z_\theta$ itself. Autoregressive models reparameterize exactly this sum: they model it directly with a network $e_\theta$,
\begin{equation}\label{eqn:ar_reparam}
    \sum_{\vx_{>i}}\exp\rbr{-E_\theta(\vx)} \;\approx\; \exp\rbr{-e_\theta(x_i;\vx_{<i})},
\end{equation}
which turns the exact conditional in~\Eqref{eqn:ebm_chain} into
\begin{equation}\label{eqn:ar_step}
    p_\theta(x_i = v\mid\vx_{<i}) = \frac{\exp\rbr{-e_\theta(v;\vx_{<i})}}{\sum_{v'\in\mathcal{V}}\exp\rbr{-e_\theta(v';\vx_{<i})}}.
\end{equation}
The marginalization over the unvisited variables is never computed; it is absorbed into $e_\theta$, which plays the role of the free energy of $\vx_{\le i}$ after $\vx_{>i}$ is summed out. What remains is a normalization over the $\abr{\mathcal{V}}$ values of a single variable, with $-e_\theta(\cdot;\vx_{<i})$ as the logits, \ie, the multi-class logistic regression of~\Secref{sec:disc_model}. An autoregressive model is therefore ancestral sampling of~\Eqref{eqn:ebm_chain} with its intractable conditionals replaced by learned ones. NADE~\citep{uria2016neural} gives a concrete instance: its forward pass is a single pass of mean-field inference that approximates the conditionals $p(x_i\mid\vx_{<i})$ of an RBM, with the unvisited variables initialized to zero rather than summed out, and its weight sharing is inherited from the RBM.

\paragraph{Exact Learning.} The same factorization makes learning exact: maximum likelihood decomposes into a sum of per-step cross-entropies,
\begin{equation}\label{eqn:ar_mle}
    \mathcal{L}(\theta) = -\EE_{\vx\sim p_d}\sbr{\sum_{i=1}^{d}\log p_\theta(x_i\mid\vx_{<i})},
\end{equation}
each of which is the exact maximum likelihood of a conditional EBM~\Eqref{eqn:ar_step}, and all of which can be computed in parallel from a data sequence.

\paragraph{Relation to Pseudolikelihood.} Pseudolikelihood~\Eqref{eqn:lpseudo} is the closest learning method for EBMs, and comparing the two shows what the reparameterization changes. Both fit single-variable conditionals that are softmaxes over $\mathcal{V}$, and both are tractable for the same reason: only the normalization of one variable over $\abr{\mathcal{V}}$ values is needed. They differ in the conditioning set (Figure~\ref{fig:conditioning}). Pseudolikelihood fits the full conditionals $p_\theta(x_j\mid\vx_{-j})$, the steps of the Gibbs sampler of~\Secref{sec:inference}, which can be computed from $E_\theta$ in closed form because nothing is summed out. An autoregressive model fits the chain-rule conditionals $p_\theta(x_i\mid\vx_{<i})$, the steps of ancestral sampling in~\Eqref{eqn:ebm_chain}, which require summing out $\vx_{>i}$ and are therefore reparameterized by $e_\theta$. This single difference leads to opposite trade-offs:

\begin{figure}[t]
    \centering
    \includegraphics[width=\textwidth]{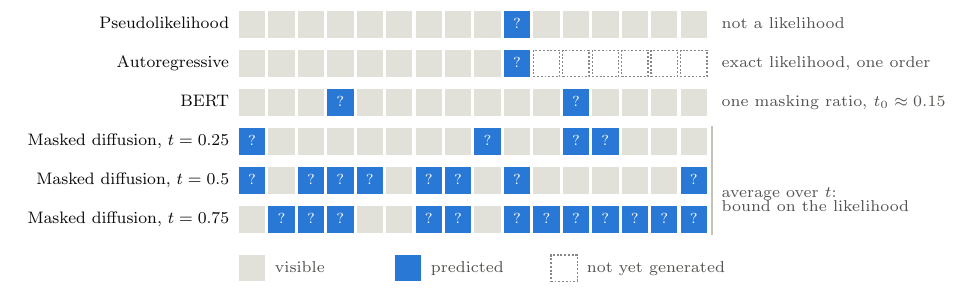}
    \caption{\textbf{Illustration of different conditioning sets.}  Pseudolikelihood predicts one token from all the others; an autoregressive model predicts the next token from the prefix, and the remaining tokens are not yet generated; BERT predicts a fixed fraction of masked tokens; masked diffusion predicts the masked tokens at every masking level $t$, and averaging over $t$ gives a bound on the likelihood.}
    \label{fig:conditioning}
\end{figure}

\begin{itemize}
    \item \textbf{Objective.} By the chain rule,~\Eqref{eqn:ar_mle} is exactly the negative $\log$-likelihood, so autoregressive training is exact MLE. The sum of full conditionals in~\Eqref{eqn:lpseudo} is not a likelihood. In the ordering view of~\Secref{sec:learning}, pseudolikelihood keeps only the last term of every ordering, whereas an autoregressive model keeps all the terms of one fixed ordering.
    \item \textbf{Energy.} Pseudolikelihood keeps the global energy: all its conditionals are derived from a single $E_\theta$, which it estimates consistently~\citep{besag1975statistical}. An autoregressive model gives the global energy up, as discussed below.
    \item \textbf{Sampling.} The full conditionals learned by pseudolikelihood do not produce a sample in one pass, and still require many Gibbs sweeps. The chain-rule conditionals do: one ordered pass from the empty state returns an exact sample.
\end{itemize}
In the language of~\Secref{sec:unified}, pseudolikelihood corresponds to a single Gibbs step started at the data, while an autoregressive model corresponds to ancestral sampling from the empty state. Masked discrete diffusion, discussed next, lies between the two: it hides a random subset of the variables, ranging from a single one, as in pseudolikelihood, to all of them, as in the first step of an autoregressive pass.

\paragraph{Connection to Discrete Diffusion.} The order of the sweep is a choice. Averaging~\Eqref{eqn:ar_mle} over random orders gives an any-order autoregressive model~\citep{uria2014deep,hoogeboom2022autoregressive}, and, as discussed with pseudolikelihood in~\Secref{sec:learning}, the masked discrete diffusion objective~\Eqref{eqn:masked_diffusion} is equivalent to it~\citep{ou2024your}. Masked diffusion models are therefore autoregressive samplers whose order is random and which may reveal several variables per step. Revealing several variables at once, from conditionals predicted independently given the same context, trades the exactness of a sequential sweep for parallel generation, and, as noted there, learned conditionals that are parameterized directly need not be consistent with a single joint distribution, which the autoregressive factorization~\Eqref{eqn:ebm_chain} guarantees.

\paragraph{The Price of the Reparameterization.} The reparameterization~\Eqref{eqn:ar_reparam} trades the energy for the sampler. On the gain side, sampling takes a single exact pass of~\Eqref{eqn:ar_step} with no burn-in or mixing, and learning is exact maximum likelihood~\Eqref{eqn:ar_mle}, since each step only requires a normalization over $\abr{\mathcal{V}}$ values. On the cost side, $e_\theta$ is no longer tied to a global energy. In~\Eqref{eqn:ebm_chain}, all conditionals are derived from a single $E_\theta$, whereas an autoregressive model approximates each marginalized sum separately, so in general the parameters of an underlying EBM cannot be recovered from it. Moreover, summing out $\vx_{>i}$ couples variables that $E_\theta$ does not couple directly: even when $E_\theta$ is simple, such as a pairwise MRF with few parameters, the sums $\sum_{\vx_{>i}}\exp\rbr{-E_\theta(\vx)}$ are generally complex functions of the whole prefix $\vx_{<i}$, and they change with the chosen order. Approximating them by $\exp\rbr{-e_\theta}$ therefore requires a far more flexible $e_\theta$ than the energy it replaces, and the compact structure of $E_\theta$ is lost.

\paragraph{Large Language Models (LLMs).} LLMs are the setting in which this reparameterization fits best, since both of its costs are naturally mitigated. First, the order is given. Text is a sequence, and the left-to-right order in which it is written and read is a natural choice for the fixed order in~\Eqref{eqn:ebm_chain}, so no ordering has to be chosen or learned~\citep{bengio2003neural}. Second, the marginalized sums in~\Eqref{eqn:ar_reparam} are modeled by a highly flexible network. A Transformer decoder~\citep{vaswani2017attention} computes $-e_\theta(\cdot;\vx_{<i})$ as the logits over the token vocabulary from the entire prefix, which provides exactly the flexibility over the whole context that the approximation requires. With causal masking, all $d$ conditionals in~\Eqref{eqn:ar_mle} are computed in one forward pass over a training sequence, so learning remains exact maximum likelihood at scale~\citep{radford2019language,brown2020language}, and generation is ancestral sampling, one token at a time.

\section{Representation Learning}\label{sec:repr}

So far, the energy itself has been the object of interest: in~\Secref{sec:disc_model} it scores outputs for prediction, and in~\Secref{sec:gen_model} it defines the distribution to sample from. In representation learning, the energy becomes a means rather than an end. The goal is a feature map $\phi_\theta(\vx)$ that is useful for downstream tasks, and the energy is the pretext through which it is learned: $\phi_\theta$ is placed inside an energy defined on a self-supervised task, such as matching two views of the same input, predicting the masked parts of an input, or denoising a corrupted input, and the energy is trained with one of the methods of~\Secref{sec:learning}. After training, the energy is discarded and $\phi_\theta$ is kept. From this view, self-supervised methods differ mainly in two choices, the structure of the energy and the learning method, and the learning methods they use are precisely those that avoid sampling from the model.

\subsection{Contrastive Representation Learning}\label{sec:contrastive}

Following~\citet{dai2026spectral}, contrastive methods can be read as learning a single conditional EBM between two related variables. Let $(\vx,\vy)\sim p(\vx,\vy)$ be a positive pair, such as two augmented views of the same image, or an image and its caption, and let $\phi_\theta$ and $\psi_\theta$ be the encoders of the two sides, which share parameters when $\vx$ and $\vy$ come from the same modality. The model is
\begin{equation}\label{eqn:repr_ebm}
\begin{aligned}
    &\frac{p_\theta(\vx,\vy)}{p(\vx)\,p(\vy)} \propto \exp\rbr{\phi_\theta(\vx)^\top\psi_\theta(\vy)}, \\
    &p_\theta(\vy\mid\vx) = \frac{p(\vy)\exp\rbr{\phi_\theta(\vx)^\top\psi_\theta(\vy)}}{Z_\theta(\vx)}, \qquad
    Z_\theta(\vx) = \EE_{p(\vy)}\sbr{\exp\rbr{\phi_\theta(\vx)^\top\psi_\theta(\vy)}}, \\
    &p_\theta(\vx\mid\vy) = \frac{p(\vx)\exp\rbr{\phi_\theta(\vx)^\top\psi_\theta(\vy)}}{Z_\theta(\vy)}, \qquad
    Z_\theta(\vy) = \EE_{p(\vx)}\sbr{\exp\rbr{\phi_\theta(\vx)^\top\psi_\theta(\vy)}},
\end{aligned}
\end{equation}
where the two conditionals have their own partition functions. This is an EBM with the bilinear energy $E_\theta(\vx,\vy) = -\phi_\theta(\vx)^\top\psi_\theta(\vy)$ relative to the base measure $p(\vy)$, and it models the pointwise mutual information between the two sides. The representation lives inside the energy, and~\citet{dai2026spectral} show that it is sufficient for downstream prediction. The structure of~\Eqref{eqn:repr_ebm} makes it convenient to learn: the base measure is the marginal $p(\vy)$, which is exactly the distribution of the other samples in a batch, so the natural noise distribution for NCE is $p_n = p(\vy)$, and the $\log p_n$ terms in~\Eqref{eqn:nce_posterior} and~\Eqref{eqn:nce_rank_posterior} cancel against the base measure, leaving the inner product as the logit. The remaining difficulty is the partition functions $Z_\theta(\vx)$ and $Z_\theta(\vy)$, which are intractable and differ for every input, and existing methods differ mainly in how they handle them.

\paragraph{SimCLR: Ranking NCE within One Modality.} SimCLR~\citep{chen2020simple} takes $\vx$ and $\vy$ to be two augmentations of the same input, with a shared encoder $\phi_\theta=\psi_\theta$. Applying ranking NCE~\Eqref{eqn:lnce_rank} to $p_\theta(\vy\mid\vx)$, with the other $K$ samples in the batch as the noise, gives
\begin{equation}\label{eqn:simclr}
    \mathcal{L}_{\text{SimCLR}}(\theta) = -\EE\sbr{\phi_\theta(\vx)^\top\phi_\theta(\vy_0) - \log\sum_{k=0}^{K}\exp\rbr{\phi_\theta(\vx)^\top\phi_\theta(\vy_k)}},
\end{equation}
which is the InfoNCE objective~\citep{oord2018representation}. Normalizing the features, $\phi_\theta(\vx)/\nbr{\phi_\theta(\vx)}_2$, and dividing the inner product by a temperature recovers SimCLR exactly~\citep{dai2026spectral}. Ranking NCE needs no partition function, since $Z_\theta(\vx)$ cancels within the softmax, but its gradient~\Eqref{eqn:nce_rank_grad} is a self-normalized importance sampling estimate whose bias decreases only as $K$ grows, which is why these methods depend on large batches.

\paragraph{CLIP: Ranking NCE across Modalities.} CLIP~\citep{radford2021learning} applies the same estimator to an image $\vx$ and its caption $\vy$ with separate encoders, and fits both conditionals of~\Eqref{eqn:repr_ebm},
\begin{equation}\label{eqn:clip}
\begin{aligned}
    \mathcal{L}_{\text{CLIP}}(\theta) = &-\EE\sbr{\phi_\theta(\vx_0)^\top\psi_\theta(\vy_0) - \log\sum_{k=0}^{K}\exp\rbr{\phi_\theta(\vx_0)^\top\psi_\theta(\vy_k)}} \\
    &-\EE\sbr{\phi_\theta(\vx_0)^\top\psi_\theta(\vy_0) - \log\sum_{k=0}^{K}\exp\rbr{\phi_\theta(\vx_k)^\top\psi_\theta(\vy_0)}},
\end{aligned}
\end{equation}
where the first term is ranking NCE for $p_\theta(\vy\mid\vx_0)$, which identifies the caption $\vy_0$ of the image $\vx_0$ among the $K+1$ captions in the batch with the other captions drawn from the noise $p(\vy)$, and the second term is ranking NCE for $p_\theta(\vx\mid\vy_0)$, which identifies the image $\vx_0$ of the caption $\vy_0$ among the $K+1$ images with the other images drawn from the noise $p(\vx)$. It inherits the large-batch requirement of ranking NCE.

\begin{figure}[t]
    \centering
    \includegraphics[width=\textwidth]{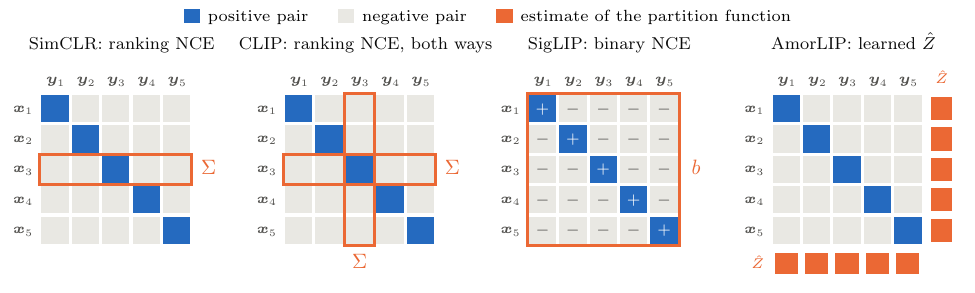}
    \vspace{-15pt}
    \caption{\textbf{Contrastive methods as estimators of the same EBM.} Each panel shows the scores $\phi_\theta(\vx_i)^\top\psi_\theta(\vy_j)$ of a batch for the bilinear EBM~\Eqref{eqn:repr_ebm}, with the positive pairs on the diagonal. The methods differ in what estimates the partition functions: SimCLR sums each row~($\Sigma$), and CLIP also each column; SigLIP classifies every pair separately, as positive or negative, with one learned bias~$b$ shared by all pairs; AmorLIP predicts the partition functions~($\hat Z$) from the representations instead of summing over the batch.}
    \label{fig:similarity}
    \vspace{-10pt}
\end{figure}

\paragraph{SigLIP: Binary NCE across Modalities.} SigLIP~\citep{zhai2023sigmoid} replaces the softmax by a sigmoid on each pair. This is binary NCE~\Eqref{eqn:lnce_bin} with positive pairs drawn from $p(\vx,\vy)$ as data and mismatched pairs drawn from $p(\vx)p(\vy)$ as noise,
\begin{equation}\label{eqn:siglip}
    \mathcal{L}_{\text{SigLIP}}(\theta, b) = -\EE_{p(\vx,\vy)}\sbr{\log\sigmoid\rbr{\phi_\theta(\vx)^\top\psi_\theta(\vy) + b}} - \EE_{p(\vx)p(\vy)}\sbr{\log\rbr{1-\sigmoid\rbr{\phi_\theta(\vx)^\top\psi_\theta(\vy) + b}}},
\end{equation}
SigLIP thus learns the EBM~\Eqref{eqn:repr_ebm} through its joint distribution, which has a single partition function,
\begin{equation}\label{eqn:siglip_joint}
    p_\theta(\vx,\vy) = \frac{p(\vx)\,p(\vy)\exp\rbr{\phi_\theta(\vx)^\top\psi_\theta(\vy)}}{Z_\theta}, \qquad
    Z_\theta = \EE_{p(\vx)p(\vy)}\sbr{\exp\rbr{\phi_\theta(\vx)^\top\psi_\theta(\vy)}}.
\end{equation} The bias $b = -\log Z_\theta$ is the $\log$-partition constant $-c$ of binary NCE in~\Eqref{eqn:nce_density}, so the objective is consistent for the joint model, and the conditionals with their partition functions $Z_\theta(\vx)$ and $Z_\theta(\vy)$ follow from it without being estimated. Each pair contributes a separate term, so the gradient is unbiased for any batch size, which is the practical advantage of SigLIP over the ranking NCE of CLIP.

\paragraph{AmorLIP: Maximum Likelihood with Amortized Partition Functions.} Instead of an NCE surrogate, one can fit both conditionals of~\Eqref{eqn:repr_ebm} by maximum likelihood,
\begin{equation}\label{eqn:clip_mle}
    \max_\theta\; 2\,\EE_{p(\vx,\vy)}\sbr{\phi_\theta(\vx)^\top\psi_\theta(\vy)} - \EE_{p(\vx)}\sbr{\log Z_\theta(\vx)} - \EE_{p(\vy)}\sbr{\log Z_\theta(\vy)},
\end{equation}
which has the same gradient as the surrogate objective
\begin{equation}\label{eqn:clip_mle_grad}
    \max_\theta\; 2\,\EE_{p(\vx,\vy)}\sbr{\phi_\theta(\vx)^\top\psi_\theta(\vy)} - \EE_{p(\vx)p(\vy)}\sbr{\frac{\exp\rbr{\phi_\theta(\vx)^\top\psi_\theta(\vy)}}{\texttt{sg}\rbr{Z_\theta(\vx)}}} - \EE_{p(\vx)p(\vy)}\sbr{\frac{\exp\rbr{\phi_\theta(\vx)^\top\psi_\theta(\vy)}}{\texttt{sg}\rbr{Z_\theta(\vy)}}},
\end{equation}
with $\texttt{sg}$ the stop-gradient operator. Given the values of $Z_\theta(\vx)$ and $Z_\theta(\vy)$, every term is an expectation that can be estimated without bias from small batches. The difficulty is that these partition functions are intractable and change with every input. Ranking NCE estimates them by the batch sum, which is biased for small batches, and binary NCE avoids them by fitting the joint model with a single constant. AmorLIP~\citep{sun2025amorlip} instead amortizes them: since $Z_\theta(\vx) = \EE_{p(\vy)}\sbr{\exp\rbr{\phi_\theta(\vx)^\top\psi_\theta(\vy)}}$ depends on $\vx$ only through $\phi_\theta(\vx)$, it can be fitted by a lightweight network on top of the learned representation, trained with a variational objective alongside the encoders, and plugged into~\Eqref{eqn:clip_mle_grad}. This keeps the consistency of maximum likelihood while removing the dependence on large batches.

In summary, SimCLR, CLIP, SigLIP and AmorLIP learn the same EBM~\Eqref{eqn:repr_ebm} and differ only in the learning method of~\Secref{sec:learning}: ranking NCE, which cancels the partition function within the batch; binary NCE, which fits the joint model with a single constant; and maximum likelihood, which estimates it with an amortized model (Figure~\ref{fig:similarity}). The trade-offs among them are those of the estimators, between the batch size, the consistency of the estimator, and the cost of modeling the partition function.

\paragraph{Fixed-Point Learning of EBM Representation.} The methods above fit both sides of the energy~\Eqref{eqn:repr_ebm} jointly. \citet{dai2026spectral} show that the representation can also be learned by a fixed-point iteration, in the spirit of power iteration for spectral decomposition, in which one side of the energy is frozen at the previous iterate and only the other side is updated. With a shared encoder as in SimCLR, the $\ell$-th iteration fits the conditional EBM $p_{\theta}(\vy\mid\vx)\propto p(\vy)\exp\rbr{\phi_\theta(\vx)^\top\phi^{(\ell)}(\vy)}$, in which the representation $\phi^{(\ell)}$ of the target is fixed, by ranking NCE,
\begin{equation}\label{eqn:moco_iter}
    \phi^{(\ell+1)} = \argmin_{\phi}\; -\EE\sbr{\phi(\vx)^\top\phi^{(\ell)}(\vy_0) - \log\sum_{k=0}^{K}\exp\rbr{\phi(\vx)^\top\phi^{(\ell)}(\vy_k)}},
\end{equation}
and a fixed point $\phi^{(\ell+1)} = \phi^{(\ell)}$ is a representation consistent with the EBM. MoCo~\citep{he2020momentum} is a practical implementation of this iteration. It approximates the $\argmin$ by a few stochastic gradient steps on the query encoder $\phi$, maintains the frozen key encoder $\phi^{(\ell)}$ as an exponential moving average of the query encoder, and draws the negatives $\vy_k$ from a queue of keys computed in recent iterations. Since the keys are produced by a slowly changing $\phi^{(\ell)}$ that receives no gradient, the stale keys in the queue remain consistent with the current iterate, which provides many negatives without large batches.

\subsection{Word2vec}\label{sec:word2vec}

Word2vec~\citep{mikolov2013efficient,mikolov2013distributed} is an early and simple instance of the same construction, in which the encoders are lookup tables. It comes in two variants, which correspond to~\Secref{sec:contrastive} and~\Secref{sec:bert}, respectively.

\paragraph{Skip-Gram: Binary NCE for a Pairwise EBM.} Let $(w,c)$ be a word and a word in its context window. Skip-gram models their co-occurrence with the bilinear EBM~\Eqref{eqn:repr_ebm}~\citep{dai2026spectral},
\begin{equation}\label{eqn:w2v_ebm}
    \frac{p_\theta(w,c)}{p(w)\,p_n(c)} \propto \exp\rbr{\bm{u}_w^\top\bm{v}_c},
\end{equation}
where $\bm{u}_w$ and $\bm{v}_c$ are the embeddings of $w$ as a word and of $c$ as a context, and $p_n(c)$ is the noise distribution over contexts used below; with $p_n(c) = p(c)$, the ratio is the pointwise mutual information. Skip-gram with negative sampling fits~\Eqref{eqn:w2v_ebm} by binary NCE~\Eqref{eqn:lnce_bin}, with observed pairs as data and $k$ pairs drawn from $p(w)\,p_n(c)$ as noise,
\begin{equation}\label{eqn:sgns}
    \mathcal{L}_{\text{SGNS}}(\theta) = -\EE_{p(w,c)}\sbr{\log\sigmoid\rbr{\bm{u}_w^\top\bm{v}_c}} - k\,\EE_{p(w)\,p_n(c)}\sbr{\log\sigmoid\rbr{-\bm{u}_w^\top\bm{v}_c}},
\end{equation}
where the noise distribution over contexts $p_n(c)$ is the unigram distribution raised to the power $3/4$ and renormalized. Compared with~\Eqref{eqn:nce_posterior}, the logit is the inner product alone, so the bias $b=-c-\log\nu$ is fixed at zero rather than learned as in SigLIP. With $\nu=k$, the optimal classifier then gives
\begin{equation}\label{eqn:w2v_pmi}
    \bm{u}_w^{\star\top}\bm{v}^\star_c = \log\frac{p(w,c)}{p(w)\,p_n(c)} - \log k,
\end{equation}
which is the shifted pointwise mutual information that skip-gram is known to factorize implicitly~\citep{levy2014neural}. In the EBM view, the shift $-\log k$ is the $\log$-partition constant that binary NCE would otherwise have learned.

\paragraph{CBOW: Masked Prediction of a Single Word.} The continuous bag-of-words model predicts a word from the average of its context embeddings,
\begin{equation}\label{eqn:cbow}
    p_\theta(w\mid\vx_{\text{ctx}}) = \frac{\exp\rbr{\bm{h}(\vx_{\text{ctx}})^\top\bm{u}_w}}{\sum_{w'\in\mathcal{V}}\exp\rbr{\bm{h}(\vx_{\text{ctx}})^\top\bm{u}_{w'}}}, \qquad
    \bm{h}(\vx_{\text{ctx}}) = \frac{1}{\abr{\vx_{\text{ctx}}}}\sum_{c\in\vx_{\text{ctx}}}\bm{v}_c.
\end{equation}
This is a special case of the masked prediction EBM of BERT, discussed next in~\Secref{sec:bert}, with a single hidden position and a context window as the visible tokens, and with the contextual encoder replaced by an average of embeddings. With one hidden position, it is the pseudolikelihood~\Eqref{eqn:lpseudo} of the window. Since the partition function sums over the whole vocabulary, which was costly at the time, CBOW is trained in practice with the same negative sampling or with a hierarchical softmax~\citep{mikolov2013distributed}.

\subsection{BERT}\label{sec:bert}

Masked language modeling, as in BERT~\citep{devlin2019bert}, learns an EBM of the same bilinear form as~\Eqref{eqn:repr_ebm}, now between a corrupted sequence and the clean one. Given a sequence $\vx = (x_1,\ldots,x_d)$ over the vocabulary $\mathcal{V}$, let $\tilde\vx$ be the corrupted sequence in which a random subset $M$ of the positions is replaced by \texttt{[MASK]}. BERT models the denoising distribution of the hidden tokens $\vx_M$ given $\tilde\vx$ as an EBM with the potential
\begin{equation}\label{eqn:bert_ebm}
    p_\theta(\vx_M\mid\tilde\vx) = \frac{\exp\rbr{-E_\theta(\vx_M;\tilde\vx)}}{Z_\theta(\tilde\vx)}, \qquad
    E_\theta(\vx_M;\tilde\vx) = -\sum_{j\in M}\bm{h}_{\theta,j}(\tilde\vx)^\top\bm{e}_{x_j} = -\phi_\theta(\tilde\vx)^\top\psi(\vx_M),
\end{equation}
where $\bm{h}_{\theta,j}(\tilde\vx)$ is the contextual representation that a bidirectional Transformer computes at position $j$, $\bm{e}_v$ is the embedding of token $v$, and $\phi_\theta(\tilde\vx) = (\bm{h}_{\theta,j}(\tilde\vx))_{j\in M}$ and $\psi(\vx_M) = (\bm{e}_{x_j})_{j\in M}$ stack them over the masked positions. As in~\Secref{sec:contrastive}, the representation $\phi_\theta$ of the corrupted input enters the energy through an inner product with a representation of the target. The same structure appears in the multi-modal spectral representation of~\citet[Sec.~7.2]{dai2026spectral}, where, for discrete variables such as text, the diffusion loss is replaced by the masked diffusion loss, which scores the clean tokens through an inner product with the representation of the corrupted text. Since the potential is a sum over the masked positions, the partition function factorizes,
\begin{equation}\label{eqn:bert_cond}
    Z_\theta(\tilde\vx) = \prod_{j\in M}\sum_{v\in\mathcal{V}}\exp\rbr{\bm{h}_{\theta,j}(\tilde\vx)^\top\bm{e}_v}, \qquad
    p_\theta(\vx_M\mid\tilde\vx) = \prod_{j\in M}\frac{\exp\rbr{\bm{h}_{\theta,j}(\tilde\vx)^\top\bm{e}_{x_j}}}{\sum_{v\in\mathcal{V}}\exp\rbr{\bm{h}_{\theta,j}(\tilde\vx)^\top\bm{e}_v}},
\end{equation}
so it is a product of $\abr{M}$ sums over $\abr{\mathcal{V}}$ values and can be computed exactly. Maximum likelihood of~\Eqref{eqn:bert_ebm} is therefore exact, and it is precisely the BERT objective,
\begin{equation}\label{eqn:bert}
    \mathcal{L}_{\text{BERT}}(\theta) = \EE_{\vx\sim p_d}\EE_{M\sim q_{t_0}}\sbr{-\log p_\theta\rbr{\vx_M\mid\tilde\vx}} = \EE_{\vx\sim p_d}\EE_{M\sim q_{t_0}}\sbr{-\sum_{j\in M}\log p_\theta\rbr{x_j\mid\tilde\vx}},
\end{equation}
with a fixed masking ratio $t_0$, about $15\%$ in BERT. BERT is thus a neural network generalization of word2vec, specifically of CBOW~\Eqref{eqn:cbow}: it replaces the averaged context embeddings by a deep bidirectional Transformer, and it predicts several masked tokens at once under an independence assumption on the potential, which is a sum of per-position terms, so that the masked tokens are conditionally independent given $\tilde\vx$. This assumption is also what makes the partition function~\Eqref{eqn:bert_cond} tractable. After training, the energy, \ie, the output layer over $\bm{e}_v$, is discarded and the representation $\bm{h}_\theta$ is kept.

\paragraph{Connection to Masked Discrete Diffusion and Pseudolikelihood.} Starting from BERT, its masking ratio $t_0$ connects it to the two objectives related in~\Secref{sec:learning}. Shrinking the mask to a single position, $\abr{M}=1$, turns~\Eqref{eqn:bert} into pseudolikelihood~\Eqref{eqn:lpseudo}, in which every other token is visible. Averaging~\Eqref{eqn:bert} over all ratios with weight $1/t$ turns it into the masked diffusion objective~\Eqref{eqn:masked_diffusion}, which bounds the negative $\log$-likelihood. \citet{austin2021structured} make the second connection precise: BERT is a one-step discrete diffusion model, whose transition matrix mixes the absorbing state \texttt{[MASK]} with a uniform transition, matching BERT's practice of replacing some selected tokens by \texttt{[MASK]} and others by random tokens. BERT thus lies between the two: it hides more than one position, so it trains the model to predict tokens whose neighbors are also unknown, which pseudolikelihood never does, but it keeps a single ratio, so it learns the denoiser, and its representation, at one noise level only.

\subsection{Diffusion Representations}\label{sec:diff_repr}

The methods in~\Secref{sec:contrastive} learn the bilinear EBM~\Eqref{eqn:repr_ebm} by NCE, which needs negative samples and, in its ranking form, large batches. Denoising avoids both. We first derive the classical denoising auto-encoder~\citep{vincent2008extracting,vincent2011connection} from a joint EBM over the input and a latent variable; the derived model justifies the bilinear structure of~\Eqref{eqn:repr_ebm}, with the DAE as a conditional EBM of this form. We then present the diffusion spectral representation~\citep{shribak2024diffusion,dai2026spectral} as its conditional extension, learned by score matching.

\paragraph{Denoising Auto-Encoders.} A denoising auto-encoder~(DAE) corrupts the input as $\tilde\vx = \vx + \sigma\bm{\epsilon}$. We derive it by parameterizing the distribution $p_\sigma$ of the corrupted input with a latent-variable EBM $p_\theta$, in which the latent variable $\bm{h}$ plays the role of the representation,
\begin{equation}\label{eqn:dae_lv}
\begin{aligned}
    &p_\sigma(\tilde\vx) = \int\Ncal\rbr{\tilde\vx;\vx,\sigma^2 I}\,p_d(\vx)\,d\vx, \\
    &p_\theta(\tilde\vx) = \frac{1}{Z_\theta}\sum_{\bm{h}}\exp\rbr{-E_\theta(\tilde\vx,\bm{h})} = \frac{\exp\rbr{-F_\theta(\tilde\vx)}}{Z_\theta}, \qquad
    \bm{h}_\theta(\tilde\vx) = \EE_{p_\theta(\bm{h}\mid\tilde\vx)}\sbr{\bm{h}},
\end{aligned}
\end{equation}
where $F_\theta(\tilde\vx) = -\log\sum_{\bm{h}}\exp\rbr{-E_\theta(\tilde\vx,\bm{h})}$ is the free energy, the sum becomes an integral for continuous $\bm{h}$, and the representation of $\tilde\vx$ is the posterior mean of the latent. Here $p_\sigma$ is the data distribution $p_d$ convolved with the Gaussian corruption, and $p_\theta$ is its parameterization: the latent-variable form is the modeling assumption, and learning fits $p_\theta$ to $p_\sigma$.
We fit $p_\theta$ to $p_\sigma$ by denoising score matching~\Eqref{eqn:dsm}, which avoids the partition function and needs only the score of the model, an expectation under the posterior of the latent,
\begin{equation}\label{eqn:dae_lv_score}
    -\nabla_{\tilde\vx}F_\theta(\tilde\vx) = -\EE_{p_\theta(\bm{h}\mid\tilde\vx)}\sbr{\nabla_{\tilde\vx}E_\theta(\tilde\vx,\bm{h})}.
\end{equation}
For a general joint energy, which couples the components of $\bm{h}$, this posterior, and hence both the score and the representation, require approximate inference over $\bm{h}$; a closed form needs the structure~\Eqref{eqn:harmonium} below.

\paragraph{Closed Form from Linearity in the Latent.} The posterior in~\Eqref{eqn:dae_lv_score} becomes tractable when the energy is linear in $\bm{h}$, up to terms that act on each component separately,
\begin{equation}\label{eqn:harmonium}
    E_\theta(\tilde\vx,\bm{h}) = U_\theta(\tilde\vx) - \bm{h}^\top g_\theta(\tilde\vx) + \sum_{j=1}^{m}V_j(h_j),
\end{equation}
where $V_j$ is a fixed base measure for the $j$-th component, which may be discrete or continuous. The components of $\bm{h}$ are then conditionally independent given $\tilde\vx$, and each posterior is an exponential family whose natural parameter is the feature $g_{\theta,j}(\tilde\vx)$, with $\log$-partition function $A_j(\eta) = \log\int\exp\rbr{\eta h_j - V_j(h_j)}\,dh_j$. The posterior, the representation, and the free energy are therefore available in closed form,
\begin{equation}\label{eqn:harmonium_post}
\begin{aligned}
    &p_\theta(\bm{h}\mid\tilde\vx) = \prod_{j=1}^{m}\exp\rbr{h_j\,g_{\theta,j}(\tilde\vx) - V_j(h_j) - A_j\rbr{g_{\theta,j}(\tilde\vx)}}, \\
    &h_{\theta,j}(\tilde\vx) = A_j'\rbr{g_{\theta,j}(\tilde\vx)}, \qquad
    F_\theta(\tilde\vx) = U_\theta(\tilde\vx) - \sum_{j=1}^{m}A_j\rbr{g_{\theta,j}(\tilde\vx)},
\end{aligned}
\end{equation}
and the score~\Eqref{eqn:dae_lv_score} is linear in the representation,
\begin{equation}\label{eqn:dae_score}
    -\nabla_{\tilde\vx}F_\theta(\tilde\vx) = -\nabla_{\tilde\vx}U_\theta(\tilde\vx) + \nabla_{\tilde\vx}g_\theta(\tilde\vx)^\top\bm{h}_\theta(\tilde\vx).
\end{equation}
For binary components, $V_j = 0$ on $\cbr{0,1}$, the posterior is Bernoulli, $A_j$ is the softplus, and the encoder is the sigmoid $\sigmoid\rbr{g_\theta(\tilde\vx)}$; for Gaussian components, $V_j(h_j) = \frac{1}{2}h_j^2$, the posterior is $\Ncal\rbr{g_\theta(\tilde\vx), I}$, and the encoder is $g_\theta(\tilde\vx)$ itself. This is the structure of exponential-family harmoniums~\citep{welling2004exponential}. Closed form requires linearity in $\bm{h}$, not in the input: the features $g_\theta$ may be a deep network.

\paragraph{From the Score to the DAE.} Substituting the closed-form score~\Eqref{eqn:dae_score} into denoising score matching~\Eqref{eqn:dsm} and multiplying by $2\sigma^4$ gives
\begin{equation}\label{eqn:dae_recon}
\begin{aligned}
    &\sigma^4\,\EE_{\vx\sim p_d,\,\bm{\epsilon}}\sbr{\Big\|-\nabla_{\tilde\vx}F_\theta(\tilde\vx) + \frac{\tilde\vx - \vx}{\sigma^2}\Big\|_2^2}
    = \EE_{\vx\sim p_d,\,\bm{\epsilon}}\sbr{\big\|r_\theta(\tilde\vx) - \vx\big\|_2^2}, \\
    &r_\theta(\tilde\vx) = \tilde\vx - \sigma^2\nabla_{\tilde\vx}U_\theta(\tilde\vx) + \sigma^2\nabla_{\tilde\vx}g_\theta(\tilde\vx)^\top\bm{h}_\theta(\tilde\vx),
\end{aligned}
\end{equation}
a reconstruction loss in which $r_\theta$ is a decoder that reads out the representation $\bm{h}_\theta$ through $\nabla_{\tilde\vx}g_\theta(\tilde\vx)^\top$~\citep{vincent2011connection}. By Tweedie's formula~\Eqref{eqn:tweedie}, the optimal reconstruction is the posterior mean $\EE\sbr{\vx\mid\tilde\vx}$, attained when $p_\theta = p_\sigma$. A nonlinear $g_\theta$ gives a deep DAE, whose decoder matrix depends on the input. The classical DAE is the simplest case: with binary latents, a linear projection $g_\theta(\tilde\vx) = W\tilde\vx + \bm{b}$, and a quadratic $U_\theta(\tilde\vx) = \frac{1}{\sigma^2}\rbr{\frac{1}{2}\nbr{\tilde\vx}_2^2 - \bm{c}^\top\tilde\vx}$, the joint energy is the Gaussian--binary RBM of~\Secref{sec:prelim}, and the encoder and the reconstruction become
\begin{equation}\label{eqn:dae_energy}
    \bm{h}_\theta(\tilde\vx) = \sigmoid\rbr{W\tilde\vx + \bm{b}}, \qquad
    r_\theta(\tilde\vx) = \bm{c} + \sigma^2 W^\top\bm{h}_\theta(\tilde\vx),
\end{equation}
a sigmoid encoder and a linear decoder whose weights are tied to the encoder up to the factor $\sigma^2$~\citep{vincent2008extracting,vincent2011connection}. With Gaussian latents instead, the same construction gives a linear encoder $\bm{h}_\theta(\tilde\vx) = W\tilde\vx + \bm{b}$, \ie, a linear DAE.

\paragraph{DAE as a Conditional EBM.} The DAE can be derived as maximum likelihood for a Gaussian model of the denoising posterior whose mean is the reconstruction $r_\theta(\tilde\vx)$, the model of $\EE\sbr{\vx\mid\tilde\vx}$ in~\Eqref{eqn:dae_recon},
\begin{equation}\label{eqn:dae_ebm}
\begin{aligned}
    p_\theta(\vx\mid\tilde\vx) = \Ncal\rbr{\vx;\, r_\theta(\tilde\vx), I} &\propto e^{-\frac{1}{2}\nbr{\vx}_2^2}\exp\rbr{r_\theta(\tilde\vx)^\top\vx} \\
    &\propto e^{-\frac{1}{2}\nbr{\vx}_2^2 + \bm{c}^\top\vx}\exp\rbr{\bm{h}_\theta(\tilde\vx)^\top\psi_\theta(\vx)}, \qquad \psi_\theta(\vx) = \sigma^2W\vx,
\end{aligned}
\end{equation}
whose negative $\log$-likelihood recovers the original DAE reconstruction loss~\citep{vincent2008extracting},
\begin{equation}\label{eqn:dae_loss}
    \mathcal{L}_{\text{DAE}}(\theta) = -\EE_{\vx\sim p_d,\,\bm{\epsilon}}\sbr{\log p_\theta(\vx\mid\tilde\vx)} = \frac{1}{2}\,\EE_{\vx\sim p_d,\,\bm{\epsilon}}\sbr{\big\|r_\theta(\tilde\vx) - \vx\big\|_2^2} + \text{const}.
\end{equation}
The second line of~\Eqref{eqn:dae_ebm} substitutes the classical reconstruction $r_\theta(\tilde\vx) = \bm{c} + \sigma^2W^\top\bm{h}_\theta(\tilde\vx)$ of~\Eqref{eqn:dae_energy}: since the reconstruction is affine in the representation, the encoder enters the energy bilinearly. The derivation thus justifies the bilinear structure, and brings the DAE into the framework of~\Eqref{eqn:repr_ebm}, as a conditional EBM with the corrupted input as the condition, the encoder $\bm{h}_\theta(\tilde\vx)$ as its representation, and the clean input as the target. The original DAE used masking noise, the discrete counterpart that BERT later adopted.

\paragraph{Diffusion Spectral Representation.} The diffusion spectral representation keeps the conditional bilinear form~\Eqref{eqn:repr_ebm} and extends the DAE in three ways. The condition is a separate clean variable $\vx$, such as an image, a prompt, or a state--action pair, rather than a corrupted copy of the target; the target is a corrupted variable $\tilde\vy$, at every noise level $t$ of the forward process~\Eqref{eqn:diff_forward}, $\tilde\vy = \alpha_t\vy + \sigma_t\bm{\epsilon}$, with one representation $\phi_\theta(\vx)$ shared across noise levels; and the target features $\mu_\theta(\tilde\vy,t)$ are nonlinear, so the partition function over $\tilde\vy$ is no longer available, and the model is learned by score matching instead of maximum likelihood. The DAE and the diffusion spectral representation are thus the same kind of conditional EBM, with the DAE as the single-level special case whose condition is the corrupted input itself, in the same way that BERT is the single-level special case of masked diffusion in~\Secref{sec:bert}.
Following~\citet{dai2026spectral}, the noisy conditional $p_t(\tilde\vy\mid\vx) = \int q_t(\tilde\vy\mid\vy)\,p(\vy\mid\vx)\,d\vy$ is parameterized with the bilinear form of~\Eqref{eqn:repr_ebm},
\begin{equation}\label{eqn:diff_repr_ebm}
    p_{\theta,t}(\tilde\vy\mid\vx) \propto p_t(\tilde\vy)\exp\rbr{\phi_\theta(\vx)^\top\mu_\theta(\tilde\vy, t)},
\end{equation}
where the base measure $p_t(\tilde\vy)$ is the distribution of the corrupted target, inherited from the base measure $p(\vy)$ of~\Eqref{eqn:repr_ebm}. In practice, it is absorbed into the inner product by appending a constant coordinate $1$ to $\phi_\theta(\vx)$, whose corresponding coordinate of $\mu_\theta(\tilde\vy,t)$ then represents $\log p_t(\tilde\vy)$.
Since $\phi_\theta(\vx)$ does not depend on $\tilde\vy$, the score with respect to $\tilde\vy$ is linear in $\phi_\theta(\vx)$, read out through the Jacobian of $\mu_\theta$, and the partition function drops out,
\begin{equation}\label{eqn:diff_repr_score}
    \nabla_{\tilde\vy}\log p_{\theta,t}(\tilde\vy\mid\vx) = \nabla_{\tilde\vy}\mu_\theta(\tilde\vy,t)^\top\phi_\theta(\vx),
\end{equation}
with the base measure included through the constant coordinate. By Tweedie's formula applied to the conditional, $\alpha_t\EE\sbr{\vy\mid\tilde\vy,\vx} = \tilde\vy + \sigma_t^2\nabla_{\tilde\vy}\log p_t(\tilde\vy\mid\vx)$, so denoising score matching fits $\phi_\theta$ and $\mu_\theta$ by regression onto the clean target,
\begin{equation}\label{eqn:diff_repr_loss}
\begin{aligned}
    \mathcal{L}_{\text{diff}}(\theta) &= \EE_{t}\,\EE_{(\vx,\vy)\sim p_d,\,\bm{\epsilon}}\sbr{\Big\|\tilde\vy + \sigma_t^2\,\nabla_{\tilde\vy}\mu_\theta(\tilde\vy,t)^\top\phi_\theta(\vx) - \alpha_t\vy\Big\|_2^2} \\
    &= \EE_{t}\,\EE_{(\vx,\vy)\sim p_d,\,\bm{\epsilon}}\sbr{\sigma_t^2\,\Big\|\bm{\epsilon} + \sigma_t\,\nabla_{\tilde\vy}\mu_\theta(\tilde\vy,t)^\top\phi_\theta(\vx)\Big\|_2^2}.
\end{aligned}
\end{equation}
The second form shows that this is a conditional diffusion model whose noise predictor, $-\sigma_t\nabla_{\tilde\vy}\mu_\theta(\tilde\vy,t)^\top\phi_\theta(\vx)$, is linear in the representation of the condition. In practice, the Jacobian $\nabla_{\tilde\vy}\mu_\theta$ is parameterized directly by a network, which avoids second-order gradients~\citep{shribak2024diffusion}; after training, it is discarded and $\phi_\theta$ is kept.
For discrete targets such as text, the Gaussian corruption is replaced by masking and the regression by the masked diffusion loss~\citep[Sec.~7.2]{dai2026spectral}, which returns to the bilinear denoising EBM of BERT~\Eqref{eqn:bert_ebm}.

\subsection{Representation-Induced Decisive Models}\label{sec:disc_fm}

Many uses of large language models are decisions rather than open-ended generation: classifying an input, judging whether a response satisfies a requirement, or retrieving the documents that answer a request. A generative LLM makes such a decision by decoding text, which takes a full autoregressive pass and returns a string rather than a distribution over the possible answers. Such inference is expensive: each decision costs one forward pass per generated token, often including a long reasoning trace, and estimating the distribution over the answers requires sampling many responses. Decisive foundation models~\citep{typesafe2026jev} instead turn each query into a discriminative question: they take a context, a query, and a set of candidate answers, and return a distribution over the candidates. Unlike the discriminative models of~\Secref{sec:disc_model}, whose classes are fixed in advance, the classes of a decisive model are specified by the instruction and vary from one query to another. Instruction-following retrievers~\citep{zhuang2025towards} implement decisive foundation models by generalizing embedding models.

\paragraph{Instruction-Following Retrieval.} Instruction-following retrieval asks for the contexts $X$, such as passages of a corpus, that satisfy an instructional query $Q = (I, q)$, which consists of a query $q$ and an instruction $I$ that specifies requirements such as a perspective, format, or level of detail, and possibly the candidate answers. Each pair $(X,Q)$ thus poses a discriminative question with a yes-or-no answer: does $X$ satisfy $Q$? InF-Embed~\citep{zhuang2025towards} generalizes an embedding model to answer it. The score is the cosine similarity between the embeddings of the two sides,
\begin{equation}\label{eqn:inf_embed_score}
    s_\theta(X,Q) = \frac{\bm{x}^\top\bm{q}}{\nbr{\bm{x}}_2\nbr{\bm{q}}_2}, \qquad \bm{x} = g_{\theta}^{X}(X), \qquad \bm{q} = g_{\theta}^{Q}(I,q),
\end{equation}
where $g_\theta^X$ is the context encoder, and the instructional-query encoder $g_\theta^Q$ either encodes the concatenation of the instruction and the query, or encodes them separately and combines their embeddings by cross-attention. The score serves as the logit of the answer: with the two candidates yes and no, $p_\theta(\text{yes}\mid X,Q) = \sigma\rbr{s_\theta(X,Q) + b}$ with an offset $b$, and when the instruction lists $K$ candidate answers, each candidate receives a score and the answer distribution is the softmax of the scores over the $K$ candidates, of which the logistic form is the case $K=2$. Retrieval asks this question for every context in the corpus and ranks the contexts by their scores; since the context encoder does not depend on the query, the context embeddings can be precomputed and indexed offline, so answering a new query costs only one encoding of the query and one inner product per context, which makes inference fast and scalable to large corpora with approximate nearest-neighbor search.

\paragraph{Generalizing Embedding Models.} Standard embedding models such as BERT~\citep{devlin2019bert} learn an embedding $\bm{x}$ of the data only, and each new downstream question requires training a new classifier $\bm{w}^\top\bm{x}$ on top of it. In InF-Embed, the context embedding plays the same role, but the more important part is the embedding head $g_\theta^Q$ for the query and the instruction, which maps each instructional query to the weights $\bm{q} = g_\theta^Q(I,q)$ of this classifier. A new question is therefore answered at inference time by one encoding, without training a new classifier, and the same context can be relevant to one instruction and irrelevant to another. InF-Embed thus generalizes embedding models from embedding the data to also embedding the questions asked about it, while the context embeddings remain query-independent and can still be precomputed.

\paragraph{Learning InF-Embed.} InF-Embed learns the score by ranking NCE~\Eqref{eqn:lnce_rank}: treating the score as a potential on both sides, it fits the conditionals of the contexts given the query and of the queries given the context. For a batch $\cbr{(X_i,Q_i)}$, the negatives replace one side at a time,
\begin{equation}\label{eqn:inf_embed_loss}
    \ell(\theta) = -\EE_i\sbr{\log\frac{\exp\rbr{s_\theta(X_i,Q_i)}}{\sum_{m}\exp\rbr{s_\theta(X_m,Q_i)} + \sum_{Q'\in\mathcal{N}_i}\exp\rbr{s_\theta(X_i,Q')}}},
\end{equation}
where the first sum provides the negative contexts, and the set $\mathcal{N}_i$ of negative instructional queries contains those of the other examples as well as those that pair the query $q_i$ with the instruction of another example, so that the energy is contrasted against changes of the instruction alone. As in CLIP~(\Secref{sec:contrastive}), fitting both conditionals trains both sides of the energy, and replacing one side at a time keeps the denominator linear in the batch size. The training data supply the noise distribution: each positive pair $(X,Q)$ with $Q = (I,q)$ comes with two negative pairs $(X_1^-,(I^-,q))$ and $(X_2^-,(I,q^-))$, in which an LLM alters the instruction or the query slightly so that a different context becomes relevant. These hard negatives bring the noise distribution close to the data, which makes NCE statistically efficient, and they force the energy to depend on the instruction rather than on the query alone.

\paragraph{From Embeddings to Decisions.} InF-Embed is a bilinear representation following~\Eqref{eqn:repr_ebm}, with the context and the instructional query on the two sides. The major difference is in how it is used: in representation learning, the energy is discarded after training and the embedding is used alone for downstream tasks, whereas a decisive foundation model uses the learned model itself, which generalizes across queries and instructions, to make the decision. In general, the instructional query $Q$ itself may contain the candidate answers, which form a set $\mathcal{O}(Q)$, and the decision is
\begin{equation}\label{eqn:dfm_ebm}
    p_\theta(o\mid X,Q) = \frac{\exp\rbr{\phi_\theta(X)^\top\psi_\theta^{o}(Q)}}{\sum_{o'\in\mathcal{O}(Q)}\exp\rbr{\phi_\theta(X)^\top\psi_\theta^{o'}(Q)}}, \qquad o\in\mathcal{O}(Q),
\end{equation}
where $\phi_\theta$ encodes the context, and $\psi_\theta$ encodes the instructional query once and returns an embedding $\psi_\theta^{o}(Q)$ for each of its candidates. InF-Embed is the case of the two candidates yes and no, whose logit $\phi_\theta(X)^\top\rbr{\psi_\theta^{\text{yes}}(Q) - \psi_\theta^{\text{no}}(Q)}$ is parameterized by the score~\Eqref{eqn:inf_embed_score}.

The main difference from an LLM, which is also a softmax over bilinear energies, is that the potential in~\Eqref{eqn:dfm_ebm} encodes the entire instructional query $Q$ at once and gives the answer in a single step, whereas an LLM factorizes the answer autoregressively and generates it token by token, each step a softmax over the vocabulary. Hence a decision costs one forward pass rather than one per generated token, which makes inference much faster than with an LLM.

\section{Reinforcement Learning and Planning through EBMs}\label{sec:rl}

EBMs enter reinforcement learning~(RL) at several levels, depending on which part of the decision-making problem the energy models. In model-free RL, the energy is placed on the actions only: adding an entropy or KL regularizer to the reward or to the policy update makes the optimal policy an EBM over actions at each state, whose energy is the negative $Q$-function~(\Secref{sec:reg_rl}). Acting with such a policy requires sampling from this EBM~(\Secref{sec:ebm_policy}), and inverse RL and imitation learning fit EBMs over actions or trajectories to expert demonstrations~(\Secref{sec:irl}). Beyond the policy, the energy can be placed on the dynamics themselves, where an EBM world model yields a representation in which value functions are linear and can be learned by temporal-difference learning~(\Secref{sec:ebm_world_model}), and on whole trajectories, where planning becomes inference in a trajectory EBM that composes dynamics, behavior priors, goals, and constraints~(\Secref{sec:planning}).

\subsection{Naturally Induced EBM Policies}\label{sec:reg_rl}

\paragraph{Entropy-Regularized MDP.} Consider a Markov decision process with states $s$, actions $a$, reward $r(s,a)$, transition $P(s'\mid s,a)$, and discount $\gamma\in(0,1)$. Maximum-entropy RL seeks the policy
\begin{equation}\label{eqn:maxent_rl}
    \pi^\star = \argmax_\pi\; \EE_\pi\sbr{\sum_{t=0}^{\infty}\gamma^t\rbr{r(s_t,a_t) + \lambda\,H\rbr{\pi(\cdot\mid s_t)}}},
\end{equation}
where $H$ is the entropy and $\lambda>0$ is a temperature. Its solution is an EBM over actions, whose energy is the negative soft $Q$-function and whose $\log$-partition function is the soft value function,
\begin{equation}\label{eqn:soft_bellman}
\begin{aligned}
    &\pi^\star(a\mid s) = \exp\rbr{\frac{Q^\star(s,a) - V^\star(s)}{\lambda}}, \qquad
    V^\star(s) = \lambda\log\sum_{a}\exp\rbr{\frac{Q^\star(s,a)}{\lambda}}, \\
    &Q^\star(s,a) = r(s,a) + \gamma\,\EE_{s'\sim P(\cdot\mid s,a)}\sbr{V^\star(s')}.
\end{aligned}
\end{equation}
The last two equations form the soft Bellman equation, which replaces the $\max$ over actions in the standard Bellman optimality equation by a $\log$-sum-$\exp$, and recovers it as $\lambda\to0$.

The entropy-regularized objective~\Eqref{eqn:maxent_rl} has been proposed from several different motivations, all of which lead to the same EBM~\Eqref{eqn:soft_bellman}:
\begin{itemize}
    \item \textbf{Probabilistic inference.} Maximum-entropy inverse RL~\citep{ziebart2008maximum} applies the maximum entropy principle~(\Secref{sec:prelim}) to trajectories, and control as inference~\citep{levine2018reinforcement} casts optimal control as posterior inference in a graphical model whose potentials are exponentiated rewards.
    \item \textbf{Tractability.} Linearly-solvable MDPs~\citep{todorov2006linearly} penalize the KL divergence of the controlled dynamics from passive dynamics, which makes the Bellman equation linear in the exponentiated value, so that optimal control reduces to an eigenvalue problem.
    \item \textbf{Optimization smoothing.} SBEED~\citep{dai2018sbeed} interprets the entropy as Nesterov smoothing of the non-smooth $\max$ in the Bellman optimality equation, using that the negative entropy is the convex conjugate of $\log$-sum-$\exp$. The smoothed Bellman operator remains a contraction, and the smoothed Bellman equation admits a primal--dual reformulation that yields a convergent algorithm with nonlinear function approximation. The same operator appears as an alternative softmax operator for value updates~\citep{asadi2017alternative}.
    \item \textbf{Robustness to noise.} Viewing the policy as a channel with limited information, \citet{rubin2012trading} trade value against information, and soft $Q$-learning~\citep{fox2016taming} uses the resulting soft updates to reduce the overestimation bias caused by noisy value estimates. Maximum-entropy RL also provably solves certain robust RL problems with perturbed rewards and dynamics~\citep{eysenbach2022maximum}. Conversely, robustness to uncertain dynamics induces the same exponential form on the transition side: in soft robust MDPs, an adversary perturbs the transition kernel subject to a KL penalty, the worst case is the tilted kernel $\tilde P(s'\mid s,a)\propto P(s'\mid s,a)\exp\rbr{-V(s')/\lambda}$, an EBM over next states, and the problem is equivalent to a risk-sensitive MDP with the entropic risk measure~\citep{zhang2024soft}.
    \item \textbf{Exploration.} The entropy keeps the policy stochastic, and path consistency learning~\citep{nachum2017bridging}, soft $Q$-learning~\citep{haarnoja2017reinforcement}, and soft actor--critic~\citep{haarnoja2018soft} use it to explore multiple modes of the action space.
    \item \textbf{Unifying value- and policy-based methods.} With the entropy, soft $Q$-learning and policy gradient methods become equivalent~\citep{schulman2017equivalence}, and both are instances of a single convex-optimization view of entropy-regularized MDPs~\citep{neu2017unified}.
\end{itemize}
Whatever the motivation, the outcome is an EBM: acting is sampling from~\Eqref{eqn:soft_bellman}, and learning amounts to fitting its energy, so the tools of~\Secref{sec:prelim} apply directly.

\paragraph{KL-Regularized Policy Updates.} The entropy in~\Eqref{eqn:maxent_rl} modifies the MDP itself. A different use of the KL divergence keeps the vanilla MDP, \ie, $\lambda=0$, and regularizes each policy update instead. Given the current policy $\pi_k$, trust-region and mirror-descent policy optimization~\citep{schulman2015trust,schulman2017proximal,geist2019theory} improve the policy while penalizing, or constraining, its KL divergence from $\pi_k$. In the penalized form,
\begin{equation}\label{eqn:md_update}
    \pi_{k+1}(\cdot\mid s) = \argmax_{\pi}\; \EE_{a\sim\pi}\sbr{Q^{\pi_k}(s,a)} - \frac{1}{\eta}\,\mathrm{KL}\rbr{\pi\,\|\,\pi_k(\cdot\mid s)}
    \quad\Longrightarrow\quad
    \pi_{k+1}(a\mid s) \propto \pi_k(a\mid s)\exp\rbr{\eta\,Q^{\pi_k}(s,a)}.
\end{equation}
Each iterate is therefore a residual EBM~\Eqref{eq:residual_ebm} whose base measure is the previous policy. Unrolling the recursion from $\pi_0$ gives $\pi_k(a\mid s)\propto\pi_0(a\mid s)\exp\rbr{\eta\sum_{j<k}Q^{\pi_j}(s,a)}$, an EBM whose energy accumulates the $Q$-functions of all previous iterates. Unlike~\Eqref{eqn:soft_bellman}, the temperature is not fixed: the energy grows with $k$, so the iterates concentrate on the greedy actions and converge to the optimal policy of the vanilla MDP. The KL thus regularizes the optimization path rather than the solution, and the EBM structure appears in every update rather than at the optimum. KL regularization can also be taken toward a fixed reference policy $\pi_0$ rather than the previous iterate, which changes the objective rather than the update; the optimal policy is then the residual EBM with base measure $\pi_0$. This is the setting of RLHF for language models~\citep{ouyang2022training}, where $\pi_0$ is the pretrained model and, in the single-step case, $\pi^\star(a\mid s)\propto\pi_0(a\mid s)\exp\rbr{r(s,a)/\lambda}$~\citep{rafailov2023direct}.

These EBM policies have also been used to balance exploration and exploitation through optimism and pessimism, both theoretically and practically:
\begin{itemize}
    \item \textbf{Value-incentivized actor--critic~(VAC).} \citet{yang2025exploration} adopt the log-linear, \ie, energy-based, policy class derived from the maximum-entropy policy in a primal--dual actor--critic method that incentivizes exploration and achieves near-optimal regret, and extend it to KL-regularized MDPs.
    \item \textbf{Value-incentivized preference optimization.} For RLHF, VPO~\citep{cen2025value} regularizes the maximum likelihood estimate of the reward by the corresponding optimal KL-regularized value, which is the expected $\log$-partition function $\lambda\,\EE\sbr{\log Z(s)}$ of the EBM policy; its sign selects optimism for online exploration or pessimism for offline learning.
\end{itemize}

\subsection{RL for EBM Policies}\label{sec:ebm_policy}

The regularized policy updates of~\Secref{sec:reg_rl} are solved in closed form by EBMs over actions,~\Eqref{eqn:soft_bellman} and~\Eqref{eqn:md_update}. The optimization is thus easy, but acting requires sampling actions from these EBMs, which brings back the sampling difficulty of~\Secref{sec:inference}. For discrete actions, the partition function is a sum over $\abr{\mathcal{A}}$ actions, the policy is a softmax over $Q$-values, and sampling is exact. For continuous actions, the partition function is an integral, and the commonly used methods fall into the following three classes.

\paragraph{Variational Policy Optimization.} The first restricts the policy to a family $\pi_\phi$ that is easy to sample from, such as a Gaussian with the reparameterization trick, and optimizes~\Eqref{eqn:maxent_rl} or~\Eqref{eqn:md_update} within this family. By the variational bound~\Eqref{eq:vi_bound}, this projects the EBM onto the family,
\begin{equation}\label{eqn:vi_policy}
\begin{aligned}
    &\max_\phi\; \EE_{a\sim\pi_\phi(\cdot\mid s)}\sbr{Q(s,a)} + \lambda\,H\rbr{\pi_\phi(\cdot\mid s)}
    = -\lambda\,\mathrm{KL}\rbr{\pi_\phi(\cdot\mid s)\,\Big\|\,\frac{\exp\rbr{Q(s,\cdot)/\lambda}}{Z(s)}} + \lambda\log Z(s), \\
    &\max_\phi\; \EE_{a\sim\pi_\phi(\cdot\mid s)}\sbr{Q^{\pi_k}(s,a)} - \frac{1}{\eta}\,\mathrm{KL}\rbr{\pi_\phi(\cdot\mid s)\,\|\,\pi_k(\cdot\mid s)}.
\end{aligned}
\end{equation}
The first is the actor update of soft actor--critic~(SAC)~\citep{haarnoja2018soft}, and the second is mirror-descent policy optimization with a parametric policy~\citep{tomar2022mirror}, of which TRPO and PPO are constrained and clipped variants. Sampling is cheap, but the family limits the policy: a Gaussian cannot represent the multimodal EBMs that arise when several actions are nearly optimal.

\paragraph{MCMC Sampling.} The second samples from the EBM directly. The score of the policy is the action gradient of the $Q$-function, $\nabla_a\log\pi(a\mid s) = \nabla_a Q(s,a)/\lambda$, so Langevin dynamics~\Eqref{eq:langevin} draws actions by $a\leftarrow a + \frac{\epsilon}{\lambda}\nabla_a Q(s,a) + \sqrt{2\epsilon}\,\bm{\xi}$. This represents any EBM, but each action requires many gradient evaluations of $Q$. Soft $Q$-learning~\citep{haarnoja2017reinforcement} amortizes the sampler: a network $a = f_\phi(s,\bm{\xi})$ is trained by amortized Stein variational gradient descent~(SVGD)~\citep{liu2016stein} so that its samples follow the EBM, which combines the expressiveness of the EBM with the cost of a single forward pass, as with the implicit samplers of~\Secref{sec:inference}.

\paragraph{Diffusion Policies.} The third represents the policy by a diffusion model, \ie, a sequence of noise-perturbed EBMs~(\Secref{sec:diffusion}), which keeps the expressiveness of the EBM while sampling by a fixed number of denoising steps. The obstacle is training: denoising score matching~\Eqref{eqn:dsm} requires samples from the target policy, which are unavailable in online RL, since the target is known only through its energy. We write $a^\tau$ for the noisy action at diffusion time $\tau$, with $a^0$ the clean action, to distinguish it from the environment step $t$, and $\bm{s}_\theta(a^\tau;s,\tau)$ for the score network. \citet{ma2025efficient} observe that the score matching loss can be reweighted by any positive function $g$ without changing its minimizer, $\bm{s}_\theta(a^\tau;s,\tau) = \nabla_{a^\tau}\log p_\tau(a^\tau\mid s)$,
\begin{equation}\label{eqn:rsm}
    \mathcal{L}^{g}(\theta;s,\tau) = \int g(a^\tau;s)\,\big\|\bm{s}_\theta(a^\tau;s,\tau) - \nabla_{a^\tau}\log p_\tau(a^\tau\mid s)\big\|_2^2\,da^\tau,
\end{equation}
where $p_\tau$ is the noise-perturbed target policy, and choose $g$ so that the loss becomes tractable. For the KL-regularized update~\Eqref{eqn:md_update}, choosing $g$ proportional to $p_\tau$ times the partition function gives diffusion policy mirror descent~(DPMD), a denoising score matching loss on samples from the current policy, weighted by the exponentiated $Q$-function,
\begin{equation}\label{eqn:dpmd}
    \mathcal{L}_{\text{DPMD}}(\theta;s,\tau) = \EE_{a^0\sim\pi_k(\cdot\mid s),\,a^\tau\sim q_{\tau|0}(\cdot\mid a^0)}\sbr{\exp\rbr{\eta\,Q^{\pi_k}(s,a^0)}\,\big\|\bm{s}_\theta(a^\tau;s,\tau) - \nabla_{a^\tau}\log q_{\tau|0}(a^\tau\mid a^0)\big\|_2^2}.
\end{equation}
For the entropy-regularized EBM~\Eqref{eqn:soft_bellman}, there is no previous policy to sample from. Soft diffusion actor--critic~(SDAC) instead draws $a^\tau$ from a sampling distribution $h_\tau$ of our choice, draws $\tilde a^0$ from the Gaussian $\tilde q_{0|\tau}(\cdot\mid a^\tau)$ obtained by reversing the perturbation kernel, and weights the loss by $\exp\rbr{Q(s,\tilde a^0)/\lambda}$, so that only evaluations of the energy are needed,
\begin{equation}\label{eqn:sdac}
    \mathcal{L}_{\text{SDAC}}(\theta;s,\tau) = \EE_{a^\tau\sim h_\tau,\,\tilde a^0\sim\tilde q_{0|\tau}(\cdot\mid a^\tau)}\sbr{\exp\rbr{Q(s,\tilde a^0)/\lambda}\,\big\|\bm{s}_\theta(a^\tau;s,\tau) - \nabla_{a^\tau}\log\tilde q_{0|\tau}(\tilde a^0\mid a^\tau)\big\|_2^2}.
\end{equation}
Both losses cost the same as standard diffusion training and avoid backpropagating through the denoising chain, and actions are drawn by the reverse diffusion, which plays the role of annealed Langevin dynamics on the EBM policy.

\paragraph{Contrastive Policy Learning.} The closed-form EBM policy can also be used to learn the policy directly from comparisons, without sampling from it. In RL from human feedback, a prompt $s$ is answered by a response $a$, and the policy is fine-tuned by the KL-regularized objective of~\Secref{sec:reg_rl} toward a reference policy $\pi_0$, whose optimal policy is the residual EBM
\begin{equation}\label{eqn:rlhf_ebm}
    \pi^\star(a\mid s) = \frac{\pi_0(a\mid s)\exp\rbr{r(s,a)/\lambda}}{Z(s)}
    \quad\Longleftrightarrow\quad
    r(s,a) = \lambda\log\frac{\pi^\star(a\mid s)}{\pi_0(a\mid s)} + \lambda\log Z(s).
\end{equation}
The second form expresses the reward through the policy, up to the $\log$-partition function $\lambda\log Z(s)$, which depends only on the prompt. Direct preference optimization~(DPO)~\citep{rafailov2023direct} exploits this. Preferences between a chosen response $a_w$ and a rejected response $a_l$ are modeled by the Bradley--Terry model~\citep{bradley1952rank}, $p(a_w\succ a_l\mid s) = \sigmoid\rbr{r(s,a_w) - r(s,a_l)}$, which depends only on reward differences, so $Z(s)$ cancels. Parameterizing the policy $\pi_\theta$ in place of $\pi^\star$ in~\Eqref{eqn:rlhf_ebm} gives the DPO loss
\begin{equation}\label{eqn:dpo}
    \mathcal{L}_{\text{DPO}}(\theta) = -\EE_{(s,a_w,a_l)}\sbr{\log\sigmoid\rbr{\lambda\log\frac{\pi_\theta(a_w\mid s)}{\pi_0(a_w\mid s)} - \lambda\log\frac{\pi_\theta(a_l\mid s)}{\pi_0(a_l\mid s)}}}.
\end{equation}
In the EBM view, $\pi_\theta$ is a residual EBM with base measure $\pi_0$, whose negative energy is the implicit reward of~\Eqref{eqn:rlhf_ebm} with $\pi_\theta$ in place of $\pi^\star$, and~\Eqref{eqn:dpo} is a contrastive objective of the kind in~\Secref{sec:learning}: as in ranking NCE with two candidates, the partition function cancels between the chosen and the rejected response, and the gradient lowers the energy of $a_w$ and raises that of $a_l$, weighted by how badly the current policy ranks them. The policy is thus learned without sampling from it and without fitting a separate reward model. Replacing the pair by a ranking of $K$ responses gives the corresponding Plackett--Luce, \ie, ranking NCE, objective, and the value-incentivized regularizer of VPO~(\Secref{sec:reg_rl}) adds the $\log$-partition function $\lambda\,\EE\sbr{\log Z(s)}$ back for exploration.

\subsection{Inverse RL with EBMs}\label{sec:irl}

In inverse RL and imitation learning, the reward is unknown and only expert demonstrations are available. Assuming that the expert acts optimally in the entropy-regularized MDP~\Eqref{eqn:maxent_rl}, the demonstrations are samples from an EBM, and learning from them is EBM learning. Maximum-entropy IRL, implicit behavioral cloning, and GAIL fit such EBMs, and they differ mainly in which EBM they fit and in which learning method of~\Secref{sec:learning} they use.

\paragraph{Maximum-Entropy IRL: Maximum Likelihood.} Maximum-entropy IRL~\citep{ziebart2008maximum} models the distribution of expert trajectories $\tau = (s_0,a_0,s_1,a_1,\ldots)$ as an EBM whose energy is the negative return under a parameterized reward $r_\theta$,
\begin{equation}\label{eqn:maxent_irl}
    p_\theta(\tau) = \frac{1}{Z_\theta}\,\underbrace{\rho(s_0)\prod_{t}P(s_{t+1}\mid s_t,a_t)}_{\text{base measure}}\,\exp\rbr{\sum_{t}r_\theta(s_t,a_t)},
\end{equation}
\ie, a residual EBM~\Eqref{eq:residual_ebm} whose base measure is given by the dynamics. For a linear reward $r_\theta = \theta^\top\vf(s,a)$, it is the maximum entropy distribution over trajectories that matches the expected features of the expert, as in~\Secref{sec:prelim}. Maximum likelihood on the demonstrations $\mathcal{D}$ has the contrastive gradient~\Eqref{eqn:gradient_ebm},
\begin{equation}\label{eqn:maxent_irl_grad}
    \nabla_\theta\log\mathcal{L}(\theta) = \EE_{\tau\sim\mathcal{D}}\sbr{\sum_{t}\nabla_\theta r_\theta(s_t,a_t)} - \EE_{\tau\sim p_\theta}\sbr{\sum_{t}\nabla_\theta r_\theta(s_t,a_t)},
\end{equation}
which raises the reward of expert trajectories and lowers that of the model's own. The negative samples come from the EBM~\Eqref{eqn:maxent_irl}. With deterministic dynamics, summing over trajectories backward in time yields the soft Bellman equation~\Eqref{eqn:soft_bellman} with $\lambda=1$, the EBM factorizes as $p_\theta(\tau) = \rbr{\rho(s_0)\exp V_\theta(s_0)/Z_\theta}\prod_t\pi_\theta(a_t\mid s_t)$ along the trajectory with $\pi_\theta(a\mid s) = \exp\rbr{Q_\theta(s,a) - V_\theta(s)}$, and the partition function is $Z_\theta = \EE_{\rho}\sbr{\exp V_\theta(s_0)}$; the sampler is thus the optimal policy of the entropy-regularized MDP with reward $r_\theta$. Each gradient step therefore requires solving a forward maximum-entropy RL problem, the RL counterpart of running MCMC to convergence in maximum likelihood. With stochastic dynamics, the same summation gives the policy $\pi_\theta(a\mid s) = \exp\rbr{Q_\theta(s,a) - V_\theta(s)}$ with
\begin{equation}\label{eqn:maxent_irl_stoch}
    V_\theta(s) = \log\sum_{a}\exp Q_\theta(s,a), \qquad Q_\theta(s,a) = r_\theta(s,a) + \log\EE_{s'\sim P(\cdot\mid s,a)}\sbr{\exp V_\theta(s')},
\end{equation}
whose backup takes the $\log$-expectation of $\exp V_\theta$ rather than the expectation of $V_\theta$. Maximum causal entropy IRL~\citep{ziebart2010modeling} instead uses the optimal policy of the entropy-regularized MDP~\Eqref{eqn:maxent_rl}~\citep{nachum2017bridging,dai2018sbeed} with reward $r_\theta$, which is given by the standard soft Bellman equation~\Eqref{eqn:soft_bellman}, so that rolling out $\pi_\theta$ in the environment provides exact negative samples. Guided cost learning~\citep{finn2016guided} avoids the inner loop by estimating the second expectation with importance sampling, using the current policy as the proposal, and improving the policy alongside the reward.

\paragraph{Implicit Behavioral Cloning: Ranking NCE.} Implicit behavioral cloning~(IBC)~\citep{florence2022implicit} skips the reward and fits the expert policy itself as a conditional EBM, $\pi_\theta(a\mid s)\propto\exp\rbr{-E_\theta(s,a)}$, by ranking NCE~\Eqref{eqn:lnce_rank},
\begin{equation}\label{eqn:ibc}
    \mathcal{L}_{\text{IBC}}(\theta) = -\EE_{(s,a)\sim\mathcal{D}}\sbr{\log\frac{\exp\rbr{-E_\theta(s,a)}}{\exp\rbr{-E_\theta(s,a)} + \sum_{k=1}^{K}\exp\rbr{-E_\theta(s,\tilde a_k)}}},
\end{equation}
where the negative actions $\tilde a_k$ are drawn uniformly from the action space. No dynamics or RL are involved, and actions are predicted by minimizing the energy, \ie, MAP prediction~(\Secref{sec:inference}), with derivative-free optimization or Langevin dynamics. Compared with explicit regression $a = f_\theta(s)$, the EBM represents multimodal and discontinuous expert behavior instead of averaging its modes, and IBC is the discriminative conditional EBM of~\Secref{sec:disc_model} applied to actions.

\paragraph{GAIL: Adversarial Training.} Generative adversarial imitation learning~(GAIL)~\citep{ho2016generative} has the same learning target as IBC, the expert policy rather than the reward, but fits it by the adversarial training of~\Secref{sec:learning} instead of NCE. Matching the occupancy measure of the expert with an entropy-regularized policy leads to
\begin{equation}\label{eqn:gail}
    \min_{\pi}\max_{D}\; \EE_{(s,a)\sim\pi_E}\sbr{\log D(s,a)} + \EE_{(s,a)\sim\pi}\sbr{\log\rbr{1 - D(s,a)}} - \lambda\,H(\pi),
\end{equation}
where the policy is the learned sampler and the discriminator $D$ provides the contrastive signal between expert and policy samples. The policy is trained by RL with the surrogate reward $-\log\rbr{1-D}$, and at the optimum its occupancy measure matches that of the expert, while the discriminator does not recover the expert's reward. This is the structure of the entropy-regularized GAN in~\Secref{sec:gan}, with the generator replaced by a policy acting in the environment. The two methods differ in their negative samples and in how the policy is represented. IBC uses uniform negatives and needs no interaction with the environment, but the policy is an explicit energy that must be minimized at every step. GAIL draws its negatives from the rollouts of the current policy, which also corrects the states that the policy itself visits, and the policy is an amortized sampler that acts in a single forward pass. Conversely, the adversarial form also recovers IRL: \citet{finn2016connection} show that guided cost learning is a GAN whose discriminator has the form $D = \frac{1}{Z_\theta}\exp\rbr{-E_\theta}/\rbr{\frac{1}{Z_\theta}\exp\rbr{-E_\theta} + q}$, where $q$ is the density of the sampler, and adversarial IRL~\citep{fu2018learning} uses this form at the level of state--action pairs to recover a reward that transfers across dynamics.

\paragraph{Which EBM Is Learned.} The three methods fit EBMs at different levels, which are related by the soft Bellman equation~\Eqref{eqn:soft_bellman}. Maximum-entropy IRL learns a per-step energy, the negative reward $-r_\theta$, whose EBM is over trajectories and whose policy is obtained only by solving the MDP with the dynamics. IBC and GAIL learn the policy directly, IBC as an explicit energy and GAIL as a sampler; if the expert is optimal for~\Eqref{eqn:maxent_rl}, the energy of the policy is the negative soft advantage, $E(s,a) = -\rbr{Q^\star(s,a) - V^\star(s)}/\lambda$, which already integrates the future through the dynamics. The reward is recovered from it only through the soft Bellman equation, $r(s,a) = Q^\star(s,a) - \gamma\,\EE_{s'}\sbr{V^\star(s')}$. Learning the reward costs an RL problem in every step but yields an energy that transfers to new dynamics, whereas learning the policy avoids solving for the reward but is tied to the dynamics of the demonstrations.

\subsection{RL with EBM World Models}\label{sec:ebm_world_model}

\begin{figure}[t]
    \centering
    \includegraphics[width=\textwidth]{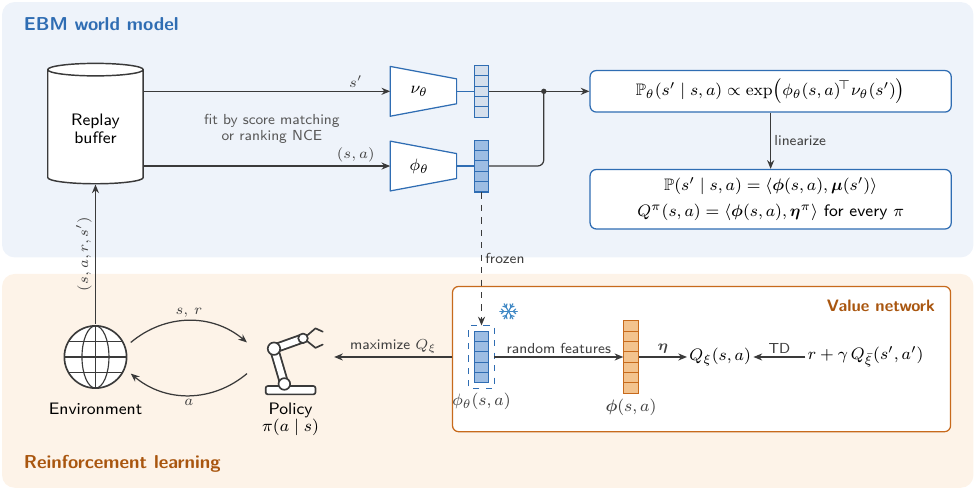}
    \caption{\textbf{RL with an EBM world model.} Top: the world model~\Eqref{eqn:ebm_world_model} scores a transition by the inner product of $\phi_\theta(s,a)$ and $\nu_\theta(s')$, and is fit to transitions from the replay buffer by score matching~\Eqref{eqn:diff_repr_loss} or ranking NCE~\Eqref{eqn:lnce_rank}. Random features linearize it~\Eqref{eqn:phi_ebm}, so that the $Q$-function of every policy is linear in $\bm{\phi}(s,a)$~\Eqref{eqn:spectral_q}. Bottom: $\phi_\theta$ is frozen and used as the input of the value network~\Eqref{eqn:spectral_td}, which computes these features with sinusoidal activations and is trained by TD learning, while the policy maximizes $Q_\xi$ and its transitions fill the buffer. Figure adapted from~\citet{gao2025spectral}.}
    \label{fig:spectral_rl}
    \vspace{-10pt}
\end{figure}

The previous subsections place the energy on actions and use no model of the dynamics. A world model, \ie, a model of the transition $\mathbb{P}(s'\mid s,a)$, is the other ingredient of RL, and it can itself be an EBM. Planners, discussed in~\Secref{sec:planning}, roll out such a model, but an EBM world model can also be used without sampling from it, by extracting from it a representation in which value functions are linear and learning the value by temporal-difference~(TD) learning~\citep{ren2022free,shribak2024diffusion,ren2025stochastic,gao2025spectral}, as summarized in~\Figref{fig:spectral_rl}. The world model is the bilinear EBM of~\Secref{sec:contrastive}, with the state--action pair as the condition and the next state as the target,
\begin{equation}\label{eqn:ebm_world_model}
    \mathbb{P}_\theta(s'\mid s,a) = \frac{\exp\rbr{\phi_\theta(s,a)^\top\nu_\theta(s')}}{Z_\theta(s,a)}, \qquad Z_\theta(s,a) = \int\exp\rbr{\phi_\theta(s,a)^\top\nu_\theta(s')}\,ds',
\end{equation}
\ie, an EBM with energy $E_\theta(s,a,s') = -\phi_\theta(s,a)^\top\nu_\theta(s')$. The same construction extends to partially observed MDPs, where the state is replaced by a short window of past observations and actions, and a bilinear EBM over the next observation gives a representation of the belief in which value functions are again linear~\citep{zhang2023energy}, \citep[Section~6]{gao2025spectral}.

\paragraph{Linear Value Functions.} Suppose that the transition and the reward factorize through a feature map $\bm{\phi}(s,a)$ in a Hilbert space $\Hcal$,
\begin{equation}\label{eqn:spectral_mdp}
    \mathbb{P}(s'\mid s,a) = \inner{\bm{\phi}(s,a)}{\bm{\mu}(s')}_{\Hcal}, \qquad r(s,a) = \inner{\bm{\phi}(s,a)}{\bm{\theta}_r}_{\Hcal},
\end{equation}
which always holds for some, possibly infinite-dimensional, $\Hcal$, \eg, by the singular value decomposition of the transition operator. Then the $Q$-function of every policy $\pi$ is linear in $\bm{\phi}$~\citep[Lemma~1]{gao2025spectral}, since by the Bellman equation
\begin{equation}\label{eqn:spectral_q}
\begin{aligned}
    Q^\pi(s,a) &= r(s,a) + \gamma\int\mathbb{P}(s'\mid s,a)\,V^\pi(s')\,ds' = \inner{\bm{\phi}(s,a)}{\bm{\eta}^\pi}_{\Hcal}, \\
    \bm{\eta}^\pi &= \bm{\theta}_r + \gamma\int\bm{\mu}(s')\,V^\pi(s')\,ds'.
\end{aligned}
\end{equation}
The representation $\bm{\phi}$ is therefore sufficient for the $Q$-function of every policy, and only the weights $\bm{\eta}^\pi$ depend on the policy.

\paragraph{Linearizing the EBM World Model.} The EBM~\Eqref{eqn:ebm_world_model} is not of the form~\Eqref{eqn:spectral_mdp}, because the inner product sits inside the exponential, but it can be brought into that form with random features. Completing the square, $\phi^\top\nu = \frac{1}{2}\nbr{\phi}^2 + \frac{1}{2}\nbr{\nu}^2 - \frac{1}{2}\nbr{\phi-\nu}^2$, turns $\exp\rbr{\phi^\top\nu}$ into a Gaussian kernel $k(\phi-\nu) = \exp\rbr{-\frac{1}{2}\nbr{\phi-\nu}^2}$ between the two representations, rescaled by functions of each side alone. A shift-invariant kernel is an expectation of products of random Fourier features~\citep[Lemma~2]{gao2025spectral},
\begin{equation}\label{eqn:rff}
    k(\vx-\vy) = \EE_{\bm{\omega}\sim p(\bm{\omega})}\sbr{\zeta_{\bm{\omega}}(\vx)\,\zeta_{\bm{\omega}}(\vy)^*}, \qquad \zeta_{\bm{\omega}}(\vx) = e^{\mathrm{i}\bm{\omega}^\top\vx},
\end{equation}
where $p(\bm{\omega})$ is the Fourier transform of $k$, which is $\Ncal\rbr{0,I}$ for the Gaussian kernel. Replacing the expectation by $N$ samples gives finite features $\bm{\zeta}_N(\vx)$ of cosines and sines, and substituting into~\Eqref{eqn:ebm_world_model} gives the factorization~\Eqref{eqn:spectral_mdp} with
\begin{equation}\label{eqn:phi_ebm}
\begin{aligned}
    \bm{\phi}_{\text{EBM}}(s,a) &= \frac{\exp\rbr{\frac{1}{2}\nbr{\phi_\theta(s,a)}^2}\,\bm{\zeta}_N\rbr{\phi_\theta(s,a)}}{Z_\theta(s,a)} = \frac{\bm{\zeta}_N\rbr{\phi_\theta(s,a)}}{\inner{\bm{\zeta}_N\rbr{\phi_\theta(s,a)}}{\bm{u}}}, \\
    \bm{\mu}_{\text{EBM}}(s') &= \exp\rbr{\tfrac{1}{2}\nbr{\nu_\theta(s')}^2}\,\bm{\zeta}_N\rbr{\nu_\theta(s')},
\end{aligned}
\end{equation}
where the second form uses that the partition function is itself linear in the same features, $Z_\theta(s,a) = \inner{\exp\rbr{\frac{1}{2}\nbr{\phi_\theta}^2}\bm{\zeta}_N(\phi_\theta)}{\bm{u}}$ with $\bm{u} = \int\bm{\mu}_{\text{EBM}}(s')\,ds'$. The $Q$-functions are thus linear in features computed from $\phi_\theta(s,a)$ alone. Stochastic nonlinear dynamics with Gaussian noise, $s' = f(s,a) + \bm{\epsilon}$, are a special case, since $\mathbb{P}(s'\mid s,a)\propto\exp\rbr{-\nbr{s'-f(s,a)}^2/(2\sigma^2)}$ is a Gaussian kernel between $f(s,a)$ and $s'$, and the same random features give a finite-dimensional representation of the known or learned dynamics directly~\citep{ren2022free,ren2025stochastic}.

\paragraph{Learning the World Model.} Since the world model is the bilinear EBM of~\Secref{sec:repr}, its representation $\phi_\theta$ is learned by the same estimators, with the state--action pair as $\vx$ and the next state as $\vy$. Diff-SR~\citep{shribak2024diffusion} uses score matching, the diffusion spectral representation of~\Secref{sec:diff_repr}: the next state is perturbed by a Gaussian kernel, $\mathbb{P}(\tilde s'\mid s';\beta) = \Ncal\rbr{\sqrt{1-\beta}\,s',\beta I}$, the perturbed transition is parameterized as $\mathbb{P}_\theta(\tilde s'\mid s,a;\beta)\propto\exp\rbr{\phi_\theta(s,a)^\top\nu_\theta(\tilde s';\beta)}$ with $\phi_\theta$ shared across noise levels, and its score $\nabla_{\tilde s'}\nu_\theta(\tilde s';\beta)^\top\phi_\theta(s,a)$ is regressed onto the score of the perturbation kernel, as in~\Eqref{eqn:diff_repr_loss}. CTRL-SR~\citep{gao2025spectral} uses ranking NCE~\Eqref{eqn:lnce_rank} instead, with the model $\mathbb{P}_\theta(s'\mid s,a)\propto\tilde{\mathbb{P}}(s')\exp\rbr{\phi_\theta(s,a)^\top\nu_\theta(s')}$ relative to a noise distribution $\tilde{\mathbb{P}}$ over next states; the noise term depends only on $s'$ and is absorbed into $\bm{\mu}_{\text{EBM}}$, so it does not change $\bm{\phi}_{\text{EBM}}$. To avoid negatives that are too easy to tell apart from the true next state, the positives and negatives are perturbed by the same Gaussian kernel at several noise levels, and the ranking NCE losses are averaged over the levels, so that the classification becomes harder, and more informative, as the noise grows.

\paragraph{Relation to Successor Features.} Successor features~\citep{dayan1993improving,barreto2017successor} also make the $Q$-function linear. With $r = \inner{\bm{\phi}}{\bm{\theta}_r}_{\Hcal}$, $Q^\pi(s,a) = \inner{\bm{\psi}^\pi(s,a)}{\bm{\theta}_r}_{\Hcal}$, where the successor features satisfy their own Bellman equation,
\begin{equation}\label{eqn:sf_td}
    \bm{\psi}^\pi(s,a) = \EE_\pi\sbr{\sum_{t\ge0}\gamma^t\bm{\phi}(s_t,a_t)\,\Big|\, s_0=s,a_0=a} = \bm{\phi}(s,a) + \gamma\,\EE_{s'\sim\mathbb{P}(\cdot\mid s,a),\,a'\sim\pi(\cdot\mid s')}\sbr{\bm{\psi}^\pi(s',a')},
\end{equation}
and are learned by TD on the right-hand side, as $Q^\pi$ is, with the vector $\bm{\phi}$ in place of the reward. Its fixed point is $\bm{\psi}^\pi = (I-\gamma\mathbb{P}^\pi)^{-1}\bm{\phi}$, with $\mathbb{P}^\pi(s',a'\mid s,a) = \mathbb{P}(s'\mid s,a)\,\pi(a'\mid s')$, and learning low-rank successor features in this way is equivalent to taking the left singular vectors of the multi-step operator $(I-\gamma\mathbb{P}^\pi)^{-1}$~\citep{ren2023spectral}. In the terms of~\Secref{sec:repr}, the two representations differ in the pair of variables that the energy couples: the EBM world model couples the state--action pair with the next state, whereas successor features couple it with the discounted future occupancy under $\pi$. Successor features therefore depend on the policy: they transfer across rewards for a fixed $\pi$, but must be relearned by TD when the policy changes. The representation $\bm{\phi}$ factorizes the single-step transition, does not depend on the policy, and is sufficient for the $Q$-function of every policy, which enters only through $\bm{\eta}^\pi$; it is learned once from transitions by NCE or score matching, which is more convenient than TD.

\paragraph{Representation-Based TD.} Once $\phi_\theta$ is learned, it is frozen, and the $Q$-function is fitted on top of it by TD learning, without sampling from the world model. Following~\Eqref{eqn:phi_ebm}, the $Q$-network computes the random features of $\phi_\theta(s,a)$ with sinusoidal activations and adds one layer for the normalization,
\begin{equation}\label{eqn:spectral_td}
\begin{aligned}
    Q_{\xi}(s,a) &= \mathrm{act}\rbr{W_2\sbr{\cos\rbr{W_1\phi_\theta(s,a)};\sin\rbr{W_1\phi_\theta(s,a)}}}^\top\bm{\eta}, \\
    \ell_{\text{TD}}(\xi) &= \EE_{(s,a,r,s')\sim\mathcal{D}}\sbr{\rbr{r + \gamma\,\EE_{a'\sim\pi}\sbr{Q_{\bar\xi}(s',a')} - Q_\xi(s,a)}^2},
\end{aligned}
\end{equation}
where $\xi = (W_1,W_2,\bm{\eta})$, $\mathrm{act}$ is a nonlinear activation, and $\bar\xi$ is a slowly updated target copy, and the policy maximizes $Q_\xi$. The TD objective does not train $\phi_\theta$: by~\Eqref{eqn:spectral_q}, the representation learned from the dynamics already spans the $Q$-function of every policy, so TD only fits the policy-dependent weights. The EBM view thus separates the two parts of model-based RL: the world model is learned as an EBM by the estimators of~\Secref{sec:learning}, and its partition function and sampler are never needed for control, since the random features linearize both the transition and its normalization.

\subsection{Planning with Trajectory EBMs}
\label{sec:planning}
 
The policies of Sections~\ref{sec:reg_rl}--\ref{sec:irl} are EBMs over the action at a single state. Planning instead reasons about whole trajectories: given a model of the dynamics, such as the world models of~\Secref{sec:ebm_world_model}, and a task, it searches for sequences of states and actions that accomplish the task. The entropy regularization of~\Secref{sec:reg_rl} again turns this search into inference in an EBM, now over trajectories~\citep{attias2003planning,toussaint2009robot,levine2018reinforcement}. This EBM is where composition~(\Secref{sec:compose}) is most useful: learned dynamics, behavior priors, goals, costs, and constraints each contribute an energy term, and a new task is a new sum of terms.
 
\paragraph{Planning as Inference over Trajectories.} Consider trajectories $\tau = (s_0,a_0,\ldots,a_{T-1},s_T)$ of horizon $T$ from a given initial state $s_0$, with the base measure 
\begin{equation}\label{eqn:plan_base}
    p_0(\tau) = \prod_{t=0}^{T-1}\pi_0(a_t\mid s_t)\,P(s_{t+1}\mid s_t,a_t),
\end{equation}
where $\pi_0$ is a prior over actions: uniform for maximum entropy, or a behavior model learned from data for KL regularization. With the return $R(\tau) = \sum_{t=0}^{T-1}r(s_t,a_t) + r_T(s_T)$, where $r_T$ scores the final state, maximum-entropy planning solves
\begin{equation}\label{eqn:plan_objective}
    \max_{q}\;\EE_{\tau\sim q}\sbr{R(\tau)} - \lambda\,\mathrm{KL}\rbr{q\,\|\,p_0}.
\end{equation}
By the same argument as~\Eqref{eqn:logz_variational}, with $p_0$ in place of the uniform base measure, the solution is the residual EBM~\Eqref{eq:residual_ebm}
\begin{equation}\label{eqn:plan_ebm}
    q^\star(\tau) = \frac{p_0(\tau)\exp\rbr{R(\tau)/\lambda}}{Z}, \qquad Z = \EE_{\tau\sim p_0}\sbr{\exp\rbr{R(\tau)/\lambda}},
\end{equation}
and the optimal value is $\lambda\log Z$. This is the trajectory EBM of maximum-entropy IRL~\Eqref{eqn:maxent_irl} with the roles reversed: IRL learns the reward from samples of this EBM, while planning draws samples from it for a given reward. As $\lambda\to0$, $q^\star$ concentrates on the trajectories of highest return in the support of $p_0$, and planning reduces to trajectory optimization, \ie, MAP prediction~\Eqref{eq:ebm_map_prediction} with this EBM. One caveat concerns stochastic dynamics. Since $q$ in~\Eqref{eqn:plan_objective} ranges over all distributions over trajectories, $q^\star$ also reweights the transitions toward favorable outcomes, which is the optimistic backup~\Eqref{eqn:maxent_irl_stoch} of~\Secref{sec:irl}; requiring $q$ to keep the true dynamics instead gives the soft Bellman backup~\Eqref{eqn:soft_bellman} and a closed-loop policy. For deterministic dynamics, the two coincide, and $q^\star$ is a distribution over action sequences, with the states given by rolling out the model. Most planners work in this setting: they sample or optimize an open-loop action sequence under a known or learned model, execute its first action, and plan again at the next step, as in model predictive control.

\begin{figure}[t]
    \centering
    \includegraphics[width=\textwidth]{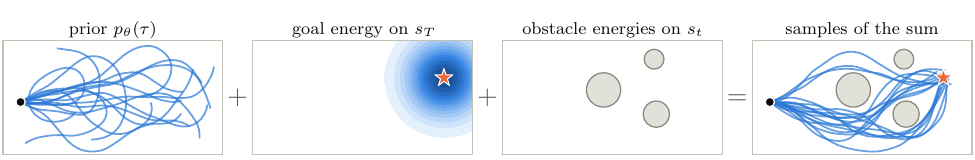}
    \caption{\textbf{Planning by composing trajectory energies.} The planning energy (\Eqref{eqn:plan_energy}) is a sum of terms that are independently specified: a prior over trajectories that start at the black dot (shown by its samples), an energy on the final state that is low near the goal (star; darker is lower), and an energy on every state that is high inside each obstacle (gray). Sampling the sum gives trajectories that reach the goal and avoid the obstacles (right), where trajectories are generated through MPPI with the prior.}
    \label{fig:planning}
\end{figure}
 
\paragraph{Sampling-Based Planners.} 
Planners that roll out a model are samplers or optimizers of~\Eqref{eqn:plan_ebm}. Model predictive path integral control~(MPPI)~\citep{williams2017model}, derived from path integral control~\citep{kappen2005linear,theodorou2010generalized}, draws $K$ action sequences from a Gaussian around the current plan, rolls them out, and moves the plan to their weighted average with weights $w_k\propto\exp\rbr{R(\tau_k)/\lambda}$. With the Gaussian as $p_0$, these are the weights of the self-normalized importance sampler of~\Secref{sec:eim}, so each iteration estimates the mean of $q^\star$. The cross-entropy method~\citep{rubinstein1999cross,deboer2005tutorial} replaces these weights by a hard selection, refitting the Gaussian to the sampled sequences with the highest returns. Iterating either method, with the refitted Gaussian as the next $p_0$, is mirror descent over action sequences~\citep{wagener2019online}, the trajectory analogue of the policy update~\Eqref{eqn:md_update}. For a differentiable model and reward, Langevin dynamics~\Eqref{eq:langevin} and gradient descent~\Eqref{eq:ebm_prediction_gd} on the action sequence, with gradients backpropagated through the rollout, respectively sample $q^\star$ and find a MAP plan.
 
\paragraph{Composing Trajectory Energies.} Taking the negative logarithm of~\Eqref{eqn:plan_ebm} writes the planning distribution as a sum of energies,
\begin{equation}\label{eqn:plan_energy}
    E(\tau) = \underbrace{-\sum_{t=0}^{T-1}\log P(s_{t+1}\mid s_t,a_t)}_{\text{dynamics}}\;
    \underbrace{-\sum_{t=0}^{T-1}\log\pi_0(a_t\mid s_t)}_{\text{behavior prior}}\;
    \underbrace{-\frac{1}{\lambda}\sum_{t=0}^{T-1}r(s_t,a_t)}_{\text{rewards and costs}}\;
    \underbrace{-\frac{1}{\lambda}\,r_T(s_T)}_{\text{goal}},
\end{equation}
which is an MRF~\Eqref{eq:mrf} over the trajectory whose factors can come from different sources: the dynamics from interaction data or a simulator, the behavior prior from demonstrations, and the rewards and goal from the task. For deterministic dynamics, the first term is a set energy~\Eqref{eqn:set_val} with $C\to\infty$ that restricts $\tau$ to dynamically feasible trajectories. Following~\Secref{sec:compose}, each term can be learned or specified separately and combined at inference time, and the task terms decompose further, \eg, into a goal energy on $s_T$ and one energy per obstacle on each $s_t$. This example, which is also discussed in the introduction, is illustrated in Figure~\ref{fig:planning}. An energy describing plausible robot trajectories, combined with energies expressing a destination and obstacle constraints enables us to form plans towards the goal that avoids obstacles. \citet{du2019model} learn an energy over state transitions in place of the dynamics and behavior terms, and plan by sampling state sequences from it together with start and goal constraints; planning in state space gives maximum-entropy inference of diverse plans and generalizes better to obstacles unseen during training.

\begin{figure}[t]
    \centering
    \includegraphics[width=\textwidth]{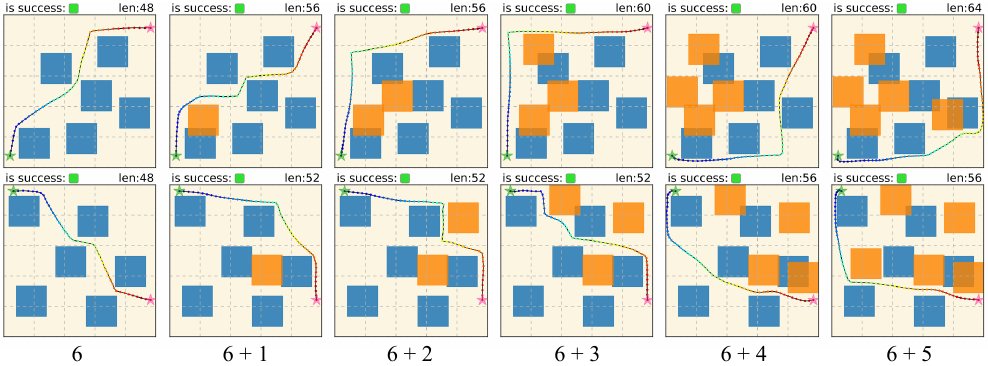}
    \caption{\textbf{Composing trajectory potentials for more obstacles.} A diffusion model learns a potential over plans in two-dimensional mazes with six obstacles. In a maze with $6+k$ obstacles (orange: the $k$ extra ones), the potentials conditioned on different groups of obstacles are added, and sampling the sum gives a collision-free plan from the start (green star) to the goal (pink star) without retraining. The two rows are different problems. Reproduced from~\citet{luo2024potential}.}
    \label{fig:potential_planning}
    \vspace{-10pt}
\end{figure}

\paragraph{Diffusion Planners.} Diffusion models can learn the dynamics and behavior terms of~\Eqref{eqn:plan_energy} jointly, as one prior over trajectories. Diffuser~\citep{janner2022planning} trains a diffusion model over state--action trajectories from offline data, \ie, a family of trajectory EBMs~(\Secref{sec:diffusion}), and plans by sampling the perturbed distribution $\tilde p(\tau)\propto p_\theta(\tau)\,h(\tau)$, where $h$ encodes rewards, goals, or constraints. Reward guidance adds the gradient of a learned return model $J_\phi(\tau)$ to the reverse process at every noise level, which is classifier guidance~\Eqref{eqn:cg} with the energy $-J_\phi$. It therefore targets the planning EBM~\Eqref{eqn:plan_ebm} with $1/\lambda$ given by the guidance scale and the learned trajectory distribution $p_\theta$ as the base measure, which keeps plans close to the trajectories that the data support. Goals are imposed by inpainting, \ie, by clamping the start and goal states, which conditions the joint energy as in~\Secref{sec:compose}. Decision Diffuser~\citep{ajay2023conditional} models state sequences, extracts actions with an inverse dynamics model, and conditions on returns, constraints, or skills by classifier-free guidance~\Eqref{eqn:cfg_poe}; conditioning on a single constraint or skill during training yields behavior at test time that satisfies several constraints together or composes skills, as in the conjunction~\Eqref{eqn:compose_bayes}. Potential-based diffusion motion planning~\citep{luo2024potential} learns potentials over motion-planning trajectories with a diffusion model and composes the potentials of different motion constraints by addition, as in classical potential-based planning. Since these planners compose by summing scores or energies at each noise level, the caveat of~\Secref{sec:diffusion} applies: reverse diffusion on summed scores only approximately samples the composed distribution, and MCMC on summed energies corrects it~\citep{du2023reduce}.

\paragraph{Sequence-Model Planners.} Sequence models such as Transformers can also serve as the prior over trajectories, treating states, actions, and returns as tokens. Autoregressive models factorize this prior in a fixed order~(\Secref{sec:ar}), and the planners built on them differ in how they bring in the reward. Decision Transformer~\citep{chen2021decision} conditions each action on the desired return-to-go and acts by conditioning on a high return, as Decision Diffuser does; with stochastic dynamics, this favors actions that were followed by lucky outcomes, the optimism discussed after~\Eqref{eqn:plan_ebm}. Trajectory Transformer~\citep{janner2021offline} plans by beam search over the model's likely continuations, keeping those with the highest predicted return, which is the greedy version of the sequential samplers of~\Secref{sec:itc_samplers}. Both generate a plan one step at a time, so they cannot revise earlier actions in light of later ones. \citet{chen2023planning} instead train a masked language model over action sequences, which defines an implicit energy over the whole plan, as in BERT~(\Secref{sec:bert}), and plan by iteratively resampling actions to lower it, in the manner of the Gibbs sampler of~\Secref{sec:inference}, so that any step can be revised and constraints can be added as energy terms.

\paragraph{Planning with Generative World Models.} When states are images, the trajectory energy can be split between a model of observation sequences and a model of actions. UniPi~\citep{du2023universal} generates a video plan with a text-conditioned video diffusion model, conditioned on the current frame and a language description $g$ of the goal, and extracts actions with an inverse dynamics model,
\begin{equation}\label{eqn:plan_video}
    p(o_{1:T}, a_{0:T-1}\mid o_0, g) = p_\theta(o_{1:T}\mid o_0, g)\prod_{t=0}^{T-1}p_\psi(a_t\mid o_t, o_{t+1}),
\end{equation}
where $o_t$ is the observation at step $t$. The video model, trained on broad data across tasks, plays the role of the dynamics and behavior terms of~\Eqref{eqn:plan_energy}, the text plays the role of the goal, and only the small inverse dynamics model depends on the action space of the robot. Compositional foundation models for hierarchical planning~\citep{ajay2023compositional} compose a language model that proposes subgoals, a video model that grounds them in visual plans, and an inverse dynamics model that turns the videos into actions, and enforce consistency among the three by iterative refinement. Video language planning~\citep{du2024video} searches over long-horizon video plans with a tree search, using vision--language models as the policy and value function and a text-to-video model as the dynamics. In each case, separately trained models supply different factors of one distribution over trajectories, which is the sense in which learned dynamics, behavior models, and task objectives contribute components of a common inference problem.

\begin{figure}[t]
    \centering
    \includegraphics[width=\textwidth]{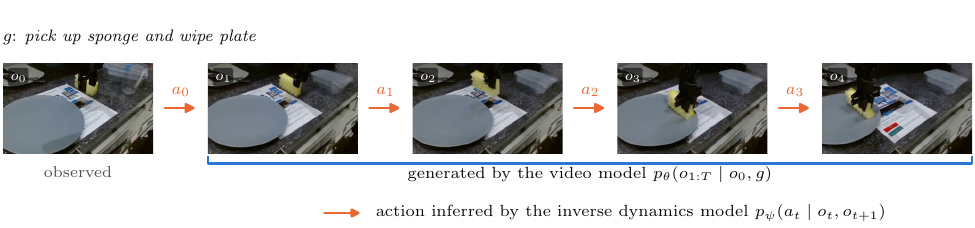}
    \caption{\textbf{Planning with a video model.} The trajectory distribution~\Eqref{eqn:plan_video} factors into a video model, which turns the current frame $o_0$ and the instruction $g$ into a video plan $o_{1:T}$ (blue), and an inverse dynamics model, which infers each action $a_t$ from a pair of consecutive frames (orange). The video model, trained across tasks, supplies the dynamics and behavior terms of~\Eqref{eqn:plan_energy}, and only the inverse dynamics model depends on the robot. Frames from~\citet{du2023universal}, where $o_1,\ldots,o_4$ are generated.}
    \label{fig:unipi}
    \vspace{-5pt}
\end{figure}
 
\paragraph{Planning versus Policies.} Policies and planners represent the same EBM at different levels. For deterministic dynamics, the trajectory EBM~\Eqref{eqn:plan_ebm} factorizes along the trajectory into the soft-optimal policy given by~\Eqref{eqn:soft_bellman}, as shown for maximum-entropy IRL in~\Secref{sec:irl}. A policy amortizes planning into one network for one reward and acts in a single forward pass. A planner keeps the terms of~\Eqref{eqn:plan_energy} separate and pays with sampling or optimization at every decision, but a new goal, cost, or constraint only requires a new energy term rather than a new policy. This is the trade-off between amortized sampling and MCMC in~\Secref{sec:inference}, and between learning the policy and learning the reward in~\Secref{sec:irl}, now applied to acting.

\section{Inference-Time Computation}\label{sec:inference_time}

A recent source of improvement in large models is to spend more computation per input at inference time rather than at training time. In the EBM view, these procedures are samplers or optimizers for one distribution, the composition of a pretrained generator with a verifier, which is the KL-regularized EBM of~\Secref{sec:reg_rl}. We first describe this target, then view common inference-time methods as the samplers of~\Secref{sec:inference},~\Secref{sec:eim}, and~\Secref{sec:planning}, then show how composing learned energies supports reasoning on problems larger than those seen in training, and finally relate inference-time computation to fine-tuning, which amortizes the same sampler.

\subsection{Verifiers Define the Target}\label{sec:itc_target}

Let $\pi_0(\vy\mid\vx)$ be a pretrained generator, such as a language model that answers a prompt $\vx$ with a response $\vy = (y_1,\ldots,y_T)$, and let $r(\vx,\vy)$ be a verifier that scores complete responses, such as a learned reward model, a learned verifier~\citep{cobbe2021training}, or a checker such as unit tests. The target of inference-time computation is the residual EBM
\begin{equation}\label{eqn:itc_target}
    \pi^\star(\vy\mid\vx) = \frac{\pi_0(\vy\mid\vx)\exp\rbr{r(\vx,\vy)/\lambda}}{Z(\vx)}, \qquad Z(\vx) = \EE_{\vy\sim\pi_0(\cdot\mid\vx)}\sbr{\exp\rbr{r(\vx,\vy)/\lambda}},
\end{equation}
which is the optimal policy of the KL-regularized objective~\Eqref{eqn:rlhf_ebm}, with the prompt as the state and the response as the action. The temperature $\lambda$ trades the verifier against the generator. For a binary verifier, $r\in\cbr{0,1}$, the limit $\lambda\to0$ gives $\pi^\star(\vy\mid\vx) = \pi_0(\vy\mid\vx, r=1)$, the generator conditioned on being correct, which is the product of $\pi_0$ with the set energy~\Eqref{eqn:set_val} of correct responses, the logical \texttt{AND} of~\Secref{sec:compose}. Several verifiers compose in the same way, as $\pi_0\prod_k\exp\rbr{r_k/\lambda_k}$, \eg, for a correctness check and a style or safety reward. The verifier can even be the generator's own likelihood, which sharpens the generator without any external signal, as discussed in~\Secref{sec:itc_samplers}. Since verifiers are learned or incomplete, the temperature also limits how far the target moves from the generator, as measured by $\mathrm{KL}\rbr{\pi^\star\|\pi_0}$, and thus how much an inaccurate verifier can be exploited. This is the requirement of~\Secref{sec:compose} that each component be accurate where the composition places its mass. The target is a distribution, and the prediction can be a sample from it or its mode.

\begin{figure}[t]
    \centering
    \includegraphics[width=\textwidth]{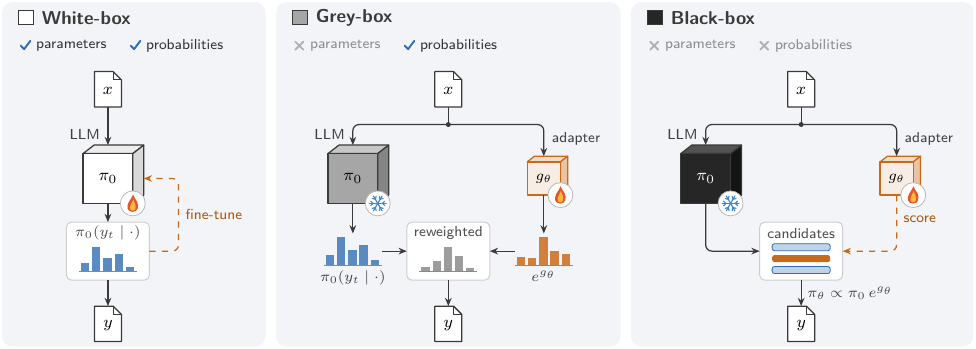}
    \caption{\textbf{White-box, grey-box, and black-box adaptation of a language model $\pi_0$.} White-box adaptation accesses both the parameters and the output probabilities of $\pi_0$ and can fine-tune it. Grey-box adaptation accesses only the token probabilities, which a small adapter $g_\theta$ reweights. Black-box adaptation accesses neither and only receives sampled text: $\pi_0$ proposes candidate sentences and $g_\theta$ scores them to decide which to extend, a sentence-level beam search for $\pi_\theta\propto\pi_0\,e^{g_\theta}$~\Eqref{eqn:itc_target}. The flame marks trainable models and the snowflake marks fixed, inaccessible ones. Adapted from~\citet{sun2024bbox}.}
    \label{fig:bbox}
    \vspace{-10pt}
\end{figure}

\paragraph{Black-Box Adaptation.} The target~\Eqref{eqn:itc_target} requires only samples from the generator and evaluations of the verifier, not the parameters or the token probabilities of $\pi_0$. This enables us to directly adapt models, even when the strongest generators are often served only through an API, as black boxes. As illustrated in~\Figref{fig:bbox}, white-box adaptation fine-tunes the parameters of $\pi_0$, and grey-box adaptation reweights its output token probabilities, but neither is available when the API returns only sampled text. The residual EBM still applies: BBox-Adapter~\citep{sun2024bbox} keeps $\pi_0$ fixed as the base measure and learns the energy as a small language model $g_\theta(\vx,\vy)$, \ie, $r/\lambda = g_\theta$ in~\Eqref{eqn:itc_target}, so that $\pi_\theta(\vy\mid\vx)\propto\pi_0(\vy\mid\vx)\exp\rbr{g_\theta(\vx,\vy)}$. The energy is trained by ranking NCE~\Eqref{eqn:lnce_rank} with the generator itself as the noise distribution: target-domain responses are the positives, the responses of $\pi_0$ are the negatives, and since the model is a residual EBM with base measure $p_n = \pi_0$, the unknown $\log\pi_0$ cancels from the logits, so its probabilities are never needed. The positives can come from ground truth or from human or AI feedback, and the negatives from previous adapted generations, which turns adaptation into an online loop that updates the verifier and the sampler together. Sampling from $\pi_\theta$ also uses $\pi_0$ only as a proposal: $\pi_0$ proposes candidate continuations and $g_\theta$ scores them in a sentence-level beam search, a sequential variant of the samplers in~\Secref{sec:itc_samplers}. The same construction adapts black-box text-to-video models: a small video model trained on domain-specific data plays the role of the energy on top of a large pretrained model that is accessed only through its score, without its weights~\citep{yang2024probabilistic}.

\subsection{Samplers for the Target}\label{sec:itc_samplers}

\paragraph{Sampling Complete Responses.} The simplest methods draw $N$ complete responses $\vy_1,\ldots,\vy_N\sim\pi_0(\cdot\mid\vx)$ and use the verifier only to select among them. Returning $\vy_i$ with probability proportional to $\exp\rbr{r(\vx,\vy_i)/\lambda}$ is the self-normalized importance sampler of~\Secref{sec:eim} with the generator as the proposal: it samples a distribution that interpolates between $\pi_0$ and $\pi^\star$, and converges to $\pi^\star$ as $N\to\infty$. Best-of-$N$, which returns the response with the highest reward, is also a residual EBM. If the reward of $\vy\sim\pi_0(\cdot\mid\vx)$ has a continuous distribution with cumulative distribution function $F_\vx$, the returned response is distributed as~\citep{gui2024bonbon}
\begin{equation}\label{eqn:itc_bon}
    \pi_{\text{BoN}}(\vy\mid\vx) = N\,\pi_0(\vy\mid\vx)\,F_\vx\rbr{r(\vx,\vy)}^{N-1},
\end{equation}
since $\vy$ is returned when it is one of the $N$ draws and the other $N-1$ draws have lower rewards. This is the KL-regularized optimum~\Eqref{eqn:itc_target} for the transformed reward $\log F_\vx(r)$, the $\log$-quantile of the reward under the generator, at temperature $\lambda = 1/(N-1)$. Increasing $N$ lowers the temperature, and the divergence from the generator is at most $\log N - (N-1)/N$, with equality for continuous rewards~\citep{beirami2025theoretical}. Rejection sampling gives exact samples instead: when $r\le r_{\max}$, accepting $\vy\sim\pi_0(\cdot\mid\vx)$ with probability $\exp\rbr{(r(\vx,\vy) - r_{\max})/\lambda}$ returns samples of~\Eqref{eqn:itc_target}. For a binary verifier with $\lambda\to0$, it resamples until the verifier accepts, at an expected cost of $1/\EE_{\vy\sim\pi_0(\cdot\mid\vx)}\sbr{r(\vx,\vy)}$ responses. Finally, majority voting over final answers, or self-consistency~\citep{wang2023self}, is a Bayes prediction. Let $a(\vy)$ be the final answer extracted from a response. The most frequent answer among the samples estimates the mode of the marginal distribution $p(a\mid\vx) = \sum_{\vy}\pi_0(\vy\mid\vx)\,\1[a(\vy) = a]$, which is the Bayes prediction under the $0$--$1$ loss on answers. It sums over the different responses that reach each answer, rather than taking the answer of the single most probable response, and weighting the votes by $\exp\rbr{r/\lambda}$ gives the same prediction under $\pi^\star$.

\paragraph{Sequential Sampling and Search.} Sampling complete responses wastes computation on responses that go wrong early. Since $\pi_0$ is autoregressive, the target~\Eqref{eqn:itc_target} also factorizes over tokens, with the conditionals
\begin{equation}\label{eqn:itc_twist}
    \pi^\star(y_t\mid\vx,\vy_{<t}) = \pi_0(y_t\mid\vx,\vy_{<t})\,\frac{\psi_t(\vx,\vy_{\le t})}{\psi_{t-1}(\vx,\vy_{<t})}, \qquad
    \psi_t(\vx,\vy_{\le t}) = \EE_{\vy_{>t}\sim\pi_0(\cdot\mid\vx,\vy_{\le t})}\sbr{\exp\rbr{r(\vx,\vy)/\lambda}},
\end{equation}
where $\psi_t$ is the expected exponentiated reward of completing the prefix with the generator, and $\psi_0 = Z(\vx)$. This is the fixed-order factorization of an EBM~\Eqref{eqn:ebm_chain}: $\pi_0(\vy_{\le t}\mid\vx)\,\psi_t(\vx,\vy_{\le t})$ is the sum over the unvisited tokens that autoregressive models reparameterize in~\Eqref{eqn:ar_reparam}. It is also the soft-optimal policy of the KL-regularized MDP of~\Secref{sec:reg_rl} whose state is the prefix, whose action appends a token, and whose reward $r$ arrives at the end, with $\lambda\log\psi_t$ as its soft value. Exact token-by-token sampling therefore needs $\psi_t$, which is as intractable as $Z(\vx)$, and practical methods approximate it. A future discriminator, which predicts an attribute of the completion from a prefix~\citep{yang2021fudge}, learns $\psi_t$ for a classifier energy, which is the autoregressive counterpart of classifier guidance~\Eqref{eqn:cg}. Value models and process reward models, which score partial solutions~\citep{lightman2024lets}, play the same role; for a binary verifier with $\lambda\to0$, $\psi_t$ becomes, up to a constant, the probability that the generator completes the prefix correctly. With an approximate $\psi_t$, sequential Monte Carlo~(SMC) recovers $\pi^\star$ as the number of particles grows: it extends $K$ partial responses in parallel, weights each extension by the ratio of successive approximate values, as in~\Eqref{eqn:itc_twist}, resamples the partial responses according to their weights, and corrects with the true reward at the end~\citep{lew2023sequential,zhao2024probabilistic}. Beam search keeps the $K$ partial responses with the highest values, a greedy and deterministic version of this resampling, and tree search adds backtracking, as in the planners of~\Secref{sec:planning}.

\paragraph{Composing Models Token by Token.} The same factorization explains a pitfall of composing autoregressive models. Sampling each token from the normalized product of the conditionals of two models gives
\begin{equation}\label{eqn:itc_token_product}
    q(\vy\mid\vx) = \prod_{t=1}^{T}\frac{p_1(y_t\mid\vx,\vy_{<t})\,p_2(y_t\mid\vx,\vy_{<t})}{Z_t(\vx,\vy_{<t})} = \frac{p_1(\vy\mid\vx)\,p_2(\vy\mid\vx)}{\prod_{t=1}^{T}Z_t(\vx,\vy_{<t})},
\end{equation}
where $Z_t(\vx,\vy_{<t}) = \sum_{v}p_1(v\mid\vx,\vy_{<t})\,p_2(v\mid\vx,\vy_{<t})$. The per-token normalizers depend on the prefix, so $q$ is not the product $p_1p_2$ of~\Secref{sec:compose}, for the same reason that summed scores do not give the score of a noised product in~\Secref{sec:diffusion}: normalizing at every step does not commute with taking products. The weights $\prod_tZ_t(\vx,\vy_{<t})$ are available as the tokens are drawn, so importance sampling or SMC with these weights is needed to remove the bias. Existing decoding-time methods use such token-level compositions directly, without this correction: DExperts~\citep{liu2021dexperts} multiplies the base model by an expert and divides it by an anti-expert, an inversion~\Eqref{eqn:compose_inversion}, and contrastive decoding~\citep{li2023contrastive} scores tokens by the ratio of a large model to a small one, restricted to the tokens the large model finds plausible, which is an inversion combined with a set energy~\Eqref{eqn:set_val}.

\begin{figure}[t]
    \centering
    \includegraphics[width=\textwidth]{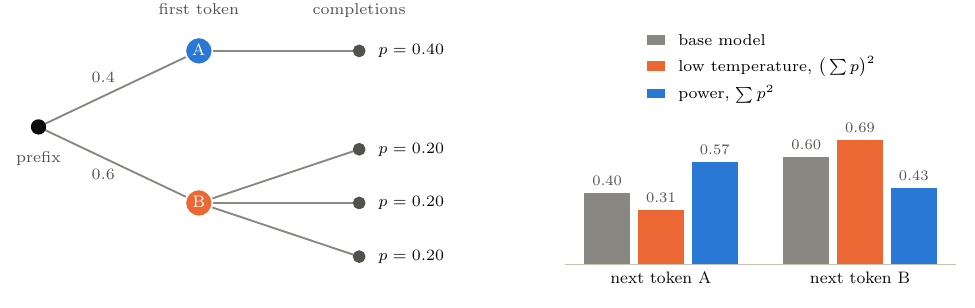}
    \vspace{-20pt}
    \caption{\textbf{Power distribution vs low-temperature sampling.} The power distribution and low-temperature decoding choose the next token differently, here with $\alpha = 2$. Token A has one completion with probability $0.4$, and token B has three completions with probability $0.2$ each. Low temperature sharpens the total probability of each token, $\rbr{\sum p}^2$, and prefers B; the power distribution sums the sharpened probabilities of the completions, $\sum p^2$, and prefers A. }
    \label{fig:power}
\end{figure}

\begin{figure}[t]
    \centering
    \includegraphics[width=0.9\textwidth]{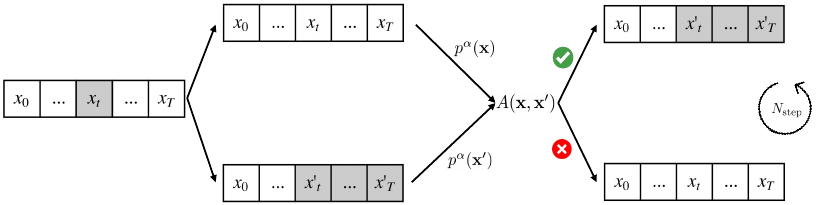}
    \caption{\textbf{Metropolis--Hastings for the power distribution (\Eqref{eqn:itc_power}).} Here the response is $\vx = (x_0,\ldots,x_T)$ and the generator is $p$ ($\vy$ and $\pi_0$ in the text). A position $t$ is chosen uniformly, the tokens from $t$ onward are resampled with a proposal model (shaded), and the candidate $\vx'$ replaces $\vx$ with the acceptance probability $A(\vx,\vx')$ of~\Eqref{eq:mh_accept}, which needs $p^\alpha$ only up to normalization; the step is repeated $N_{\mathrm{step}}$ times. Figure from~\citet{karan2025reasoning}.}
    \label{fig:power_mh}
     \vspace{-10pt}
\end{figure}

\paragraph{Sharpening without a Verifier.} The verifier can also be the generator itself. Taking $r(\vx,\vy) = \log\pi_0(\vy\mid\vx)$ and $\lambda = 1/(\alpha-1)$ in~\Eqref{eqn:itc_target} gives the power distribution
\begin{equation}\label{eqn:itc_power}
    \pi_\alpha(\vy\mid\vx) \propto \pi_0(\vy\mid\vx)^{\alpha}, \qquad \alpha>1,
\end{equation}
which is the EBM of the generator at the lower temperature $1/\alpha$, or, for integer $\alpha$, the product of $\alpha$ copies of the generator~(\Secref{sec:compose}). \citet{karan2025reasoning} show that sampling from~\Eqref{eqn:itc_power} elicits reasoning from base language models without additional training or a verifier. Its single-sample accuracy is comparable to that of reinforcement learning with GRPO on the domain used for post-training and its accuracy over multiple samples is higher, since it avoids the loss of diversity that RL post-training causes. Low-temperature decoding, which samples each token from $\pi_0(y_t\mid\vx,\vy_{<t})^{\alpha}$ after normalization, does not sample~\Eqref{eqn:itc_power}, for the same reason as the token-level product~\Eqref{eqn:itc_token_product}. Writing the conditionals through the sum over completions in~\Eqref{eqn:itc_twist}, the two choose the next token according to
\begin{equation}\label{eqn:itc_power_vs_temp}
    \pi_\alpha(y_t\mid\vx,\vy_{<t}) \propto \sum_{\vy_{>t}}\pi_0(\vy\mid\vx)^{\alpha}
    \qquad\text{and}\qquad
    \pi_0(y_t\mid\vx,\vy_{<t})^{\alpha} \propto \rbr{\sum_{\vy_{>t}}\pi_0(\vy\mid\vx)}^{\alpha},
\end{equation}
respectively. That is, the power distribution sums the sharpened likelihoods of the completions, whereas low-temperature decoding sharpens the sum of their likelihoods. The power distribution therefore prefers tokens with a few high-likelihood completions, and low-temperature decoding prefers tokens whose probability is spread over many low-likelihood completions~\citep{karan2025reasoning},  as illustrated in the example of Figure~\ref{fig:power}. The sum on the left equals $\pi_0(\vy_{\le t}\mid\vx)\,\psi_t(\vx,\vy_{\le t})$ in~\Eqref{eqn:itc_twist} for this reward and is intractable. Instead of approximating it, \citet{karan2025reasoning} sample~\Eqref{eqn:itc_power} with Metropolis--Hastings~\Eqref{eq:mh_accept}, illustrated in Figure~\ref{fig:power_mh}. The proposal resamples the response from a uniformly chosen position onward with a proposal language model, with the acceptance probability dependent only on likelihoods of the current and proposed responses under the generator and the proposal, since the partition function cancels, and the MCMC steps are applied progressively as the response is generated. In the terms of~\Secref{sec:itc_amortize}, RL fine-tuning fits a mode-seeking approximation to a sharpened distribution, whereas MCMC samples the sharpened distribution itself, which is consistent with the diversity that power sampling retains.

\paragraph{Reasoning by Energy Minimization.} When the verifier is itself a learned energy over complete outputs, $E_\theta(\vx,\vy)$, inference can refine a candidate rather than only select among candidates, by minimizing the energy~\Eqref{eq:ebm_prediction_gd}. IREM~\citep{du2022learning} trains an energy so that a few steps of gradient descent on it reach the answer~(\Secref{sec:learning}), and on algorithmic tasks it solves harder problems than those seen in training by running more steps at test time. IRED~\citep{du2024learning} learns a sequence of annealed energy landscapes, supervised through both their scores and their energies, optimizes from the smoothest landscape to the sharpest, and adapts the number of steps to the difficulty of each problem, which lets it solve harder Sudoku puzzles and plan paths in larger graphs than in training. Energy-Based Transformers~\citep{gladstone2026energy} learn such energies over inputs and candidate predictions at the scale of language and vision models, and use extra minimization steps and comparisons between candidates as a form of deliberate thinking. Inference-time computation then has two axes: more minimization steps refine each candidate, and more restarts, compared by their energies, select among candidates, which is best-of-$N$ with $r = -E_\theta$. The same inference-time computation applies to diffusion models, where a verifier enters as classifier guidance~\Eqref{eqn:cg}. The two axes become local and global search~\citep{zhang2026inference}: more MCMC steps on the summed energy of the model and the verifier at each noise level refine each sample~\citep{du2023reduce}, and more branches of the denoising process, compared by the verifier, select among samples through breadth-first or depth-first tree search.

\subsection{Compositional Reasoning}\label{sec:itc_compose}

\begin{figure}[t]
    \centering
    \includegraphics[width=\textwidth]{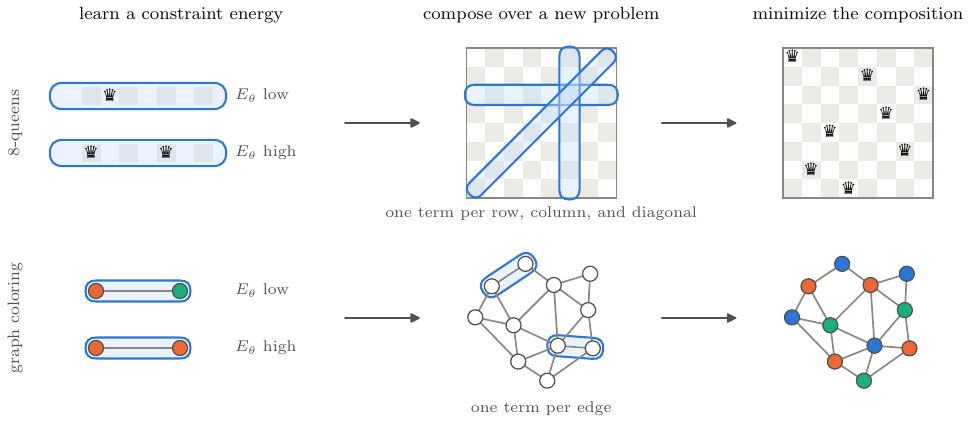}
    \caption{\textbf{Compositional reasoning by energy minimization.} An energy $E_\theta$ is trained on a single constraint, a row of a chessboard that must hold one queen or an edge whose endpoints must differ in color, to be low on valid assignments and high on invalid ones (left). For a new problem, the energies of all its constraints are summed as in~\Eqref{eqn:itc_csp}: one term for each row, column, and diagonal of an $8\times8$ board, or for each edge of a larger graph, with outlines marking the variables of a few of the terms (middle). Minimizing the sum solves the problem (right). Adapted from~\citet{oarga2025generalizable}.}
    \label{fig:compositional_reasoning}
    \vspace{-15pt}
\end{figure}

So far, verifiers and learned energies have scored complete solutions, and harder problems have been met by spending more computation on the same energy, as IREM and IRED do by running more minimization steps. Many reasoning problems, however, are defined by constraints on parts of their solutions. Each clause of a 3-SAT problem involves three variables, a valid coloring of a graph gives different colors to the two endpoints of every edge, and a solution of the $N$-queens problem places one queen in every row and column and at most one on every diagonal. A network trained end to end to map problems to solutions must learn this whole map from examples, and it often fails on problems larger than those it was trained on. The energy view suggests a different decomposition, illustrated in Figure~\ref{fig:compositional_reasoning}: learn an energy for each kind of constraint, compose the energies of the constraints that appear in a new problem, and solve the problem by minimizing the composition~\citep{oarga2025generalizable}.

\paragraph{Composed Energies.} We write a problem $\vx$ as a set of constraints $c_1,\ldots,c_K$, where $c_k$ involves the variables $\vy_{S_k}$ of the solution, such as the two endpoints of an edge or the three variables of a clause. An energy $E_\theta(c,\vy_S)$ that is low on the assignments satisfying a single constraint is composed into the energy of the whole problem,
\begin{equation}\label{eqn:itc_csp}
    E(\vx,\vy) = \sum_{k=1}^{K} E_\theta\rbr{c_k,\vy_{S_k}},
\end{equation}
which is an MRF~\Eqref{eq:mrf} whose factors share one learned energy and whose graph is given by the problem. When a solution exists, the minimizers of~\Eqref{eqn:itc_csp} satisfy every constraint, which is the logical \texttt{AND}~\Eqref{eqn:compose_and} of~\Secref{sec:compose}. A problem with more variables and constraints than any training problem is a longer sum of the same terms, so training needs to cover the kinds of constraints rather than their combinations, which is the sample-complexity argument of~\Secref{sec:compose} applied to reasoning. \citet{oarga2025generalizable} train such energies on single clauses of 3-SAT, single edges of graph coloring, and the rows of one $N$-queens solution, using the energy-parameterized denoising objective of~\Secref{sec:diffusion} together with a contrastive loss that raises the energy of invalid assignments~(\Secref{sec:learning}). They compose the energies over whole problems, including SAT instances and graphs larger than any used in training, and, since a sum of separately trained energies can have spurious minima, they can fine-tune the composition on complete solutions.

\paragraph{Minimizing the Composition.} The composed energy is harder to minimize than any of its terms: its minima are where all terms agree, it is non-convex, and gradient descent or Langevin dynamics~\Eqref{eq:langevin} from a single initialization is easily trapped. \citet{oarga2025generalizable} minimize it with parallel energy minimization~(PEM), an SMC procedure over the noise levels of the energy: a population of candidates starts from noise, and at each level, the candidates are resampled in proportion to $\exp\rbr{-E}$, perturbed with noise, and moved by a gradient step on the energy at that level. Resampling discards candidates in poor local minima, and the noise keeps the population spread over different solutions. More candidates solve more problems, so the computation spent at inference can grow with the difficulty of a problem, which a network that predicts the solution in a single pass cannot do. With this sampler, the composed energies outperform learned solvers trained end to end on most of the $N$-queens, 3-SAT, and graph-coloring benchmarks considered, including their larger instances~\citep{oarga2025generalizable}.

\paragraph{Relation to Planning and Verification.} The composed energy~\Eqref{eqn:itc_csp} has the structure of the trajectory energies of~\Secref{sec:planning}, which are composed from dynamics, goals, and obstacles, and of the constraint solvers composed from diffusion models~\citep{yang2023compositional}: in each, the factors are learned separately and the graph is chosen for each problem. In the terms of~\Secref{sec:itc_target}, each factor is a local verifier, and their sum is a verifier for the whole problem that was never trained on a whole problem. Since this verifier is a differentiable energy, it can refine candidates as well as select them, and the samplers of~\Secref{sec:itc_samplers} apply to it directly.

\subsection{Amortizing Inference-Time Computation}\label{sec:itc_amortize}

Inference-time methods spend computation on each new input to draw
approximate samples from the target distribution in~\Eqref{eqn:itc_target},
\[
    \pi^\star(\vy\mid\vx)
    =
    \frac{\pi_0(\vy\mid\vx)\exp\rbr{r(\vx,\vy)/\lambda}}{Z(\vx)},
    \qquad
    Z(\vx) = \EE_{\vy\sim\pi_0(\cdot\mid\vx)}
    \sbr{\exp\rbr{r(\vx,\vy)/\lambda}}.
\]
This distribution reweights the pretrained model $\pi_0$ toward
responses with higher verifier scores. Fine-tuning amortizes this
computation into the generator's weights: it learns a model
$\pi_\theta$ that approximates the target across inputs, allowing
high-reward responses to be generated with less search at inference
time. Below, we discuss three approaches to learning such a model.

\paragraph{RL Fine-Tuning.}
KL-regularized RL rewards high verifier scores while keeping the model
close to $\pi_0$. Substituting the definition of $\pi^\star$ into its
objective gives
\begin{equation}\label{eqn:itc_rl}
    \EE_{\vy\sim\pi_\theta(\cdot\mid\vx)}\sbr{r(\vx,\vy)}
    - \lambda\,\mathrm{KL}\rbr{\pi_\theta\,\|\,\pi_0}
    = -\lambda\,\mathrm{KL}\rbr{\pi_\theta\,\|\,\pi^\star}
    + \lambda\log Z(\vx).
\end{equation}
Since $Z(\vx)$ does not depend on $\theta$, maximizing this objective
is equivalent to minimizing the reverse KL to the target, as in
variational inference~(\Eqref{eq:vi_kl}) and policy
learning~(\Eqref{eqn:vi_policy}). This objective is often mode-seeking:
when $\pi_\theta$ cannot match $\pi^\star$ exactly, it may concentrate
on a few high-probability regions of the target.

Practical algorithms build on this objective.
GRPO~\citep{shao2024deepseekmath}, for example, uses REINFORCE-style
policy gradients~(\Eqref{eq:reinforce_grad}) with rewards normalized
within a group of responses to the same prompt. It subtracts the
group's mean reward and divides by its standard deviation, then
combines these normalized rewards with PPO-style clipping and a KL
penalty to the reference model.

\paragraph{Rejection-Sampling Fine-Tuning.}
A second approach uses the verifier to construct a training dataset.
For each input, we sample several responses from $\pi_0$, discard
those that fail the verifier, and train the model on those that pass.
This is the basic idea behind methods such as
STaR~\citep{zelikman2022star} and RFT~\citep{yuan2023scaling}, and it
amortizes the rejection sampler of~\Secref{sec:itc_samplers} into the
weights of $\pi_\theta$. For a binary verifier, $r\in\cbr{0,1}$, the
accepted responses are exact samples of
$\pi^\star(\vy\mid\vx) = \pi_0(\vy\mid\vx, r=1)$, the $\lambda\to0$
limit of the target~(\Secref{sec:itc_target}), which keeps the
relative probabilities that $\pi_0$ assigns to correct responses and
removes the incorrect ones. The expected $\log$-likelihood of these
responses is
\begin{equation}\label{eqn:itc_rft}
    \EE_{\vy\sim\pi^\star(\cdot\mid\vx)}\sbr{\log\pi_\theta(\vy\mid\vx)}
    = -\mathrm{KL}\rbr{\pi^\star\,\|\,\pi_\theta} - H\rbr{\pi^\star}.
\end{equation}
Since $H\rbr{\pi^\star}$ does not depend on $\theta$, maximum-likelihood
training minimizes the forward KL to $\pi^\star$. For a binary
verifier, the target of RL fine-tuning in~\Eqref{eqn:itc_rl} tends to
the same $\pi^\star$ as $\lambda\to0$, but optimizes the KL in the opposite direction. If a fixed number of responses is sampled per prompt
and the accepted ones are pooled, each prompt is weighted by its pass
rate under $\pi_0$, so prompts that $\pi_0$ does not solve are not trained on. The forward KL is mass-covering and when $\pi_\theta$
cannot match $\pi^\star$ exactly, it spreads its probability over all
the successful solutions found during sampling, and may also place
probability on responses that fail. In contrast, the reverse KL of RL fine-tuning
avoids failing responses but may drop some solutions.

\paragraph{Best-of-$N$ Distillation.}
A third approach takes the best-of-$N$ policy~\Eqref{eqn:itc_bon},
itself a residual EBM, as the target. BOND~\citep{sessa2024bond} fits
$\pi_\theta$ to $\pi_{\text{BoN}}$ by minimizing the Jeffreys
divergence, a weighted sum of the forward and reverse KL,
\begin{equation}\label{eqn:itc_bond}
    \min_\theta\;(1-\alpha)\,\mathrm{KL}\rbr{\pi_{\text{BoN}}\,\|\,\pi_\theta}
    + \alpha\,\mathrm{KL}\rbr{\pi_\theta\,\|\,\pi_{\text{BoN}}},
    \qquad \alpha\in[0,1],
\end{equation}
which balances the mass-covering and mode-seeking tendencies of the
two approaches above, so that a single sample of $\pi_\theta$ emulates
best-of-$N$. The forward term is supervised fine-tuning on
best-of-$N$ samples of $\pi_0$, with gradient
$-\EE_{\vy\sim\pi_{\text{BoN}}}\sbr{\nabla_\theta\log\pi_\theta(\vy\mid\vx)}$.
By~\Eqref{eqn:itc_bon},
$\log\pi_{\text{BoN}} = \log\pi_0 + (N-1)\log F_\vx(r) + \log N$, so up
to a constant the reverse term is $-(N-1)$ times the RL
objective~\Eqref{eqn:itc_rl} with the reward $\log F_\vx(r)$ and
$\lambda = 1/(N-1)$. It is therefore minimized by KL-regularized RL
with the $\log$-quantile as the reward, where the quantile
$F_\vx(r(\vx,\vy))$, the fraction of samples of $\pi_0$ with reward at
most $r(\vx,\vy)$, is estimated by Monte Carlo. Iterative BOND repeats this distillation, each round
distilling the best-of-$N$ of a moving anchor policy, the model of the
previous round, so that each round applies the
transformation~\Eqref{eqn:itc_bon} to the previous model.
\citet{yang2025faster} show that this iteration converges to the
equilibrium of a game between policies, and replace it with the win
rate dominance~(WIND) fixed point
\begin{equation}\label{eqn:itc_wind}
    \pi^\dagger(\vy\mid\vx) \propto
    \pi_0(\vy\mid\vx)\exp\rbr{w^\dagger(\vx,\vy)/\lambda},
    \qquad
    w^\dagger(\vx,\vy) = \EE_{\vy'\sim\pi^\dagger(\cdot\mid\vx)}
    \sbr{\mathbf{1}\cbr{r(\vx,\vy)\ge r(\vx,\vy')}},
\end{equation}
a residual EBM whose energy is the win rate of $\vy$ against the
policy itself, which they solve with more sample-efficient
algorithms.

\paragraph{Fine-Tuned Models as Proposals.}
A fine-tuned model can also serve as a proposal for the samplers
of~\Secref{sec:itc_samplers}. To use $\pi_\theta$ as an
importance-sampling proposal for the original target, each candidate
receives weight
\[
    \omega(\vy,\vx)
    =
    \frac{\pi_0(\vy\mid\vx)\exp\rbr{r(\vx,\vy)/\lambda}}
         {\pi_\theta(\vy\mid\vx)}.
\]
These weights correct for the change in proposal, provided
$\pi_\theta$ assigns nonzero probability wherever $\pi^\star$ does.
When $\pi_\theta=\pi^\star$, every weight equals $Z(\vx)$:
the generator already samples from the target.
More generally, proposals that cover the target and produce less
variable weights can reduce the number of candidates needed.
A proposal concentrated on only a few solutions may rarely generate
other responses favored by the target.

\paragraph{Tradeoff Between Finetuning and Inference.}
More broadly, fine-tuning pays a training cost to reduce search across many later
inputs, sometimes making a single sample sufficient.
Inference-time methods spend computation on each input but can adapt
to a new verifier without retraining the generator.
The same trade-off appears between learned samplers and MCMC
in~\Secref{sec:inference}, and between policies and planners
in~\Secref{sec:planning}: computation can be spent during training
to make later inference cheaper, or during inference to adapt to
the problem at hand.

\subsection{Best-of-$N$ Sampling-Aware Fine-Tuning}\label{sec:itc_bon}

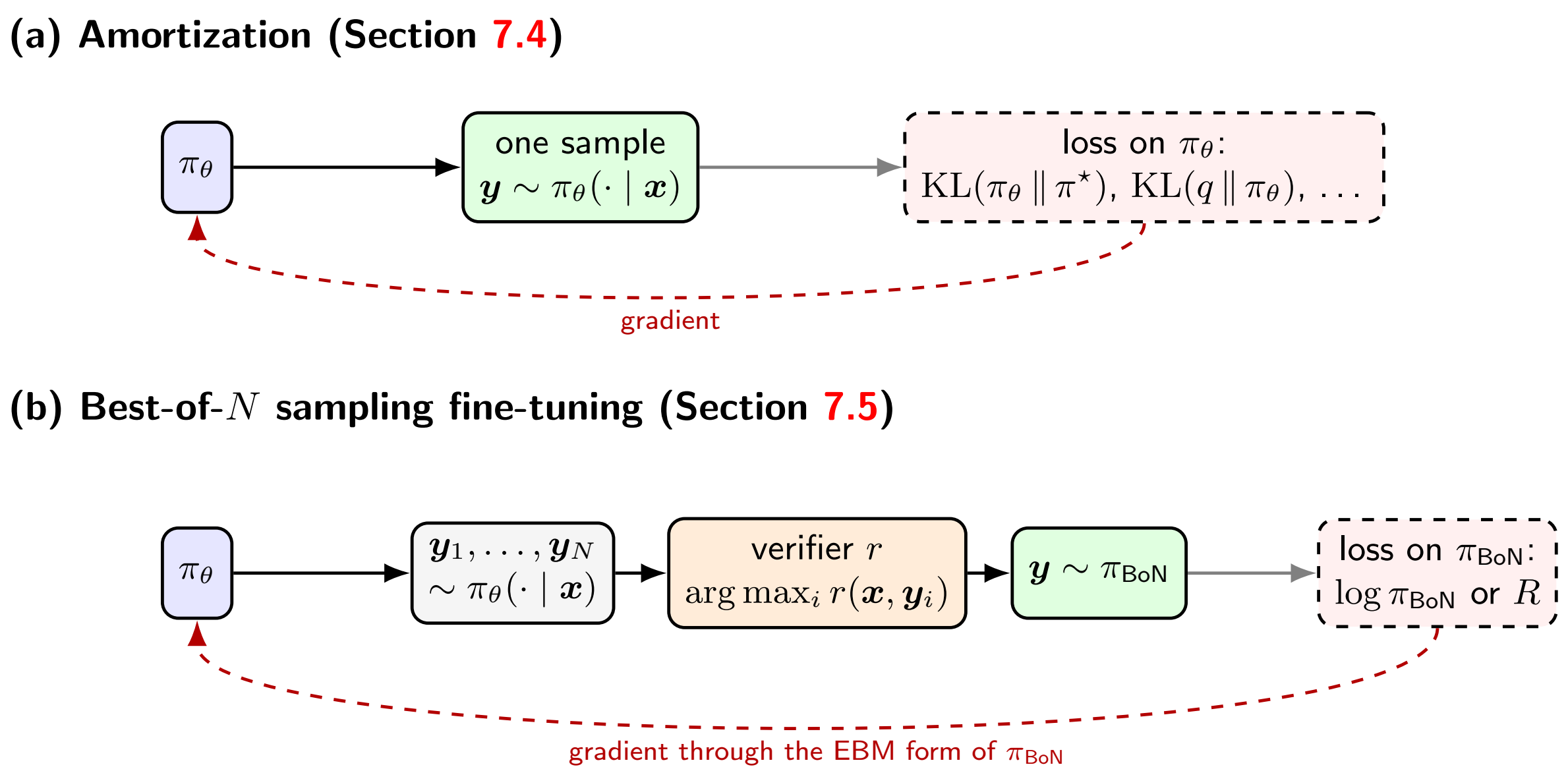
\begin{figure}[t]
    \centering
    \resizebox{0.95\textwidth}{!}{%
    \begin{tikzpicture}[font=\sffamily\small,
        box/.style={draw,rounded corners,thick,minimum height=0.8cm,align=center,inner sep=4pt},
        gen/.style={box,fill=blue!10},
        ver/.style={box,fill=orange!15},
        res/.style={box,fill=green!12},
        loss/.style={box,dashed,fill=red!6},
        arr/.style={-{Latex},thick},
        grad/.style={-{Latex},thick,red!70!black,dashed}]
    \node[anchor=west,font=\sffamily\bfseries] at (-0.6,1.15) {(a) Amortization (\Secref{sec:itc_amortize})};
    \node[gen] (a1) at (1.2,0) {$\pi_\theta$};
    \node[res] (a2) at (4.6,0) {one sample\\$\vy\sim\pi_\theta(\cdot\mid\vx)$};
    \node[loss] (a3) at (9.6,0) {loss on $\pi_\theta$:\\ $\mathrm{KL}(\pi_\theta\,\|\,\pi^\star)$, $\mathrm{KL}(q\,\|\,\pi_\theta)$, \dots};
    \draw[arr] (a1)--(a2);
    \draw[grad] (a3.south) .. controls +(0,-0.9) and +(0,-0.9) .. node[below,font=\sffamily\scriptsize]{gradient} (a1.south);
    \draw[arr,gray] (a2)--(a3);
    \node[anchor=west,font=\sffamily\bfseries] at (-0.6,-2.15) {(b) Best-of-$N$ sampling fine-tuning (\Secref{sec:itc_bon})};
    \node[gen] (b1) at (1.2,-3.6) {$\pi_\theta$};
    \node[box,fill=gray!8] (b2) at (4.0,-3.6) {$\vy_1,\ldots,\vy_N$\\$\sim\pi_\theta(\cdot\mid\vx)$};
    \node[ver] (b3) at (6.7,-3.6) {verifier $r$\\$\arg\max_i r(\vx,\vy_i)$};
    \node[res] (b4) at (9.2,-3.6) {$\vy\sim\pi_{\text{BoN}}$};
    \node[loss] (b5) at (12.2,-3.6) {loss on $\pi_{\text{BoN}}$:\\ $\log\pi_{\text{BoN}}$ or $R$};
    \draw[arr] (b1)--(b2); \draw[arr] (b2)--(b3); \draw[arr] (b3)--(b4); \draw[arr,gray] (b4)--(b5);
    \draw[grad] (b5.south) .. controls +(0,-1.2) and +(0,-1.2) .. node[below,font=\sffamily\scriptsize]{gradient through the EBM form of $\pi_{\text{BoN}}$} (b1.south);
    \end{tikzpicture}}
    \caption{\textbf{Two ways to fine-tune for inference-time computation.} (a) Amortization trains $\pi_\theta$ so that its own samples approximate the target, and inference draws a single sample. (b) Best-of-$N$ sampling fine-tuning keeps best-of-$N$ at inference time and trains $\pi_\theta$ through the distribution $\pi_{\text{BoN}}$ of the selected response, whose EBM form~\Eqref{eqn:itc_bon_ft} replaces the non-differentiable selection.}
    \label{fig:itc_bon}
\end{figure}

One round of best-of-$N$ distillation~(\Secref{sec:itc_amortize})
makes $\pi_\theta$ only as good as best-of-$N$ of $\pi_0$. To improve
further, iterative BOND distills best-of-$N$ of the distilled model again, round
after round, and by~\citet{yang2025faster} these rounds reach the WIND
equilibrium~\Eqref{eqn:itc_wind} only after distilling repeatedly
until convergence. This motivates a different
approach: keep best-of-$N$ with a fixed $N$ at inference time, and
update the base policy for it directly.
Inference-aware fine-tuning~\citep{chow2025inference} places the loss
not on $\pi_\theta$ but on the distribution $\pi_{\text{BoN}}$ of the
response that best-of-$N$ returns from $\pi_\theta$, as compared
in~\Figref{fig:itc_bon}. Since only the best of $N$ samples is used,
$\pi_\theta$ does not need to put most of its mass on good responses;
it only needs one of its $N$ samples to be good, which leaves room for
diverse responses.

\paragraph{Best-of-$N$ Policy as an EBM.} Best-of-$N$ selects by an
$\arg\max$ over $N$ samples, which cannot be differentiated with
respect to $\theta$. Its output distribution, however, has an explicit
density: by~\Eqref{eqn:itc_bon} with $\pi_\theta$ in place of $\pi_0$,
$\pi_{\text{BoN}}$ is a residual EBM on $\pi_\theta$ whose energy is
the $\log$ win rate of $\vy$ against samples of $\pi_\theta$.
\citet{chow2025inference} use the variational approximation
\begin{equation}\label{eqn:itc_bon_ft}
    \pi_{\text{BoN}}(\vy\mid\vx) = \frac{\pi_\theta(\vy\mid\vx)\exp\rbr{\beta_N\,w_\theta(\vx,\vy)}}{Z_\theta(\vx)},
    \qquad
    w_\theta(\vx,\vy) = \EE_{\vy'\sim\pi_\theta(\cdot\mid\vx)}
    \sbr{\mathbf{1}\cbr{r(\vx,\vy)\ge r(\vx,\vy')}},
\end{equation}
with $\beta_N$ increasing in $N$. Since both $\pi_\theta$ and the win
rate $w_\theta$ depend on $\theta$, the unnormalized $\log$-density
$f_\theta = \log\pi_\theta + \beta_N w_\theta$ has the gradient
\begin{equation}\label{eqn:itc_bon_score}
    \nabla_\theta f_\theta(\vx,\vy) = \nabla_\theta\log\pi_\theta(\vy\mid\vx)
    - \beta_N\,\EE_{\vy'\sim\pi_\theta(\cdot\mid\vx)}\sbr{\nabla_\theta\log\pi_\theta(\vy'\mid\vx)\,\mathbf{1}\cbr{r(\vx,\vy)<r(\vx,\vy')}},
\end{equation}
where the second term differentiates the win rate through the samples
of $\pi_\theta$. The two losses below are written in terms of this
gradient.

\paragraph{Supervised Fine-Tuning.} Given expert responses
$(\vx,\vy)\sim\Dcal$, inference-aware supervised fine-tuning maximizes their
likelihood under $\pi_{\text{BoN}}$ rather than under $\pi_\theta$,
\begin{equation}\label{eqn:itc_bon_sft}
    \max_\theta\;\EE_{(\vx,\vy)\sim\Dcal}\sbr{\log\pi_{\text{BoN}}(\vy\mid\vx)}
    = \EE_{(\vx,\vy)\sim\Dcal}\sbr{\log\pi_\theta(\vy\mid\vx) + \beta_N\,w_\theta(\vx,\vy) - \log Z_\theta(\vx)},
\end{equation}
where $Z_\theta(\vx) = \EE_{\vy\sim\pi_\theta(\cdot\mid\vx)}\sbr{\exp\rbr{\beta_N w_\theta(\vx,\vy)}}$.
Its gradient is that of maximum likelihood for an EBM~\Eqref{eqn:gradient_ebm_reduce},
\begin{equation}\label{eqn:itc_bon_sft_grad}
    \EE_{(\vx,\vy)\sim\Dcal}\sbr{\nabla_\theta f_\theta(\vx,\vy)} - \EE_{\vx\sim\Dcal,\,\vy\sim\pi_{\text{BoN}}(\cdot\mid\vx)}\sbr{\nabla_\theta f_\theta(\vx,\vy)},
\end{equation}
in which the expert responses are the positives and the responses
selected by best-of-$N$ are the negatives. The first term increases
the likelihood of the expert responses and their win rate against
$\pi_\theta$, and the second lowers the likelihood of the responses
that best-of-$N$ already selects, which avoids over-fitting.

\paragraph{Reinforcement Learning.} With an environment reward
$R(\vx,\vy)$, which may differ from the verifier score $r$,
inference-aware RL maximizes the reward of the selected response,
\begin{equation}\label{eqn:itc_bon_rl}
    \max_\theta\; J(\theta) = \EE_{\vx\sim\Dcal,\,\vy\sim\pi_{\text{BoN}}(\cdot\mid\vx)}\sbr{R(\vx,\vy)},
    \qquad
    \nabla_\theta J(\theta) = \EE_{\vx\sim\Dcal,\,\vy\sim\pi_{\text{BoN}}}\sbr{\nabla_\theta f_\theta(\vx,\vy)\rbr{R(\vx,\vy)-b(\vx)}},
\end{equation}
with a baseline $b(\vx) = \EE_{\vy\sim\pi_{\text{BoN}}(\cdot\mid\vx)}\sbr{R(\vx,\vy)}$.
This is REINFORCE~\Eqref{eq:reinforce_grad} with the samples drawn
from best-of-$N$ instead of from $\pi_\theta$, and since the reward
is $R$ rather than $r$, it also tolerates errors of the verifier.

Amortization~(\Secref{sec:itc_amortize}) trains
$\pi_\theta$ to approximate the target on its own, whereas best-of-$N$
sampling-aware fine-tuning keeps best-of-$N$ at inference time and
trains $\pi_\theta$ for it. Writing $\pi_{\text{BoN}}$ as a residual
EBM on $\pi_\theta$~\Eqref{eqn:itc_bon_ft} gives the selection a
differentiable density, so supervised fine-tuning becomes maximum
likelihood for this EBM, with best-of-$N$ samples as negatives, and
RL becomes REINFORCE on best-of-$N$ samples.
In both cases the generator is trained as a proposal for best-of-$N$
rather than as a sampler on its own, and it learns to interleave its
best responses with more diverse ones, trading exploitation for
exploration as $N$ grows.

\section{Discussion}\label{sec:discussion}

Throughout this tutorial, we have used EBMs as a single lens on machine learning. This organization grew out of the course \href{https://bo-dai.github.io/CSE6243-fall2024/}{CSE 6243} at Georgia Tech, where students appreciated learning machine learning along a single thread, in which each new method is introduced as a different choice within the same EBM framework, rather than as a catalog of separate model families, tasks, and algorithms, as in the traditional organization of machine learning courses. Any EBM is specified by three choices: the energy, which determines what the model can represent; the inference procedure, which samples from or optimizes the energy; and the learning method, which fits the energy to data. We close by summarizing how this view serves three purposes: understanding existing methods, organizing them into one framework, and developing new algorithms.

\paragraph{Understanding Existing Literature.} Many methods that were developed separately, often under different names and motivations, turn out to be EBMs with particular choices of the three components. In discriminative learning, logistic regression, CRFs, and structured prediction energy networks are conditional EBMs that differ in the structure of the energy over outputs~(\Secref{sec:disc_model}). In generative modeling, GANs pair a potential with a learned sampler, normalizing flows and autoregressive models restrict the energy so that the partition function becomes tractable and sampling takes a single pass, diffusion and flow models learn the score of noise-perturbed EBMs, and energy-inspired and deep equilibrium models turn an inference procedure into the model itself~(\Secref{sec:gen_model}). In representation learning, contrastive methods including MoCo, word2vec, BERT, diffusion representations, and denoising auto-encoders all learn a bilinear EBM whose factors are the representation, and they differ mainly in the learning method; for the denoising auto-encoder, the bilinear form follows from score matching of a latent-variable EBM~(\Secref{sec:repr}). Instruction-following retrievers use the same bilinear EBM as a decisive model, by embedding the question as well as the data~(\Secref{sec:disc_fm}). In decision making, maximum-entropy RL, mirror-descent policy updates, and KL-regularized fine-tuning of language models from human feedback yield EBM policies in closed form, and inverse RL and imitation learning fit EBMs from demonstrations~(\Secref{sec:rl}). Beyond the policy, an EBM world model of the dynamics gives a representation in which value functions are linear~(\Secref{sec:ebm_world_model}), and planning is inference in an EBM over trajectories~(\Secref{sec:planning}). Inference-time computation in large models, such as best-of-$N$ selection, search over partial solutions, and iterative refinement, consists of samplers for one residual EBM that composes a pretrained generator with a verifier~(\Secref{sec:inference_time}). Seen this way, the differences among these methods are differences in how they handle the partition function and sampling, rather than differences in what they model.

\paragraph{Organizing the Framework.} The three components also organize the methods into a common structure. On the learning side, maximum likelihood, contrastive divergence, NCE, pseudolikelihood, score matching, and adversarial training share the same gradient and differ only in the sampler that produces the negative samples~(\Secref{sec:unified}). The same axis reappears in every application: SimCLR and CLIP use ranking NCE, SigLIP and word2vec use binary NCE, AmorLIP uses maximum likelihood with an amortized partition function, MoCo uses a fixed-point iteration, and diffusion representations and denoising auto-encoders use score matching~(\Secref{sec:repr}); maximum-entropy IRL, implicit behavioral cloning, and GAIL fit EBMs from demonstrations by maximum likelihood, NCE, and adversarial training~(\Secref{sec:irl}). On the inference side, MCMC, variational inference, and amortized samplers appear in the same roles, whether the samples are images, tokens, or actions~(\Secref{sec:inference},~\Secref{sec:ebm_policy},~\Secref{sec:inference_time}), and fine-tuning and inference-time computation target the same KL-regularized EBM, differing only in whether its cost is paid during training or at inference~(\Secref{sec:itc_amortize}), and best-of-$N$ sampling fine-tuning lies between the two, training the generator for the sampler that uses it~(\Secref{sec:itc_bon}). Placing a method on these axes makes its trade-offs explicit, such as the dependence of NCE on the batch size, the bias of short MCMC chains, or the loss of the global energy in autoregressive reparameterizations, and shows which alternatives are available for each component. Table~\ref{tab:overview} in Appendix~\ref{app:overview} places every method of this tutorial on these components.

\paragraph{Developing New Algorithms.} Finally, the framework suggests new algorithms by replacing one component while keeping the others. Replacing the learning method of a known EBM gives new estimators: fitting the bilinear representation EBM by score matching instead of NCE gives diffusion spectral representations, and by maximum likelihood with an amortized partition function gives AmorLIP~(\Secref{sec:repr}). Replacing the policy class in RL by an EBM, and its learning by reweighted score matching, gives diffusion policy optimization~(\Secref{sec:ebm_policy}). Using the structure of the EBM itself gives new objectives: the smoothed Bellman equation of SBEED, and the value-incentivized regularizers of VAC and VPO, which use the $\log$-partition function of the EBM policy for optimism and pessimism~(\Secref{sec:reg_rl}). Placing the energy on the dynamics gives EBM world models, whose representation makes value functions linear and learnable by temporal-difference learning~(\Secref{sec:ebm_world_model}). Treating the best-of-$N$ policy itself as a residual EBM gives objectives for distilling it and for fine-tuning the generator for it~(\Secref{sec:itc_amortize},~\Secref{sec:itc_bon}). Using the learned energy itself rather than the representation, with an embedding head for the query and the instruction, gives decisive models that answer new questions in a single forward pass~(\Secref{sec:disc_fm}). Composition is perhaps the most distinctive capability: since the energies of independently trained models can be added to form products of distributions~(\Secref{sec:compose}), as already exploited by classifier and classifier-free guidance~(\Secref{sec:diffusion}), EBMs provide a principled way to combine pretrained models, constraints, and verifiers at inference time without retraining. In planning, learned dynamics, behavior priors, goals, and constraints each contribute an energy term, and a new task is a new sum of terms~(\Secref{sec:planning}); in reasoning, energies learned for individual constraints are composed to solve problems larger than those seen in training~(\Secref{sec:itc_compose}). We expect this view to remain useful as new models are developed: whenever a model defines an unnormalized score over its outputs, the tools for inference and learning surveyed here apply to it directly.

\paragraph{Limitations.} The EBM view explains how methods relate, but it does not by itself tell which one works better. Two methods that fit the same energy can behave very differently in practice, since their performance also depends on how the energy is parameterized, on inductive biases such as the architecture and, in contrastive learning, the data augmentations that define the positive pairs, and on the statistical and computational properties of the estimator. The intended use also matters. Methods with the same optimum may serve different purposes, as the denoising auto-encoder and a diffusion model fit the same score at a given noise level but are used for representation and for sampling, respectively; and different uses favor different learning methods: maximum likelihood when calibrated likelihoods are needed, as for density evaluation, score matching when the model is used for sampling, since it learns the score that the sampler follows, and NCE when only relative scores between candidates matter, as in retrieval and ranking. Composition has limitations of its own: the composed distribution is meaningful only when the component energies have compatible scales and the product is normalizable, and sampling from or optimizing the composed landscape can be much harder than for each component. We therefore suggest that readers weigh these factors together for their own setting, and choose the parameterization of the energy and the methods for learning and inference to the characteristics of the problem.

\section*{Acknowledgements}
This paper is supported by NSF ECCS-2401391, ONR N000142512173, NSF IIS-2403240, NSF IIS-2552007, DARPA Young Faculty Award, and the Toyota Research Institute.

\bibliographystyle{plainnat}
\bibliography{ref}

\newpage
\appendix
\section{Overview of Methods}\label{app:overview}

Table~\ref{tab:overview} summarizes the methods discussed in this tutorial by the three components of an EBM, the energy, the inference procedure, and the learning method, together with how each handles the partition function. Each entry follows the corresponding section, listed in the last column.

{\footnotesize
\setlength{\tabcolsep}{3pt}
\renewcommand{\arraystretch}{1.25}
\begin{longtable}{>{\raggedright\arraybackslash}p{2.2cm}>{\raggedright\arraybackslash}p{3.0cm}>{\raggedright\arraybackslash}p{2.8cm}>{\raggedright\arraybackslash}p{2.6cm}>{\raggedright\arraybackslash}p{3.0cm}>{\raggedright\arraybackslash}p{1.0cm}}
\caption{Methods in this tutorial as EBMs.}\label{tab:overview}\\
\toprule
Method & Energy & Inference & Learning & Partition function & Sec.\\
\midrule\endfirsthead
\toprule
Method & Energy & Inference & Learning & Partition function & Sec.\\
\midrule\endhead
\bottomrule\endlastfoot
Logistic regression & linear in $f_b(\vx)$ & energy minimization & MLE & exact by enumeration & \ref{sec:crf}\\
Chain CRF & unary and pairwise terms & Viterbi & MLE; max-margin & exact by the forward algorithm & \ref{sec:crf}\\
SPEN & deep, non-decomposable over $\vy$ & projected gradient descent on relaxed $\vy$ & structured SVM; end-to-end & intractable; no probabilistic model & \ref{sec:spen}\\
\midrule
GAN & discriminator as potential & learned generator & adversarial & potential need not be normalized; generator samples the EBM only with entropy regularization & \ref{sec:gan}\\
Normalizing flow & $-\log p$ by change of variables & exact, one pass $\vx=g_\theta(\bm{z})$ & exact MLE, no negative phase & $Z_\theta=1$ for every $\theta$ & \ref{sec:flow}\\
Diffusion / flow matching & family of EBMs $\{p_t\}$, one per noise level & reverse SDE/ODE; annealed Langevin & denoising score matching; velocity regression & drops out of the score & \ref{sec:diffusion}\\
Energy-inspired model & learned $E_\theta$ times proposal $\pi$ & SNIS / Hamiltonian importance sampling & lower bound on sampler likelihood (ranking NCE) & batch denominator $\to Z_\theta$ as $K\to\infty$ & \ref{sec:eim}\\
Modern Hopfield & log-sum-exp energy of stored patterns & CCCP update, \ie, attention & --- & --- & \ref{sec:hopfield}\\
DEQ & fixed point; energy minimizer only if $f$ is a gradient step & root finding & implicit differentiation & --- & \ref{sec:deq}\\
Autoregressive & sum of normalized conditionals & ancestral sampling & exact MLE, per-step cross-entropies & softmax over $|V|$ per step & \ref{sec:ar}\\
\midrule
SimCLR / CLIP & $-\phi_\theta(\vx)^\top\psi_\theta(\vy)$ & --- (energy discarded) & ranking NCE & cancels in softmax; batch sum, biased for small $K$ & \ref{sec:contrastive}\\
SigLIP & $-\phi_\theta(\vx)^\top\psi_\theta(\vy)$ & --- & binary NCE & learned $b=-\log Z_\theta$ & \ref{sec:contrastive}\\
Word2vec & $u_w^\top v_c$ & --- & binary NCE & bias fixed, shift $-\log k$ & \ref{sec:word2vec}\\
MoCo & $\phi_\theta(\vx)^\top\phi^{(\ell)}(\vy)$, target frozen & --- & fixed-point iteration via ranking NCE & cancels in softmax; negatives from queue & \ref{sec:contrastive}\\
AmorLIP & $-\phi_\theta(\vx)^\top\psi_\theta(\vy)$ & --- & MLE & amortized by a light network & \ref{sec:contrastive}\\
BERT & $-\phi_\theta(\tilde\vx)^\top\psi(\vx_M)$, summed over masked positions & --- & MLE at a fixed mask ratio $t_0$ & factorizes over masked positions & \ref{sec:bert}\\
DAE & $h_\theta(\tilde\vx)^\top\psi_\theta(\vx)$ & --- & score matching (Gaussian MLE) & avoided & \ref{sec:diff_repr}\\
Diffusion spectral repr. & $\phi_\theta(\vx)^\top\mu_\theta(\tilde\vy,t)$ & --- & score matching & drops out of the score & \ref{sec:diff_repr}\\
InF-Embed / decisive model & $s_\theta(X,Q)$; $\phi_\theta(X)^\top\psi^o_\theta(Q)$ & sigmoid or softmax over candidates, one forward pass & ranking NCE & over candidates $\mathcal{O}(Q)$ & \ref{sec:disc_fm}\\
\midrule
MaxEnt RL / soft $Q$ & $-Q/\lambda$ & exact softmax; Langevin; amortized SVGD; Gaussian (SAC) & soft Bellman & $V=\lambda\log Z$ & \ref{sec:reg_rl}, \ref{sec:ebm_policy}\\
KL-regularized RL / RLHF & $\pi_0\exp(r/\lambda)$ & closed-form residual EBM & mirror descent (TRPO/PPO); DPO & cancels in DPO & \ref{sec:reg_rl}\\
Diffusion policy & $-Q/\lambda$ & reverse diffusion & reweighted score matching & absorbed by the reweighting & \ref{sec:ebm_policy}\\
MaxEnt IRL & $-$return over trajectories & soft-optimal policy & MLE & $Z_\theta=\EE_\rho[\exp V_\theta]$ & \ref{sec:irl}\\
IBC / GAIL & policy energy $E_\theta(s,a)$ & energy minimization / policy sampler & ranking NCE / adversarial & cancels (NCE); avoided (adversarial) & \ref{sec:irl}\\
EBM world model & $-\phi_\theta(s,a)^\top\nu_\theta(s')$ & no sampling: representation and TD & ranking NCE; score matching & linearized by random features & \ref{sec:ebm_world_model}\\
Trajectory planning & dynamics, behavior prior, cost, and goal & MPPI/CEM; Langevin; diffusion; search & terms learned or specified separately & $Z=\EE_{p_0}[\exp(R/\lambda)]$ & \ref{sec:planning}\\
\midrule
Best-of-$N$ / SMC / power sampling & $\pi_0\exp(r/\lambda)$; $\pi_0^\alpha$ & rejection sampling; IS; SMC; MH & --- & IS weights; cancels in MH & \ref{sec:itc_samplers}\\
BBox-Adapter & $\pi_0(\vy\mid\vx)\exp(g_\theta(\vx,\vy))$ & $\pi_0$ proposes, $g_\theta$ scores in beam search & ranking NCE with $\pi_0$ as noise & $\log\pi_0$ cancels from the logits & \ref{sec:itc_target}\\
IREM / IRED / EBT & learned energy $E_\theta(\vx,\vy)$ over complete outputs & gradient descent with restarts & few-step descent; annealed landscapes & --- & \ref{sec:itc_samplers}\\
Compositional reasoning & MRF sharing one learned energy & parallel energy minimization & denoising and contrastive, per constraint & --- & \ref{sec:itc_compose}\\
RL / rejection-sampling fine-tuning & $\pi_0\exp(r/\lambda)$ & less search, sometimes a single sample & reverse KL (RL); forward KL (rejection sampling) & $Z(\vx)$ drops out & \ref{sec:itc_amortize}\\
BOND & $\pi_{\text{BoN}}$, energy $\log F_\vx(r)$ on $\pi_0$ & a single sample & Jeffreys divergence to $\pi_{\text{BoN}}$: SFT on best-of-$N$ samples and RL with $\log F_\vx(r)$ & $\pi_{\text{BoN}}$ normalized in closed form & \ref{sec:itc_amortize}\\
Iterative BOND & best-of-$N$ of the previous model & a single sample & BOND repeated with a moving anchor & normalized in closed form & \ref{sec:itc_amortize}\\
WIND & win rate against the policy itself, on $\pi_0$ & a single sample & fixed point of iterative BOND, with sample-efficient solvers & --- & \ref{sec:itc_amortize}\\
Inference-aware fine-tuning & $\pi_\theta\exp(\beta_N w_\theta)$, $w_\theta$ the win rate & best-of-$N$ over $\pi_\theta$ & likelihood or reward under $\pi_{\text{BoN}}$ & samples of $\pi_{\text{BoN}}$ in the gradient & \ref{sec:itc_bon}\\
\end{longtable}}

\end{document}